%% file: main.tex
\documentclass[10pt,twocolumn,letterpaper]{article}

\input{style/packages}

\cvprfinalcopy 

\ifcvprfinal\fi

\begin{document}

\title{Identity-Disentangled Neural Deformation Model for Dynamic Meshes}

\author[1,2,*]{Binbin Xu}
\author[1,*]{Lingni Ma}
\author[1]{Yuting Ye}
\author[1]{Tanner Schmidt}
\author[1]{Christopher D. Twigg}
\author[1]{Steven Lovegrove}

\affil[1]{Facebook Reality Labs}
\affil[2]{Imperial College London}

 \twocolumn[{%
 \renewcommand\twocolumn[1][]{#1}%
 \maketitle
  \begin{center}
 \centering
   \begin{tikzpicture}[inner sep=0pt]
     \node(p1) at(0,0){\includegraphics[width=\textwidth]{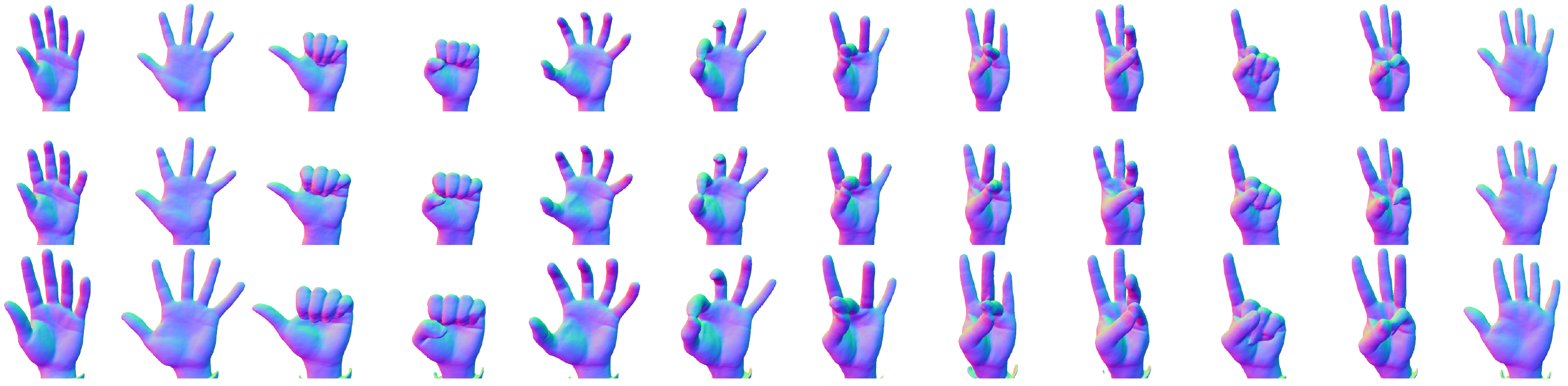}};
   \end{tikzpicture}
   \captionof{figure}{We propose \method to model deformable shapes where identity related deformation is disentangled into separate embedding spaces. This figure demonstrates \method for hand modeling, where 12 different deformation codes drive 2 identity codes (top and bottom) to perform the same gestures. The mid row shows the same deformation applying on the interpolated identity code. }

 \label{fig:teaser}
 \end{center}
 }]

\renewcommand*{\thefootnote}{\fnsymbol{footnote}}
\footnotetext{* The first two authors contributed equally. }
\renewcommand*{\thefootnote}{\arabic{footnote}}
\setcounter{footnote}{0}


\begin{abstract}
\input{section/abstract}

\end{abstract}

\input{section/introduction}

\input{section/related_work}

\input{section/method}

\input{section/experiments}

\input{section/conclusions}

{\small
	\bibliographystyle{ieee_fullname}
	\bibliography{robotvision}
}

\clearpage
\end{document}


\title{\large{Identity-Disentangled Neural Deformation Model for Dynamic Meshes}\\ \Large Supplementary Material}

\author[1,2,*]{Binbin Xu}
\author[1,*]{Lingni Ma}
\author[1]{Yuting Ye}
\author[1]{Tanner Schmidt}
\author[1]{Christopher D. Twigg}
\author[1]{Steven Lovegrove}

\affil[1]{Facebook Reality Labs}
\affil[2]{Imperial College London}

\maketitle

This supplementary material contains three sections. \cref{sec:dataset} provides details on the \dataset dataset. \cref{sect:impl} discuss the implementation details of our methods. \cref{sec:exp} provides additional experiments that was not included in the main submission due to the page limit.

\input{supp/dataset.tex}

\input{supp/implement.tex}

\input{supp/experiments.tex}

{\small
	\bibliographystyle{ieee_fullname}
	\bibliography{robotvision}
}

%% file: style/packages.tex
\usepackage{style/cvpr}
\usepackage{tikz}
\usetikzlibrary{positioning}

\usepackage{times}
\usepackage{epsfig}
\usepackage{graphicx}
\usepackage{amssymb, amsthm, amsfonts, amsmath, mathrsfs, mathtools}
\usepackage{array, tabularx, booktabs, multirow}
\usepackage{subcaption}
\usepackage{textcomp}  %
\usepackage{authblk}
                                                
\usetikzlibrary{snakes, decorations, shapes, shapes.arrows, chains, positioning, shapes.symbols, patterns}

\pgfdeclarelayer{background}
\pgfdeclarelayer{foreground}
\pgfsetlayers{background,main,foreground}

\usepackage{caption}

\usepackage{xcolor}

\usepackage{pifont}

\usepackage{lipsum}

\usepackage[pagebackref=true,breaklinks=true,colorlinks,bookmarks=false]{hyperref}
\hypersetup{colorlinks=true, linkcolor=blue, citecolor=green, filecolor=cyan, urlcolor=cyan}

\usepackage[capitalize]{cleveref}
\crefname{section}{Sec.}{Secs.}
\Crefname{section}{Section}{Sections}
\Crefname{table}{Table}{Tables}
\crefname{table}{Tab.}{Tabs.}

 \definecolor{c1}{HTML}{EC222E}
 \definecolor{c2}{HTML}{F46833}
 \definecolor{c3}{HTML}{FFC417}
 \definecolor{c4}{HTML}{29B04A}
 \definecolor{c5}{HTML}{0B6ED4}
 \definecolor{c6}{HTML}{1620FF}
 \definecolor{c7}{HTML}{7724FF}

\long\def\draft#1{}

\newcommand{\method}{DiForm~}
\newcommand{\dataset}{3DH~}
\newcommand{\datasetDynamic}{3DH-4D~}

\newcommand*\rot{\rotatebox{90}}
\newcolumntype{L}[1]{>{\raggedright\let\newline\\\arraybackslash\hspace{0pt}}m{#1}}
\newcolumntype{C}[1]{>{\centering\let\newline\\\arraybackslash\hspace{0pt}}m{#1}}
\newcolumntype{R}[1]{>{\raggedleft\let\newline\\\arraybackslash\hspace{0pt}}m{#1}}
\input{style/notation-math-defs} %

%% file: style/notation-math-defs.tex
\usepackage{xifthen}
\usepackage{xparse}
\usepackage{subdepth}
\usepackage{leftidx}
\usepackage{bm}
\usepackage{amsmath}

\newcommand{\bbm}{\begin{bmatrix}}
	\newcommand{\ebm}{\end{bmatrix}}

\DeclareMathAlphabet{\mbf}{OT1}{ptm}{b}{n}
\newcommand{\mbs}[1]{{\bm{#1}}}

\newcommand{\mbsbar}[1]{{\overline{\boldsymbol{#1}}}}
\newcommand{\mbshat}[1]{{\hat{\boldsymbol{#1}}}}
\newcommand{\mbstilde}[1]{{\tilde{\boldsymbol{#1}}}}

\newcommand{\mbsdot}[1]{{\dot {\boldsymbol{#1}}}}

\newcommand{\mbfbar}[1]{{\overline{\mbf{#1}}}}
\newcommand{\mbfhat}[1]{{\hat{\mbf{#1}}}}
\newcommand{\mbftilde}[1]{{\tilde{\mbf{#1}}}}

\newcommand{\mbfdot}[1]{{\dot{\mbf{#1}}}}

\DeclareMathAlphabet{\mathbfit}{OML}{cmm}{b}{it}

\newcommand{\trans}[3]{\leftidx{_{#1}}{\mbf t}{\IfValueTF{#2}{_{#2#3\hspace{2pt}}}{}}} %

\newcommand{\vel}[3]{\leftidx{_{#1}}{\mbf v}{\IfValueTF{#2}{_{#2#3\hspace{2pt}}}{}}} %
\newcommand{\veltilde}[3]{\leftidx{_{#1}}{\mbftilde v}{\IfValueTF{#2}{_{#2#3\hspace{2pt}}}{}}} %
\newcommand{\velbar}[3]{\leftidx{_{#1}}{\mbfbar v}{\IfValueTF{#2}{_{#2#3\hspace{2pt}}}{}}} %
\newcommand{\velhat}[3]{\leftidx{_{#1}}{\mbfhat v}{\IfValueTF{#2}{_{#2#3\hspace{2pt}}}{}}} %
\newcommand{\veldot}[3]{\leftidx{_{#1}}{\mbfdot v}{\IfValueTF{#2}{_{#2#3\hspace{2pt}}}{}}} %

\newcommand{\acc}[3]{\leftidx{_{#1}}{\mbf a}{\IfValueTF{#2}{_{#2#3\hspace{2pt}}}{}}} %
\newcommand{\acctilde}[3]{\leftidx{_{#1}}{\mbftilde a}{\IfValueTF{#2}{_{#2#3\hspace{2pt}}}{}}} %
\newcommand{\accbar}[3]{\leftidx{_{#1}}{\mbfbar a}{\IfValueTF{#2}{_{#2#3\hspace{2pt}}}{}}} %

\newcommand{\rotvel}[3]{\leftidx{_{#1}}{\mbs \omega}{\IfValueTF{#2}{_{#2#3\hspace{2pt}}}{}}} %
\newcommand{\rotveltilde}[3]{\leftidx{_{#1}}{\mbstilde \omega}{\IfValueTF{#2}{_{#2#3\hspace{2pt}}}{}}} %
\newcommand{\rotvelbar}[3]{\leftidx{_{#1}}{\mbsbar \omega}{\IfValueTF{#2}{_{#2#3\hspace{2pt}}}{}}} %
\newcommand{\rotvelhat}[3]{\leftidx{_{#1}}{\mbshat \omega}{\IfValueTF{#2}{_{#2#3\hspace{2pt}}}{}}} %
\newcommand{\rotveldot}[3]{\leftidx{_{#1}}{\mbsdot \omega}{\IfValueTF{#2}{_{#2#3\hspace{2pt}}}{}}} %

%% file: section/abstract.tex
Neural shape models can represent complex 3D shapes with a compact latent space. When applied to dynamically deforming shapes such as the human hands, however, they would need to preserve temporal coherence of the deformation as well as the intrinsic identity of the subject. These properties are difficult to regularize with manually designed loss functions. In this paper, we learn a neural deformation model that disentangles the identity-induced shape variations from pose-dependent deformations using implicit neural functions. We perform template-free unsupervised learning on 3D scans without explicit mesh correspondence or semantic correspondences of shapes across subjects. We can then apply the learned model to reconstruct partial dynamic 4D scans of novel subjects performing unseen actions. We propose two methods to integrate global pose alignment with our neural deformation model. Experiments demonstrate the efficacy of our method in the disentanglement of identities and pose. Our method also outperforms traditional skeleton-driven models in reconstructing surface details such as palm prints or tendons without limitations from a fixed template.

\draft{
Hands and bodies are important geometry to model and track for many computer vision applications. 
Many state-of-the-art modeling methods fail to reconstruct the detailed muscle deformation, non-trivial to design and customized, and difficult to apply for dynamic reconstruction. 
In this paper, we aim to reconstruct the complete detailed geometry for hands and bodies, given 3D point cloud scans in arbitrary coordinates. 
To this end, we made three contributions. 
First, we propose an unsupervised template-free approach to learn the shape embedding of hands and bodies. with an explicit regulation for the identity and deformation. 
Second, we propose several solutions to use the learned embedding to track and reconstruct the mesh from 4D point cloud scans. 
Shown with experiments, the proposed modeling achieves an effective embedding separation between intrinsic identity and generic deformation. 
The overall algorithm yields state-of-the-art hand and body modeling, and accurate dynamic reconstruction.
}

%% file: section/introduction.tex
 \section{Introduction}

Hands form one of our most important interfaces with the world and thus modelling and tracking them are an important problem that has recently received significant attention from the computer vision community. 
 However, it is challenging to capture the fine details of hand geometry, due to the complicated interaction between muscles, bones and tendons.  In addition, there are significant variations between individuals' hands.

The most common technique for building a multi-person model of human hands or bodies is to fit an explicit mesh representation to a large collection of scans~\cite{Romero:etal:TOG2017, SMPL-X:2019}.  These models typically combine a statistical shape basis with the standard Linear Blend Skinning (LBS) method for deforming the mesh using the skeleton.  
It is non-trivial to develop these models, which require high quality mesh registration and often rely on manual annotation.  Solving for the resulting statistical model is a complex, nonconvex optimization that requires careful regularization.

With the rapid development of deep learning, many recent works have shown successful and impressive results on learning shape embeddings using a purely data-driven approach. 
One promising direction is to train a multi-layer perceptron (MLP) to implicitly represent the shape representation using e.g. a signed distance field.  
Learning the shape embedding requires that the training data be aligned in order to remove the ambiguity introduced by the 6DoF rigid body or 7DoF similarity transformation. 
This alignment space is referred to as the canonical coordinates and is defined by the training set. 
While it may be possible to align the {\em training} shapes to the canonical coordinates by preprocessing the data, this assumption is invariably violated on real-world {\em testing} data. 
In practice, using the shape embedding to reconstruct novel inputs requires joint estimation of pose and geometry. 

This paper studies the problem of reconstructing identity-invariant deforming shapes from a dynamic sequence of 3D point clouds and provides solutions to apply it in real-world data. Our model produces complete and detailed geometry that captures identity-specific features. We have two major contributions. First, we learn a neural deformation model that factorizes its latent space into subject identity and gesture from unregistered 3D point clouds, without requiring any human registration or annotation. Second, we propose two effective solutions to jointly optimize for a global rigid pose with identity and pose embeddings in the neural deformation model. We conduct extensive ablations to validate our design choice, and show the disentangled latent space can perform identity and pose transfer as well as dynamic shape completion on new subjects. %

%% file: section/related_work.tex
\section{Related Work}

Our goal is not only to learn a disentangled deformable model with detailed geometry, but also to use the presentation to reconstruct unseen data that often misalign with the canonical shape space. 

\paragraph{Surface Deformation Models}

One widely adopted representation for articulated shapes is Linear Blend Skinning (LBS) that deforms a template mesh based on skeleton. Low resolution meshes are more efficient for pose tracking~\cite{Tompson:2014:RTC}, but cannot capture complex deformations or fine details such as wrinkles. Subdividing the mesh can improve tracking convergence \cite{Taylor:2016:EPI,khamis:2015:lem}, but does not increase details. Fixed template mesh also struggles to describe variations among individuals such as palm prints or finger nails. 
To improve the deformation quality, Pose-Space Deformation (PSD) \cite{Lewis:2000:PSD,Allen:2002:ABD,eigenskin} models add pose-dependent shapes as ``correctives'' to LBS results. Similar ideas are used in SMPL~\cite{Loper:2015:SMP} and STAR~\cite{STAR:2020} for the body and in MANO \cite{Romero:etal:TOG2017} for hands. However, these models are still limited by the template mesh resolution.

Recently, neural models offer an alternative approach to represent or deform 3D geometries. The skinning algorithm can be implemented as network modules that map skeletal pose parameters to the final deformed mesh \cite{hasson19_obman,boukhayma20193d}. The flexibility of a neural network enables the learning of more complex skinning functions \cite{Moon:etal:ECCV2020}, or a direct mapping from a latent embedding to a deformed mesh without skeleton \cite{Kulon:etal:BMVC2019}. Such flexibility also opens the door to mesh-less representations. Autodecoder networks can simultaneously learn an embedding space together with a mapping function to signed distance fields \cite{Park:etal:CVPR2019,Gropp:etal:ICML2020} or occupancy grids \cite{Mescheder:etal:CVPR2019, Chen:Zhang:CVPR2019}, where a surface can be extracted up to a desired resolution. These networks can be trained on 3D point clouds, meshes, or even images~\cite{Mildenhall:etal:ECCV2020,Sitzmann:etal:NIPS2020,Saito:etal:ICCV2019}. While initial efforts focus on learning static shapes aligned in a canonical space, they can be extended to represent deforming shapes~\cite{Niemeyer:etal:ICCV2019,deng2019neural,Karunratanakul:etal:3DV2020,yang2021s3}. 
Our work is closely related to these efforts, where we extend \cite{Park:etal:CVPR2019, Gropp:etal:ICML2020} to incorporate the 6D global transformation of deforming shapes, and further disentangle the shape space from the identity space to preserve subject-specific invariance throughout a motion. 

\paragraph{Identity-disentangled Deformable Models}
Thanks to the speed and simplicity, bilinear deformable models are widely used in pose and shape estimation for hands~\cite{Romero:etal:TOG2017}, bodies~\cite{Loper:2015:SMP}, and faces~\cite{Paysan:2009:FMP,Cao:2014:FWF,FLAME:SiggraphAsia2017}. These methods use a global linear shape basis to represent identity but are posed using LBS. They are typically built by registering a template mesh to a large 3D shape collection which capture multiple subjects with a variety of poses~\cite{Allen:2006:LCM}. Inverting the skinning transform brings the meshes into a consistent space where a statistical model such as Principle Component Analysis (PCA) can disentangle the per-subject identity from pose-dependent corrective shapes. The quality of the resulting model depends critically on the accurate template registration, which often requires manual annotations.

Neural networks are able to learn disentangled nonlinear deformable models. Zhou et al.~\cite{Zhou:etal:ECCV2020} models disentangled shape and pose from registered meshes via self-consistency loss. This approach can be applied to different subjects without domain specific designs. I3DMM \cite{yenamandra2020i3dmm} learns disentangled latent spaces of identity, face expressions, hairstyle, and color from watertight aligned face scans. Similarly we learn a neural deformable model from scans with disentangled identity and shapes. NPMs~\cite{palafox2021npms} disentangle shape and pose by learning the deformation field, which requires dense correspondences annotation that is non-trivial to obtain. Our work assumes the training scans are rigidly aligned, but needs no hole-filling, template registration, nor correspondence annotations across subjects. Unlike I3DMM \cite{yenamandra2020i3dmm} which focuses on faces, we evaluate on hands with a high degree of articulation.

\paragraph{3D Scan Neural Registration}
While a neural deformable model is conceptually similar to a bilinear model, it is not obvious how to incorporate global transformation into a neural shape model. While most existing work only demonstrates reconstruction in a canonical shape space~\cite{yenamandra2020i3dmm, Niemeyer:etal:ICCV2019,deng2019neural,yang2021s3}, there have been several concurrent exciting work fitting 3D models to 3D scans using implicit representations. Most approaches~\cite{Bhatnagar:etal:NIPS2020, Saito:CVPR:2021, LEAP:CVPR:21} combine conventional template models with neural implicit representations to get the best of both worlds. We instead explore a purely implicit representation without any templates. It allows us to learn unstructured and detailed features that would be difficult to capture by a predefined template. We bypass the template registration completely, and need not to define a consistent ``rest pose" among different identities. %

%% file: section/method.tex
\section{Method}
The overview of DiForm is presented in \cref{fig:overview}. This section describes each module in details.

\subsection{Learning Embedding for Deformable Shapes}\label{sec:embedding}

\input{section/pipeline.tex}

Consider a shape set $\mathcal{M}=\{\mathcal{M}_{ij}\}$, where $ij$ indexes the $j$-th sample from the $i$-th subject. For deformable shapes, each subject has a different intrinsic geometry that can deform in a similar manner. An example set is $X$ hands each with $Y$ different gestures, or $X$ faces each with $Y$ different expressions. We assume the set contain 3D point clouds with normals, $\mathcal{M}_{ij} = \{\boldsymbol{x}^{ij}_k\in\mathbb{R}^3, \boldsymbol{n}^{ij}_k\in\mathbb{R}^3\}_{k\in\mathbb{Z}}$. 
The prior works \cite{Park:etal:CVPR2019, Gropp:etal:ICML2020, sitzmann2019metasdf} learn the shape space using MLPs that encode SDF.
To recap, these MLPs map a 3D location and a latent \emph{shape code} $\boldsymbol{z}$ to its SDF value.
Shape information can be extracted to meshes by marching cubes or depth maps by sphere tracing, both of which query the network at many spatial locations while holding the shape code fixed.

This work focus on deformable shapes. We thus aim to distentangle the two main factors of variation: the intrinsic shape of the subject and its current configuration.
Thus, rather than a single latent code $z$, we condition the MLP on two latent codes.
The first, $\boldsymbol{z}_I\in\mathbb{R}^N$, is referred to as the identity code and is meant to store intrinsic shape information.
The second, $\boldsymbol{z}_D\in\mathbb{R}^M $, is the deformation code and is used by the network along with $\boldsymbol{z}_I\in\mathbb{R}^N$ to pose a concrete shape.
Mathematically, we learn the function
\begin{equation}\label{eq:mlp}
    f(\boldsymbol{x}; \boldsymbol{z}_I^{i}, \boldsymbol{z}_D^{ij}; \boldsymbol{\phi}) = \varphi(\boldsymbol{x}) \;,
\end{equation}
where $\varphi(\boldsymbol{x}): \mathbb{R}^3\mapsto \mathbb{R}$ is the SDF of shape $\mathcal{M}_{ij}$ at $\boldsymbol{x}$, and $\boldsymbol{\phi}$ is the network parameters. 
Similar to previous works \cite{Park:etal:CVPR2019, Gropp:etal:ICML2020, sitzmann2019metasdf}, we model $f(\boldsymbol{\phi})$ by a decoder MLP. At training, the network parameters and the latent codes are optimized by back-propagating gradients from the loss. 

We assume knowledge of which shapes belong to the same subject $i$. However the knowledge of corresponding deformation across subjects are unknown, due to the fact that configuration space is continuous and it requires expensive manual annotation. To this end, shapes from the same subject share a common identity code $\boldsymbol{z}_I^{i}$, while all shapes maintain its own deformation code $\boldsymbol{z}_D^{ij}$. 
By forcing the network to share the same identity code $z_I^i$ for all shapes belonging to subject $i$, we prevent the optimizer from storing pose-specific information in these latent codes.
As a corollary, all variation between different configurations with the same identity must be encoded in the shape codes $z_D^{ij}$ for various $j$.
We do not have any corresponding mechanism to prevent the optimizer from storing identity-specific information in the shape codes, because they are not shared among individuals. In theory, it is possible for all information about a shape be encoded in codes $z_D^{ij}$. However, our experimental results indicate this does not happen in practice (c.f. section \ref{sec:experiments}).

Inspired by IGR~\cite{Gropp:etal:ICML2020} and MetaSDF~\cite{sitzmann2019metasdf}, we solve the following optimization at training:
\begin{equation}\label{eq:loss}
   \underset{\mathcal{Z}_I, \mathcal{Z}_D, \boldsymbol{\phi}}{\arg\min} \, E
   = \underset{\mathcal{Z}_I, \mathcal{Z}_D, \boldsymbol{\phi}}{\arg\min} \,
   E_{\Omega_0} + 
   E_{\varphi} + 
   E_Z \;\;.
\end{equation}
The variable $\mathcal{Z}_I = \{\boldsymbol{z}_I^{i}\}$ and $\mathcal{Z}_D = \{\boldsymbol{z}_D^{ij}\}$ denotes all identity and deformation codes. The loss function $E$ contains three parts. The term 
\begin{equation}
	\label{eq:loss-manifold}
	E_{\Omega_0} = 
	\underset{\boldsymbol{x}\in\Omega_0}{\sum} \lambda_{s} |f(\boldsymbol{\phi}) | 
	+ \lambda_{n} \left(1- \left|\mathbf{n}^\top  \nabla_{\boldsymbol{x}} f(\boldsymbol\phi)\right|\right) \;,
\end{equation}
supervises the training with the point clouds. It enforces the SDF to be zero at the surface locations, i.e., $\varphi(\boldsymbol{x}) = 0, \forall \boldsymbol{x}\in\Omega_0$. The term $E_{\Omega_0}$ also encourages the predicted normal, as defined by the gradient of the SDF, to be as similar as possible to the input normal.
It is not sufficient to constrain the SDF using only surface points as that leaves the majority of space without supervision.
Therefore, the second term
\begin{equation}
	\label{eq:loss-sdf}
	\begin{aligned}
		\thickmuskip=0mu
		\medmuskip=0mu
		\thinmuskip=0mu
		E_\varphi = \lambda_{g} \underset{\boldsymbol{x}\in\Omega}{\sum}  \left|1-\left\|\nabla_{\boldsymbol{x}} f(\boldsymbol\phi)\right\| \right|
		+\lambda_{i} \underset{\boldsymbol{x}\in\Omega\setminus\Omega_0}{\sum} \exp(-\alpha |f(\boldsymbol{\phi})|)
	\end{aligned}
\end{equation}
is used to regularize the SDF over the continuous 3D space $\Omega$. This is done by enforcing the predicted normal to be a unit vector, which is a general property of all signed distance functions (excepting discontinuities at which the gradient is not well defined). 
In addition, the regularizer $\exp(-\alpha |f(\boldsymbol{\phi})|)$ with the positive hyper-parameter $\alpha\in\mathbb{R}$, also prevents off-surface locations to create zero-isosurface. The third term
\begin{equation}\label{eq:loss-z}
    E_Z = \lambda_{z} \left(\sideset{}{_i}\sum\big\|\boldsymbol{z}_I^i\big\| + \sideset{}{_{ij}}\sum\big\|\boldsymbol{z}_D^{ij}\big\|\right)
\end{equation}
encourages the latent space to be zero-mean and helps to prevent over-fitting. The weights $\lambda_{s}, \lambda_{n}, \lambda_{g}, \lambda_{i}, \lambda_z$ are the hyper-parameters used in the training and inference.

\subsection{Dynamic Reconstruction}
We now describe the inference of identity and deformation on novel observations using the trained model, thereby performing dynamic reconstruction of a deforming shape.

\subsubsection{Single shape inference}\label{sec:single-shape}
We first discuss the inference for a single shape. Following related work on learning SDF, we assume the training shapes are properly aligned such that the network learns the shape space according to a `canonical' coordinates. This greatly simplifies learning by removing the large-scale variations to SDFs caused by similarity transforms. However, it cannot typically be assumed that novel observations are provided in the canonical space. Instead, these observations appear in what we will call the \emph{world} coordinates. We can still apply the canonical model by simply introducing a transformation to align the world to the canonical space and optimizing it in addition to the latent codes.

Without loss of generality, we assume the transformation is described by 6DoF $\mathbb{SE}(3)$. A shape in the world $\mathcal{M}^w=\{\boldsymbol{x}_k^{w}, \boldsymbol{n}_k^{w}\}$ can be transformed to the canonical space by 
\begin{equation}\label{eq:6dof}
    \mathcal{M}^c = \{
    R_{cw}\boldsymbol{x}_k^{w} + \mathbf{t}_{cw} \;,
    R_{cw}\boldsymbol{n}_k^{w} \} \;,
\end{equation}
with rotation $R_{cw}\in\mathbb{SO}(3)$, translation $\mathbf{t}_{cw}\in\mathbb{R}^3$, where the subscript indicates the transformation from world to canonical. Rigid transformation is differentiable. Inserting \cref{eq:6dof} to the energy function of \cref{eq:loss}, optimizing the latent codes $\boldsymbol{z}_I,\boldsymbol{z}_D$ and the pose $R, \mathbf{t}$, yield a joint estimation of the shape and the pose.     

It is straight-forward to optimize this problem with gradient descent for $\boldsymbol{z}_I,\boldsymbol{z}_D$ and $\mathbf{t}$. However, 3D rotations are embedded in a non-Euclidean manifold. To take proper gradient, we use Lie algebra $\mathfrak{so}(3) = \{[\boldsymbol{\omega}]_\times \in \mathbb{R}^{3\times3} | \boldsymbol{\omega}\in\mathbb{R}^3\}$, where the operator $[\cdot]_\times$ generates a skew-symmetric matrix. The exponential map $\exp(\boldsymbol{[\omega}]_\times)$ converts $\boldsymbol{\omega}$ to the rotation matrix $R$, and the log map $\log(R)$ operates vice versa. The Jacobian of the exponential map $\partial\exp({[\boldsymbol\omega]_\times})/\partial\boldsymbol\omega$ is analytically defined for $\boldsymbol\omega=\boldsymbol{0}$, otherwise is unstable and difficult to compute. Therefore, to optimize $\boldsymbol\omega$, we rewrite the energy
$E(R) = E(R\exp([\Delta\boldsymbol{\omega}]_\times))$ with $\Delta\boldsymbol\omega=\boldsymbol{0}$. This leads to update of the rotation with  
$R^{(n+1)} = R^{(n)} \exp(\lambda \partial E /\partial\Delta \boldsymbol\omega)$, where the Jacobian is always evaluated at $\Delta\boldsymbol\omega=\boldsymbol{0}$, and $\lambda$ is the learning rate for gradient descent. This optimization is more stable as opposed to $R^{(n+1)} = R^{(n)} - \lambda\partial E/\partial \boldsymbol\omega$, with Jacobian evaluated at the current estimate.

\subsubsection{Initialization for Joint Optimization}\label{sec:init}
The optimization defined in \cref{sec:single-shape} is highly non-convex, which depends on good initialization to avoid local minima. Our initial experience suggests that it is sufficient to initialize $\boldsymbol{z}_I, \boldsymbol{z}_D$ with the $\boldsymbol{0}$-code and $\mathbf{t}$ with the shape center. 
However, rotation needs to be reasonably accurate to ensure the initial estimate is within the basin of convergence. Our experience suggests the rotation be within about 20$^\circ$ of the global optimum. To find the rotation initialization, we develop two methods: a searching solution using policy gradient (PG) from reinforcement learning (RL); and a learning solution to predict the pose with MLPs. 

{\bf Searching rotation with policy gradient}. Our experiment shows that good initial rotations converges to significantly lower cost as defined by \cref{eq:loss} and vice versa. Based on this observation, we can sample rotations and evaluate the hypothesis to seek a good guess. This strategy demands intense computation to cover the rotation space. We propose to use PG to solve the searching efficiently with a probabilistic formulation.

In the language of RL, a policy $\pi(a|\boldsymbol\theta)$ parameterized by $\boldsymbol\theta$ is the probability of taking action $a$. Each action leads to a reward $\psi(a)\in\mathbb{R}$. For simplicity, we denote the policy with $\pi_{\boldsymbol\theta}$. To find the action trajectory that maximizes the total rewards, PG shapes the action distribution by maximizing the expected rewards $J$ 
\begin{equation}\label{eq:pg}
    J(\boldsymbol\theta)
    = \underset{a\thicksim\pi_{\boldsymbol\theta}}{\mathbb{E}}[\psi(a)]
    =\underset{a}{\sum}\pi_{\boldsymbol\theta} \psi(a) 
    \;,
\end{equation}
with respect to $\boldsymbol\theta$.  With gradient ascent, policy is updated by $\boldsymbol\theta_{k+1} = \boldsymbol\theta_k + \lambda\partial J/\partial{\boldsymbol\theta}$. Note the equation $\partial f/\partial x = f\partial\log f/\partial x$, yield
\begin{equation}\label{eq:pgupdate}
    \frac{\partial J}{\partial \boldsymbol\theta}
    = \underset{a}{\sum} \pi_{\boldsymbol\theta} \frac{\partial\log\pi_{\boldsymbol\theta}}{\partial\boldsymbol\theta} \psi(a)
    = \underset{a\thicksim\pi_{\boldsymbol\theta}}{\mathbb{E}} 
    \Big[\frac{\partial\log\pi_{\boldsymbol\theta}}{\partial\boldsymbol\theta} \psi(a)\Big] \;.
\end{equation}
In RL, it is common for the policy to be rolled out across multiple actions with rewards often deferred to the end.
However, our case is much simpler as the action consists of selecting a rotation and the reward (loss) is determined immediately.
With this approach, we can sample actions to optimize the policy without differentiating $\psi$. 
Consider an action space of $K$ randomly sampled rotations $\mathcal{R} = \{R_i\in\mathbb{SO}(3)\}$, where $\mathcal{R}$ follows a multinomial distribution $\pi(\mathcal{R}|\boldsymbol\theta)$. We can optimize each action $\psi(R_i)$ and $\pi_{\boldsymbol\theta}$ in an alternative fashion. At iteration $k$, we sample $X$ rotations from $\mathcal{R}$ by the distribution $\pi_{\boldsymbol\theta}^{(k)}$. For each $R_i$, we optimize $E(R)$ as defined in \cref{eq:loss} for $Y$ steps to follow the action trajectory and set the cost $\psi(R_i) = E(R_i)$. We then update $\pi_{\boldsymbol\theta}^{(k+1)}$ with the gradient computed by \cref{eq:pgupdate}, and iterate the process till convergence. Starting from a uniform distribution, policy gradient will shift $\pi_{\boldsymbol\theta}$ to center on the preferred rotation. As a result, computation is effectively distributed to the more promising hypothesises. 

{\bf Direct pose prediction with MLP.}
Despite being more efficient than exhausted search, PG still requires evaluating hundreds of hypothesis to update the probability. For fast inference, we train an MLP to directly output the 6DoF pose $\boldsymbol\xi_{cw}$ from the world to the canonical, which we refer to as PoseNet.
To achieve this, one solution is to learn the mapping $g_a(\boldsymbol{x}^{w}) = \boldsymbol\xi_{cw}$, where $\boldsymbol{x}^{w}$ is a point cloud in the world. The problem with this design is that because the canonical coordinates are arbitrarily defined, training can get stuck at local minima. To help the network, we instead learn $g_r(\boldsymbol{x}^{w}, \boldsymbol{x}^{c}) = \boldsymbol\xi_{cw}$, where $\boldsymbol{x}^{c}$ is a \emph{reference} shape in the canonical space. With this modification, the network is changed to learn the relative instead of the absolute pose. Empirically, We found this training behaves better. 
We also bootstrap the training by with the mass center difference $\mathbf{t}^*_{cw}$ of input shapes to roughly shift the world shapes to the canonical space. With the predicted relative rotation $\tilde{R}_{cw}$ and translation $\tilde{\mathbf{t}}_{cw}$, the final transformation between the input shape and the canonical space are composed by $R_{cw} = \tilde{R}_{cw}, \mathbf{t}_{cw}= \tilde{R}_{cw} \mathbf{t}^*_{cw}  + \tilde{\mathbf{t}}_{cw}$.

\subsubsection{Continuous inference}
Given a sequential observation of shape deformation, we achieve dynamic reconstruction by estimating the geometry parameters $\boldsymbol{z}_I, \boldsymbol{z}_D$ and the 6DoF $\boldsymbol\xi_{cw}$ for each time stamp $t$. For this purpose, we adopt the incremental optimization strategy to solve the single shape inference as described in \cref{sec:single-shape}, where optimization at $t+1$ is initialized by the estimates at $t$. At $t=0$, the rotation is first solved by methods in \cref{sec:init}, followed by a full optimization for the shape and the pose. Because our shape embedding models identity and deformation separately, we can alternatively freeze $\boldsymbol{z}_I$ or reduce its learning rate drastically after an identity shape is well observed.

\subsection{\dataset~Dataset}\label{sect:dataset}
In order to evaluate the proposed algorithms, we collect a large number of 3D scans of hands. Similar to the MANO dataset~\cite{Romero:etal:TOG2017}, we use the commercial 3DMD scanner, which directly outputs 4D reconstructions by fusing depth measurements from five synchronized RGB-D cameras. 

Two types of data are collected. First, we capture the left hand from a variety of people. Each participant performs some predefined gestures, such as counting, grabbing, pointing etc. During capturing, participants rest their left arm on a fixed handle to produce aligned 3D meshes without post-processing. We refer this dataset as \emph{\dataset} for 3D hands. In total, 183 subjects are captured, which is randomly split into 150  for training and 33 for testing. We reserve two random samples per training subject for validation. This provides 13820, 300 and 734 total samples for training, validation and testing. 
For the second collection, we remove the arm rest and ask participant to perform random left hand motion freely in the space. A total of 10 sequences from 5 people are captured, who are not included in the previous data capture. We refer this data \datasetDynamic.

%% file: section/pipeline.tex
\begin{figure}

\centering
  \begin{tikzpicture}[inner sep=0pt,font=\small]
  \node(base) at(0,0) {\includegraphics[width=0.49\textwidth]{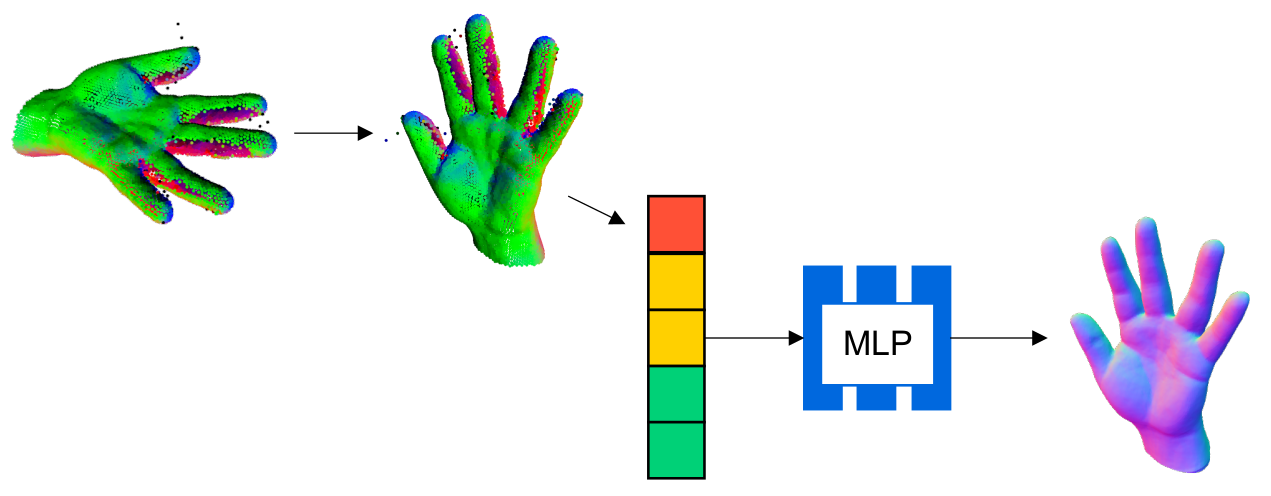}};
  \node(zi) at(-2.2,-0.4)    [text width=30mm, align=right]{\small identity $\boldsymbol{z}_I$};
  \node(zd) at(-2.2,-1.2) [text width=30mm, align=right]{\small deformation $\boldsymbol{z}_D$};
  \node(mc) at(1.2,1.4) [text width=33mm]{$\mathcal{M}^{c} $};
  \node(xi) at(-2.0,0.5) {$\boldsymbol\xi_{cw}$};
  \node(mw) at(-1.2,1.4) [text width=33mm]{$\mathcal{M}^{w}$};
  \node(f) at(1.7,-1.5) [align=center]{\small $f(\boldsymbol{x},\boldsymbol{z}_I, \boldsymbol{z}_D; \boldsymbol\phi)$};
  \draw[-latex] (zi.east)++(0.1,0) -- ++(0.5,0);
  \draw[-latex] (zd.east)++(0.1,0) -- ++(0.5,0);

\end{tikzpicture}
  \caption{DiForm overview. From a set of deformable shapes, we learn the identity code $\boldsymbol{z}_I$ and the deformation code $\boldsymbol{z}_D$ to decode 3D point clouds in the canonical space to SDF. To reconstruct unseen observations in an arbitrary space at inference, we jointly optimize the latent codes and the transformation to align the coordinates. Two solutions are proposed to seek good rotation initialization.}
  \label{fig:overview}
\end{figure}

%% file: section/experiments.tex
\section{Experiments}\label{sec:experiments}

\input{supp/fig_mano_vs_diform.tex}

\input{supp/fig_3dmd.tex}
\paragraph{Implementation details. }
We adopt DeepSDF~\cite{Park:etal:CVPR2019} auto-decoder back bone for shape embedding with the modification to train with disentangle latent space. The other modification is to add the positional encoding~\cite{Mildenhall:etal:ECCV2020} with 10 frequency bands to transform the input. Unless otherwise stated, the identity and the deformation codes are both set to be 64 dimensions. The training shapes are scaled by a global multiplier estimated from the average hand size. For each shape, we randomly sample 16K surface points and 16K off-surface points according to a Gaussian distribution. Adam is used to optimize the network with a constant 0.0005 learning rate for 1000 epochs, with batch size of 24. 

To train the pose prediction, we use a weight-sharing Siamese PointNet~\cite{Qi:etal:CVPR2017} to extract 1024-dimensional feature vector for a given point cloud. The features of query and reference shapes are concatenated before passing to an MLP and output the pose. The learning is supervised with ground truth. To avoid discontinuity, the 6D vector proposed in \cite{Zhou:6drot:2019cvpr} is used to parameterize rotation. At training, the reference shapes are augmented by adding small random perturbation to the learnt $\boldsymbol{0}$-code shape. At inference, the reference shape is set to the $\boldsymbol{0}$-code shape. We randomly sampled 2822 shapes from the \dataset training set. The network is optimized by Adam with a constant learning rate of 0.0005 for 1000 epochs with batch size of 36. 

\subsection{Shape Reconstruction in Canonical Space}
To reconstruct SDF from point clouds in the canonical coordinates, we optimize the latent codes for 2000 epochs with Adam. In each iteration, 16K surface points and 16K off-surface points are sampled. For shapes belonging to the same subject, the identity code can be optimized separately from the single observation or jointly from all available observations. We refer to the separate identity optimization as DiForm-S and the joint version as DiForm-J. 
We compare \method to the state-of-the-art algorithm IGR~\cite{Gropp:etal:ICML2020}, which is the baseline method that learns one latent space. We implement IGR with the same back-bone network and parameters. The latent space is set to be 128 dimensions with the same capacity as DiForm. For fair comparisons, we also train IGR with the positional encoding (denoted as IGR-PE). To compare to the state-of-the-art LBS-based hand modeling, we use MANO~\cite{Romero:etal:TOG2017} baseline. For fair comparison, we used the published result from MANO and subdivide the meshes to have the similar resolution as DiForm, i.e., approximately 100K vertices.

\paragraph{Shape representation power. }
We train the baseline and DiForm on \dataset training set with the same setting. Afterwards we evaluate the reconstruction on the \dataset test set and the MANO \cite{Romero:etal:TOG2017} dataset with 271 left hands from 24 subjects. The results of MANO~\cite{Romero:etal:TOG2017} is computed from the published model. To quantify the performance, the Chamfer Distance (CD) is measured that compares the raw scans and the reconstructions. Because raw scans can contain missing surfaces, we report the sided CD, where $^r_g\text{CD}$ denotes CD from reconstruction to the ground truth and $^g_r\text{CD}$ vice versa. For all methods, we report the mean $\mu$ and the standard deviation $\sigma$, as shown in \cref{tab:eval3d}. 

Overall, training with separate latent spaces yields lower CD in comparison to the LBS-based modeling and one-code training. %
Our method can even fill in the missing holes from the input point cloud by sharing the information between different samples of the same hand by back-propagating gradients to update the same identity code. In \cref{fig:reconstruct_mano}, we visualize the reconstructed meshes from different methods, where DiForm-J yields less noisy reconstructions. The LBS-based MANO~\cite{Romero:etal:TOG2017} algorithm shows very low error on $^r_g\text{CD}$. However, Manos templated meshes are significantly coarser, whereas our model can express intricate muscle deformations (see \cref{fig:mano-dyform}). The template meshes also self-penetrate when fingers pressed into each other, which is not a problem for our implicit SDF representation.

\begin{table}
	\centering
	\footnotesize
    \caption{Quantitative evaluation on canonical shape reconstruction. We report the Chamfer Distance (CD) in millimeter on \dataset~and MANO. }
	\label{tab:eval3d}
	\setlength{\tabcolsep}{2.pt}
	\begin{tabular}{clcccccc}
		\toprule
		&Methods &$\mu(\text{CD})$ &$\sigma(\text{CD})$ &$\mu\left(^r_g\text{CD}\right)$ &$\sigma\left(^r_g\text{CD}\right)$ &$\mu\left(^g_r\text{CD}\right)$ &$\sigma\left(^g_r\text{CD}\right)$ \\
		\midrule
		\multirow{4}{*}{\rot{\shortstack[c]{\dataset}}}
		&IGR~\cite{Gropp:etal:ICML2020}      & 1.1848 &0.3981 & 1.7535 &0.7281 & \textbf{0.6160} &0.0963 \\
		&IGR+PE & 1.1628 &0.3720 & 1.7073 &0.6752 & 0.6182 &\textbf{0.0929} \\
		&DiForm-S  &\textbf{1.0494} &\textbf{0.3060} & 1.4697 &\textbf{0.5257} & 0.6286 &0.1178\\
		&DiForm-J  &1.0594 &0.3180 & \textbf{1.4631} &0.5413 & 0.6552 &0.1157 \\
		\midrule
		\multirow{5}{*}{\rot{\shortstack[c]{MANO}}}
		&Mano~\cite{Romero:etal:TOG2017} &1.9549 &\textbf{0.3988}& \textbf{0.9174} &0.2336& 1.0375 &0.3416\\
		&IGR~\cite{Gropp:etal:ICML2020}  &1.9900 &0.7551 & 3.3365 &1.4862 & \textbf{0.6444} &\textbf{0.1034} \\
		&IGR+PE &2.0340 &0.7265 & 3.3905 &1.4246 & 0.6786 &0.1725 \\
		&DiForm-S  &1.8536 &0.6816& 3.0032 &1.2860& 0.7056 &0.2156 \\
		&DiForm-J  &\textbf{1.8085} &0.6243& 2.8921 &1.2145& 0.7253 &0.1568  \\
		\bottomrule
	\end{tabular}
\end{table}

\begin{table}
	\centering
	\footnotesize
  \caption{Quantitative evaluation of shape reconstruction conditioned on fixed identity code. Chamfer Distance (CD) is in millimeter.}
	\label{tab:eval_condition_identity}
	\setlength{\tabcolsep}{2pt}
	\begin{tabular}{clcccccc}
		\toprule
		&Methods &$\mu(\text{CD})$ &$\sigma(\text{CD})$ &$\mu\left(^r_g\text{CD}\right)$ &$\sigma\left(^r_g\text{CD}\right)$ &$\mu\left(^g_r\text{CD}\right)$ &$\sigma\left(^g_r\text{CD}\right)$ \\
		\midrule
		\multirow{5}{*}{\rot{\shortstack[c]{Validation set}}}
		&IGR~\cite{Gropp:etal:ICML2020}      &1.1001 &0.2516 & 1.6534 &0.5020 & 0.5471 &0.0404 \\
		&IGR-PE      &1.0707 &0.2262& 1.5951 &0.4499 & \textbf{0.5467} &\textbf{0.0396} \\
		&DiForm-S &0.9993 &0.2222 & 1.4455 &0.4187 & 0.5529 &0.0626 \\
		&DiForm-J &0.9977 &0.1983 & 1.4400 &0.3827 & 0.5557 &0.0546 \\
		&DiForm-C &\textbf{0.9955} &\textbf{0.1878} & \textbf{1.4312} &\textbf{0.3695} & 0.5596 &0.0446 \\
		\midrule
		\multirow{5}{*}{\rot{\shortstack[c]{Test set}}}
		&IGR~\cite{Gropp:etal:ICML2020}  & 1.3076 &0.3261 & 1.9410 &0.6292& \textbf{0.6740} &\textbf{0.0746} \\
		&IGR-PE &1.3681 &0.3194& 2.0266 &0.5910& 0.7100 &0.0895 \\
		&DiForm-S &\textbf{1.1920} &0.2821& 1.6770 &0.4953& 0.7071 &0.1039 \\
		&DiForm-J &1.2217 &0.2823& 1.6924 &0.4885& 0.7509 &0.1074 \\
		&DiForm-C &1.2034 &\textbf{0.2533} & \textbf{1.6563} &\textbf{0.4435}& 0.7505 &0.1024 \\
		\bottomrule
	\end{tabular}
\end{table}
\begin{figure}
	\centering
	\def\imw{10ex}
	\def\yoff{2pt}
	\begin{tikzpicture}[inner sep=0pt]
		\node (p00) {\includegraphics[width=\imw]{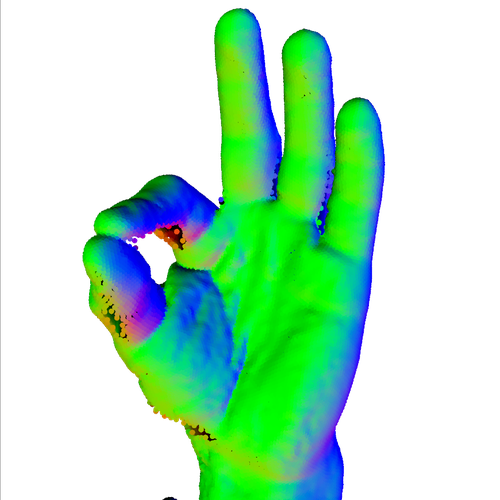}};
		\node (p10) at(p00.east) [anchor=west, xshift=-3ex] {\includegraphics[width=\imw]{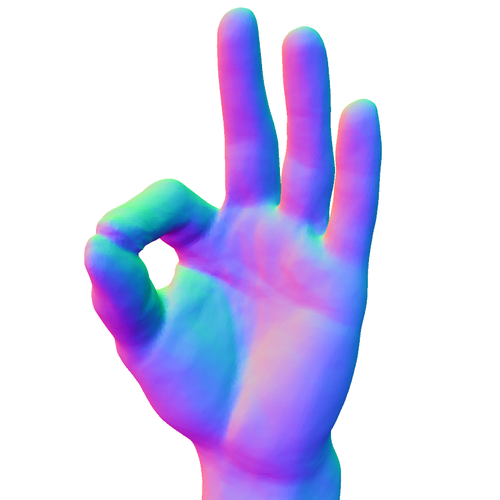}};
		
		\node (p02) at(p10.east) [anchor=west, xshift=-1ex]{\includegraphics[width=\imw]{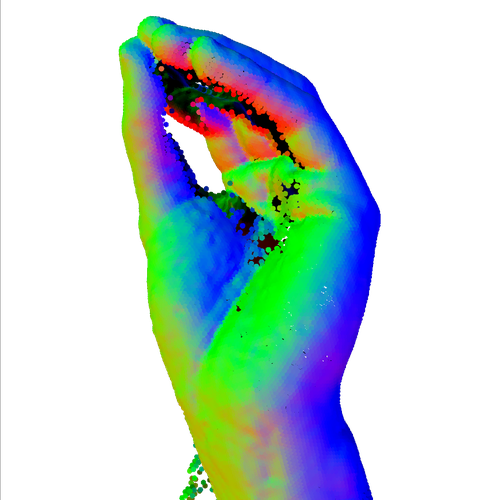}};
		\node (p12) at(p02.east) [anchor=west, xshift=-1ex] {\includegraphics[width=\imw]{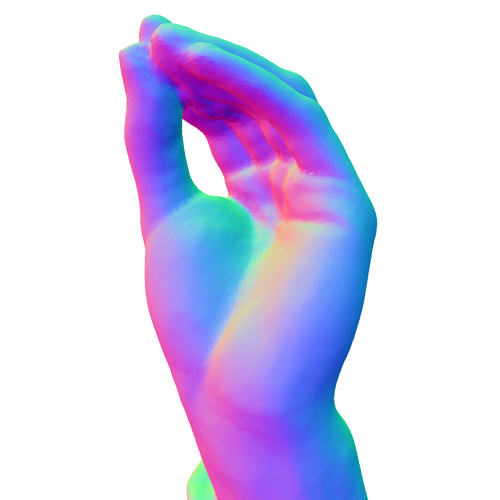}};
		
		\node (p03) at(p12.east) [anchor=west, xshift=-1ex]{\includegraphics[width=\imw]{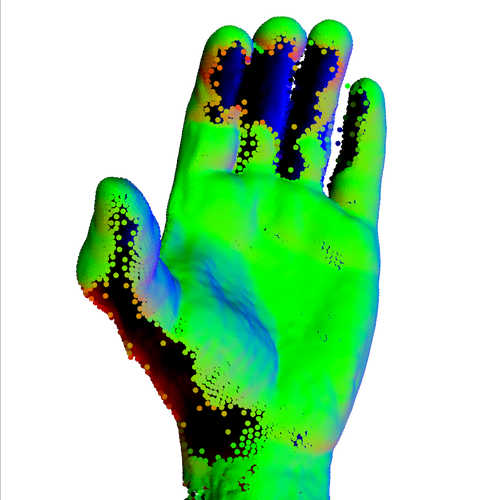}};
		\node (p13) at(p03.east) [anchor=west, xshift=-1ex] {\includegraphics[width=\imw]{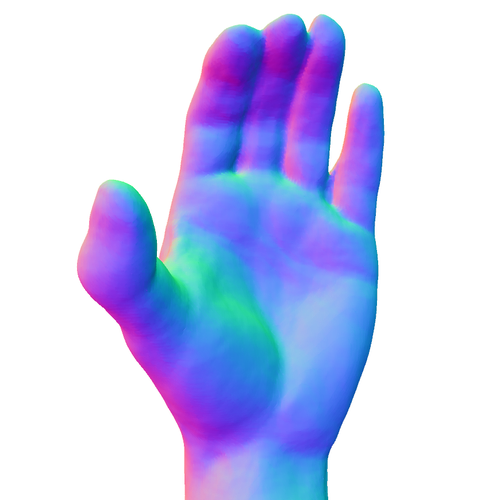}};
		
	\end{tikzpicture}
	\caption{DiForm reconstructs unseen shapes using pre-optimized identity code from the same subjects.} 
	\label{fig:condition_recon}
\end{figure}

\paragraph{Disentangle identity and deformation. }
To examine the separation between the identity and the deformation representations, we designed two experiments. 
First, we examine if the deformation can be transferred to different individuals. To this end, we randomly sampled a set of deformation codes and identity codes that were not paired in training. Then we densely pair each deformation code to each identity code. The decoded shapes can be seen in \cref{fig:teaser}. The result shows that our deformation code can express the same gesture in different identities, even though the identity codes and the deformation codes were not trained together. Similarly, the same identity code combined with different deformation codes show visually similar hand geometry. 
In addition, we linearly interpolate the deformation codes and the identity codes to visualize the interpolated shape and observe smooth transitions in both identity and gesture interpolation trajectories. Additional results are shown in the supplementary material.

We also conducted a quantitative evaluation for the identity-deformation exchange. We randomly select 100 pairs of different poses from different identities and then synthesize a new shape by taking the pose from one identity and combining it with the other identity. We fitted our model to these shapes and computed the L2 distance between the optimized codes and the known input codes. We found the optimized codes are always closest to the ground truth than to other codes sampled from embedding space. The average distance of the optimized identity and deformation codes to all the other input codes is 7.19 and 22.60 respectively, comparing to 2.29 and 4.81 to the ground truth.
This further supports the separation of identity and deformation representations.

In the third experiment, we take an optimized identity code to reconstruct unseen shapes from the same individual while freezing the identity code. To this end, we use the trained identity codes to reconstruct the \dataset validation set, which share the same identities with the training set. The experiment is conducted also on test sequences, where the identity code is first jointly optimized with 20 shapes, before we freeze it to reconstruct the remaining shapes. \cref{tab:eval_condition_identity} present the quantitative comparison (c.f. \cref{fig:condition_recon} for visual inspect). It can be seen that freezing identity code (DiForm-C) outperforms the other methods, where the identity code is optimized together with the deformation code. It shows that the optimised identity code has captured the underlying shape and driven the deformation code to the corresponding gesture, suggesting a good separation of identity and code. The respective reconstructions are shown in the supplementary material.

\begin{figure}
	\centering
	\includegraphics[width=0.485\textwidth]{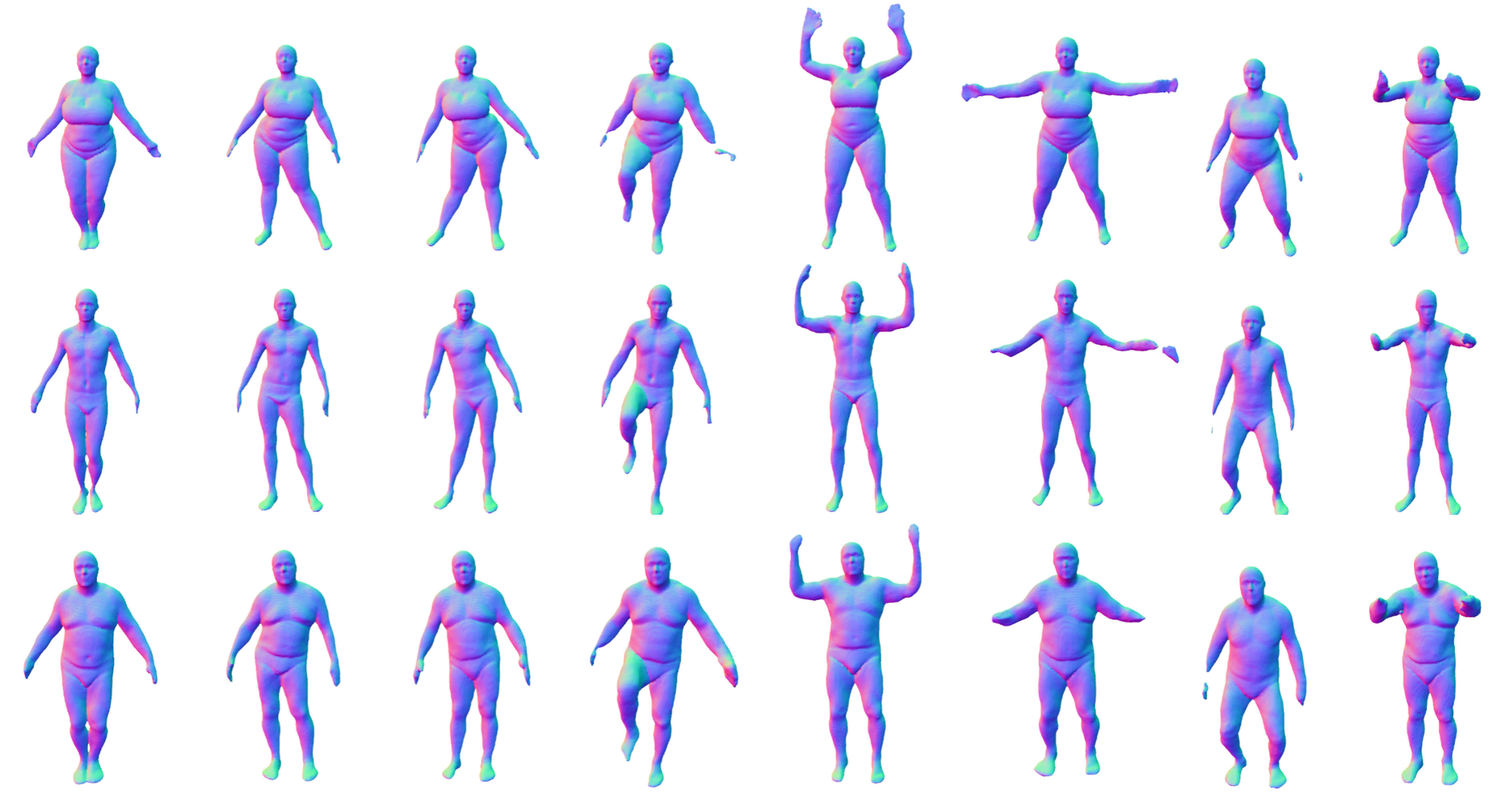}
	\caption{Deformation transfer on the DFaust~\cite{dfaust:CVPR:2017}.}
	\vskip-3ex
	\label{fig:dfaust}
\end{figure}

\paragraph{Generalization to human body.}
We further conducted an experiment to exame how DiForm performs on complex shapes. To this end, we trained DiFrom with the DFaust \cite{dfaust:CVPR:2017} dataset using 11442 body shapes from 9 people. Since there is only few subjects, \cref{fig:dfaust} visualizes the generated shapes by combining 8 random deformation codes with 3 random identity codes. The result show that the deformation codes are capable of driving the identity to deform in the similar fashion. 

\input{section/exp-4dscan.tex}

\begin{table*}
 \centering
    \footnotesize
 \caption{Quantitative evaluation of pose initialization. We compare multi-hypothesis evaluation (MHE), policy gradient (PG) and the pose network prediction (PN). We report CD to the ground truth in the canonical (can.) and in the world. }

 \label{tab:poseinit}
 \setlength{\tabcolsep}{3.5pt}
	\begin{tabular}{c|cccc|cccc}
		\hline
		\multirow{2}{*}{Methods} & \multicolumn{4}{c|}{Initial estimation (mean/standard deviation)}                                            & \multicolumn{4}{c}{After joint shape and pose optimization (mean/standard deviation) }                           \\ \cline{2-9} 
		& RPE: $\mbf R$ (degree) & RPE: $\mbf t$ (mm) & CD: can. (mm) & CD: world (mm) & RPE: $\mbf R$ (degree) & RPE: $\mbf t$ (mm) & CD: can. (mm) & CD: world (mm) \\ \hline
		MHE                   &     17.343/27.118  &  10.344/6.4190   &   7.2924/1.5676     &  7.1615/\textbf{1.2365} &    12.233/28.327 & 6.9173/6.5961  &  3.6215/1.4521 &  1.6347/0.7655  \\
		PG             &     6.2968/4.4954                   &\textbf{5.2287}/\textbf{3.1341}  &  \textbf{7.2891}/\textbf{1.5588 }             &       \textbf{6.7842}/1.3300         &5.4425/3.5907  &   5.1671/3.3695 &     3.0956/\textbf{1.1738}    &  1.4422/0.5960 \\
		PN                   &   \textbf{5.2534}/\textbf{3.3529}  & 5.8876/3.9372 &   7.8876/1.8261  & 7.2946/1.5643 &  \textbf{4.5315}/\textbf{3.1874} & \textbf{4.6401}/\textbf{2.8217}&    \textbf{2.9295}/1.2387 & \textbf{1.3861}/\textbf{0.5221} \\ \hline
	\end{tabular}
\end{table*}

\subsection{Dynamic Reconstruction}

First we evaluate different algorithms on the 6DoF initialization. Lack of baseline algorithms, we construct one that approaximates the exhausted search. To this end, we uniformly sample 2000 rotations as candidate hypothesis for initialization. We ran 400 iterations with 1000 surface points to evaluate each hypothesis and take one corresponds to the lowest loss as initialization. We refer this baseline as MHE. 
To quantify the performance, we generate a dataset with 136 random shapes randomly rotated. In \cref{tab:poseinit}, we report the pose and the geometry error evaluated on the predicted initialization and after the joint optimization. The results show similar performances for PG and PN. The advantage of PG is no training is required and it guarantees to find the solution. The disadvantage is PG takes significantly longer to compute than PN which outputs the prediction with a single forward inference.
\cref{fig:4dscan} shows a demo of DiForm reconstructing a dynamic 4D point cloud in the world. Our method robustly tracked the 6DoF pose and simutaneously optimizes the geometry with fine details.

\subsection{Limitations and Discussions}

\input{supp/fig_failure_case.tex}

\cref{fig:failurecase} shows some failure cases. We observe DiForm is under constrained by highly incomplete point clouds. This drawback is compensated in 4D reconstruction, where DiForm can easily leverage all available data over time. We also observe DiForm struggles to extrapolate shapes, which can be improved by more diverse training data. When linearly interpolating latent codes, the resulting shapes can contain artifacts (c.f. supplementary). This suggests the latent space is not always smooth, causing linear interpolation to deviate from the manifold. %
Since DiForm is template free, it cannot support conventional animation. Despite we demonstrate motion transfer and dynamic reconstruction capability, it is difficult to interpreate DiForm parameters semantically. %

%% file: supp/fig_mano_vs_diform.tex
\begin{figure}
    \centering
    \begin{tikzpicture}[inner sep=1pt]
    \def\imw{9.4ex}
    \node (p00) {\includegraphics[width=\imw]{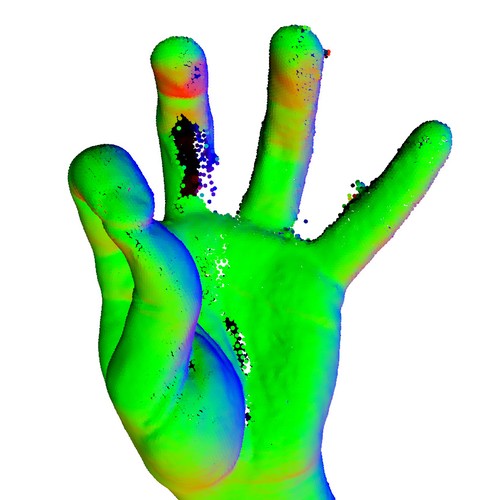}};
    \node (p01) at(p00.south)[anchor=north] {\includegraphics[width=\imw]{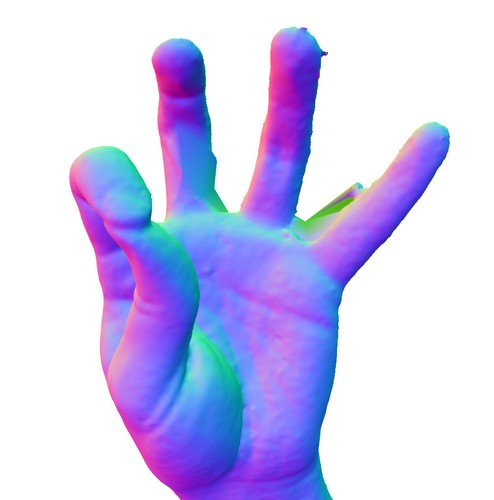}};
    \node (p02) at(p01.south)[anchor=north] {\includegraphics[width=\imw]{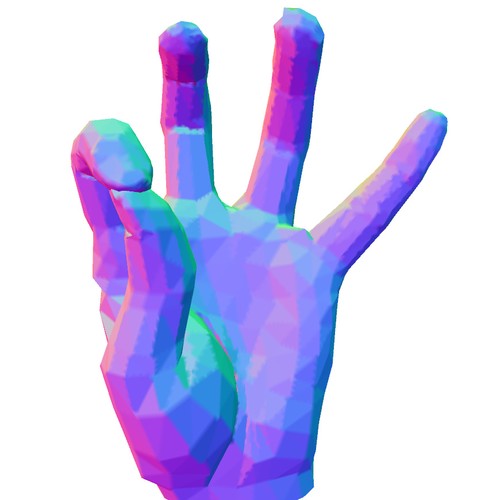}};
    \node (p03) at(p02.south)[anchor=north] {\includegraphics[width=\imw]{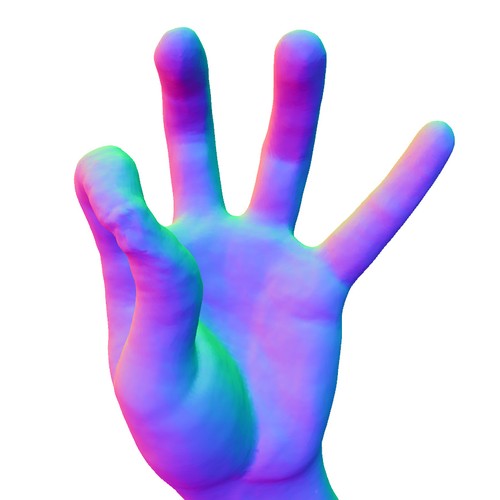}};

    \node (p40) at(p00.east)[anchor=west,xshift=-1.2ex]  {\includegraphics[width=\imw]{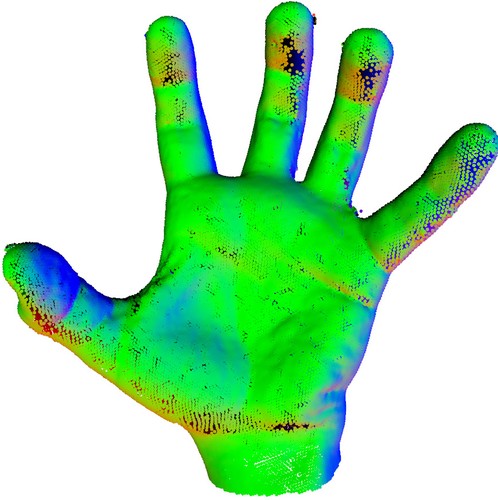}};
    \node (p41) at(p40.south)[anchor=north] {\includegraphics[width=\imw]{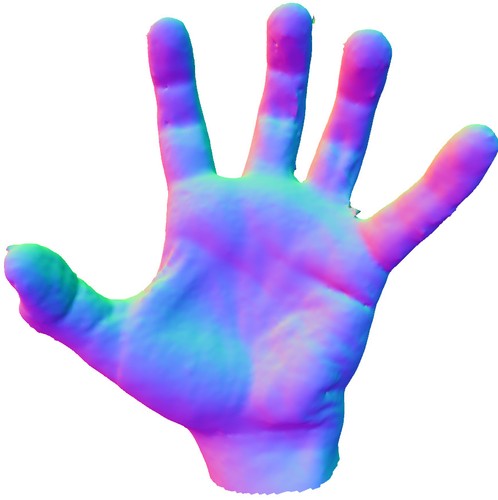}};
    \node (p42) at(p41.south)[anchor=north] {\includegraphics[width=\imw]{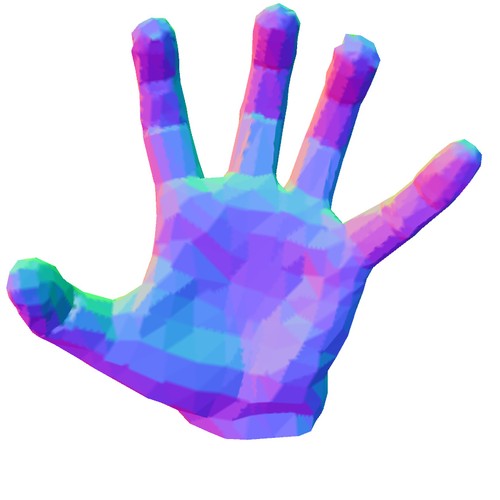}};
    \node (p43) at(p42.south)[anchor=north] {\includegraphics[width=\imw]{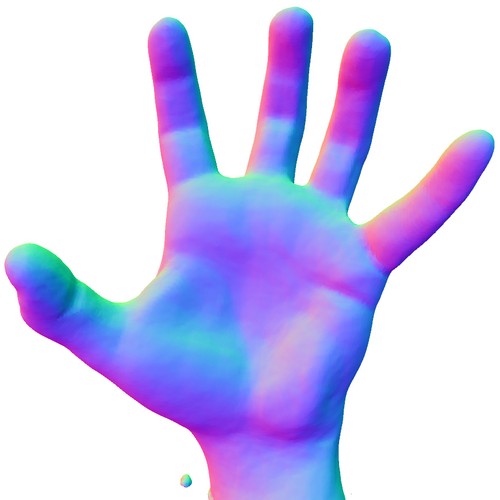}};
    
    \node (p30) at(p40.east)[anchor=west,xshift=-0.45ex]  {\includegraphics[width=\imw]{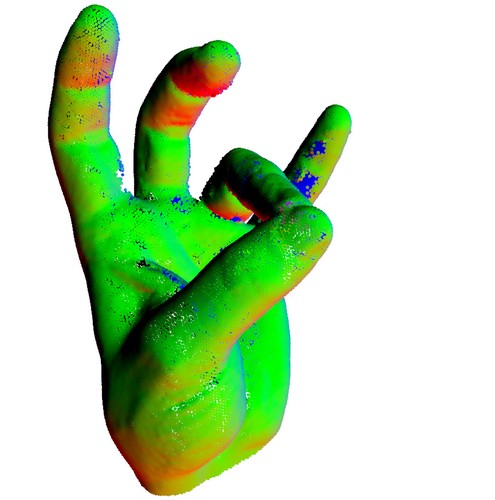}};
    \node (p31) at(p30.south)[anchor=north] {\includegraphics[width=\imw]{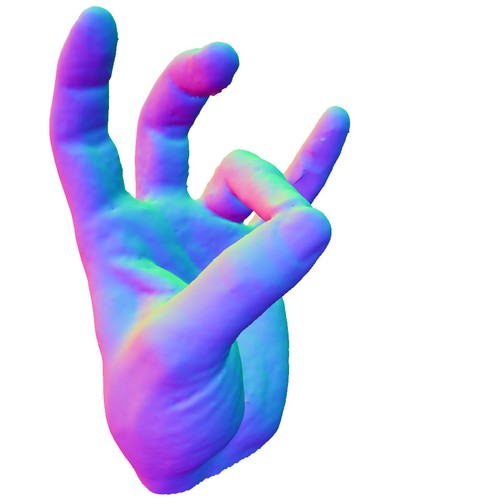}};
    \node (p32) at(p31.south)[anchor=north] {\includegraphics[width=\imw]{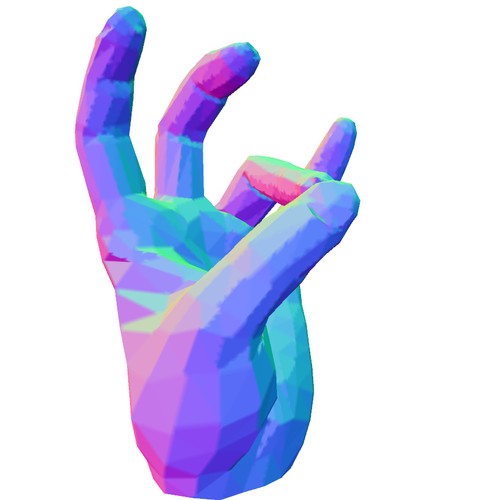}};
    \node (p33) at(p32.south)[anchor=north] {\includegraphics[width=\imw]{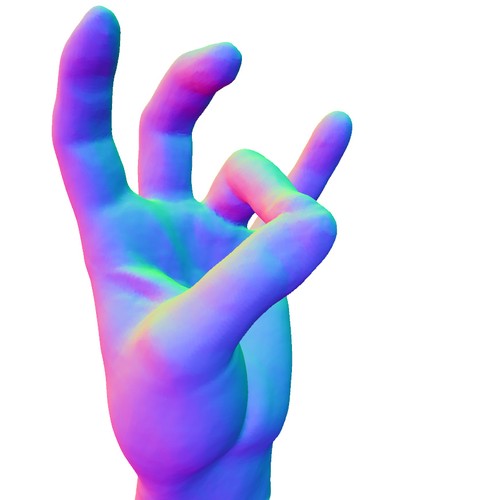}};

	\node (p60) at(p30.east)[anchor=west,xshift=-2.6ex]  {\includegraphics[width=\imw]{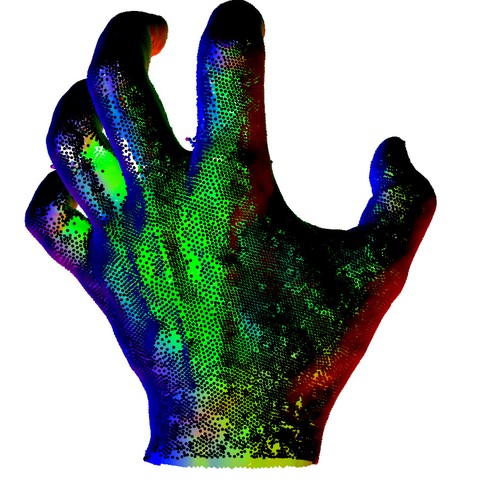}};
	\node (p61) at(p60.south)[anchor=north] {\includegraphics[width=\imw]{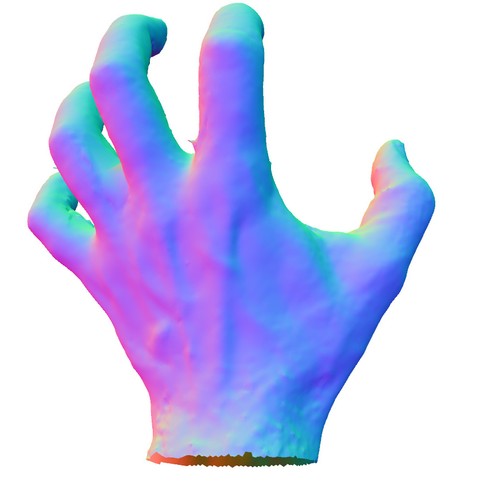}};
	\node (p62) at(p61.south)[anchor=north] {\includegraphics[width=\imw]{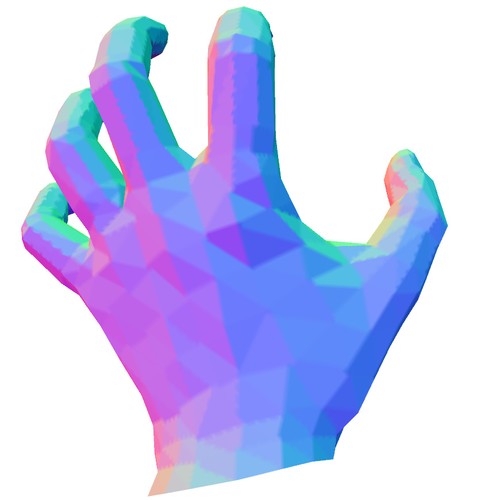}};
	\node (p63) at(p62.south)[anchor=north] {\includegraphics[width=\imw]{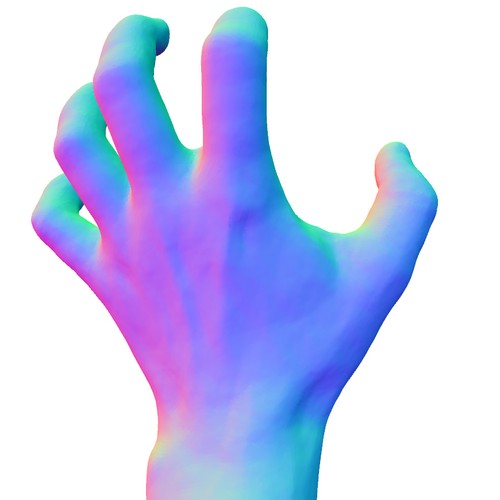}};

	\node (p70) at(p60.east)[anchor=west,xshift=-1.2ex]  {\includegraphics[width=\imw]{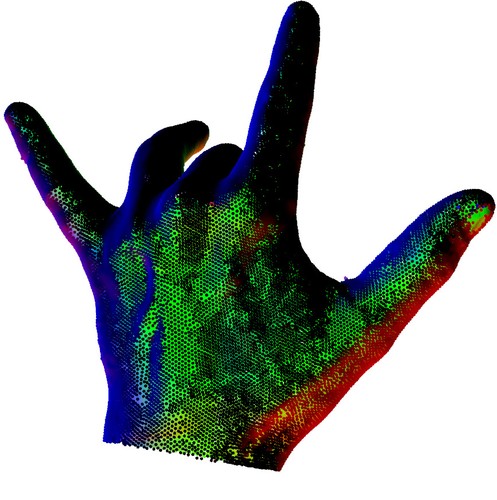}};
	\node (p71) at(p70.south)[anchor=north] {\includegraphics[width=\imw]{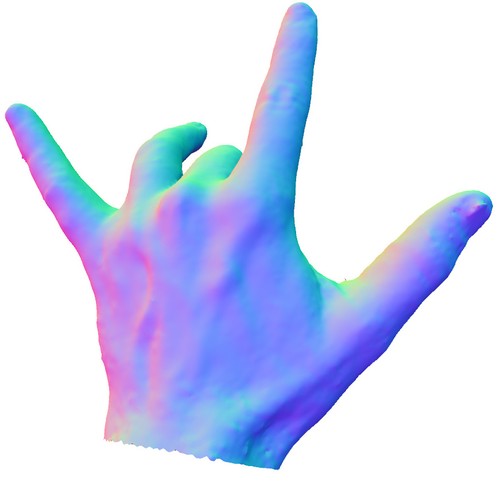}};
	\node (p72) at(p71.south)[anchor=north] {\includegraphics[width=\imw]{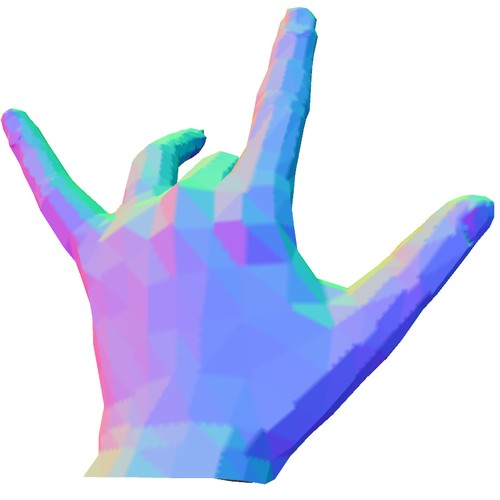}};
	\node (p73) at(p72.south)[anchor=north] {\includegraphics[width=\imw]{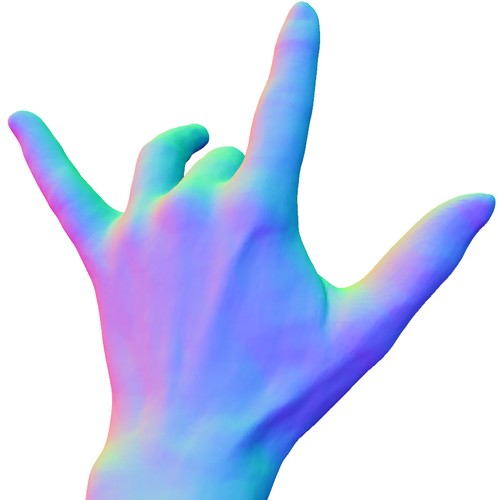}};

    \node (p50) at(p70.east)[anchor=west,xshift=-0.3ex]  {\includegraphics[width=\imw]{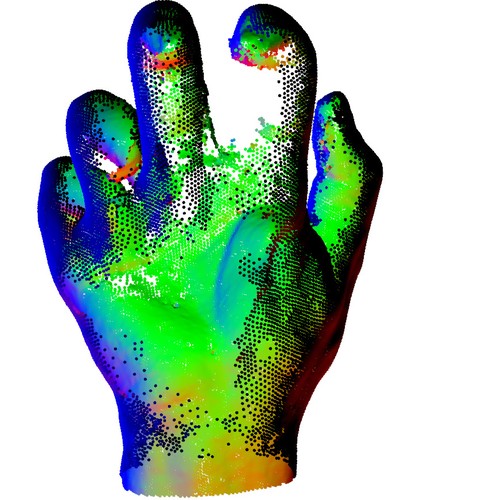}};
    \node (p51) at(p50.south)[anchor=north] {\includegraphics[width=\imw]{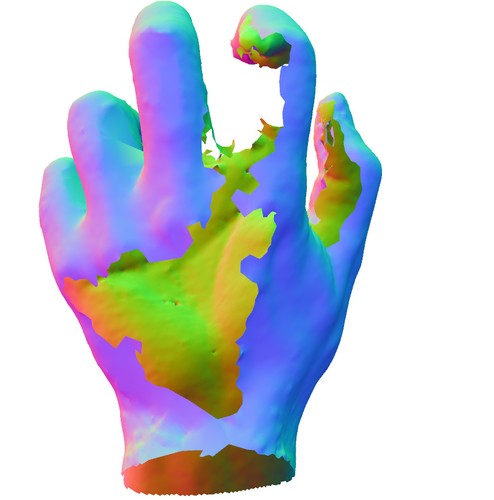}};
    \node (p52) at(p51.south)[anchor=north] {\includegraphics[width=\imw]{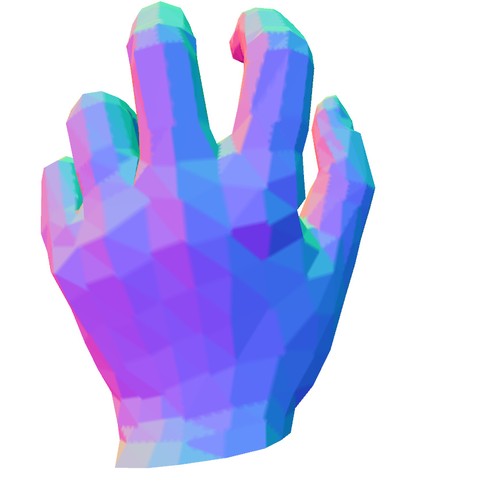}};
    \node (p53) at(p52.south)[anchor=north] {\includegraphics[width=\imw]{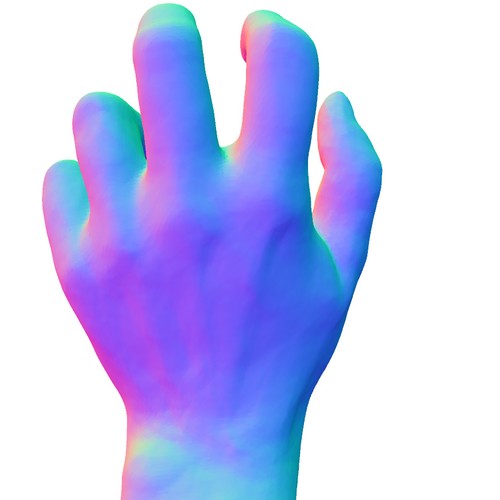}};

    \node at(p00.west)[anchor=south, xshift=1ex, rotate=90, font=\scriptsize] {point cloud};
    \node at(p01.west)[anchor=south, xshift=1ex, rotate=90, font=\scriptsize] {ground truth};
    \node at(p02.west)[anchor=south, xshift=1ex, rotate=90, font=\scriptsize] {MANO};
    \node at(p03.west)[anchor=south, xshift=1ex, rotate=90, font=\scriptsize] {\textbf{DiForm (ours)}};

    \end{tikzpicture}
    \caption{The level of details of the proposed DiForm versus MANO~\cite{Romero:etal:TOG2017}. %
    	The ground truth mesh obtained by the scanning device is shown for reference. DiForm is able to express more accurate and detailed muscle deformation, such as creases and bulging, and fill in the holes that are missing from the inputs.}
    \label{fig:mano-dyform}
\end{figure}

%% file: supp/fig_3dmd.tex
\begin{figure}
    \centering
    \begin{tikzpicture}[inner sep=0pt]
    \def\imw{9.5ex}
    \def\yoff{3pt}

    \node (px0) {\includegraphics[width=\imw]{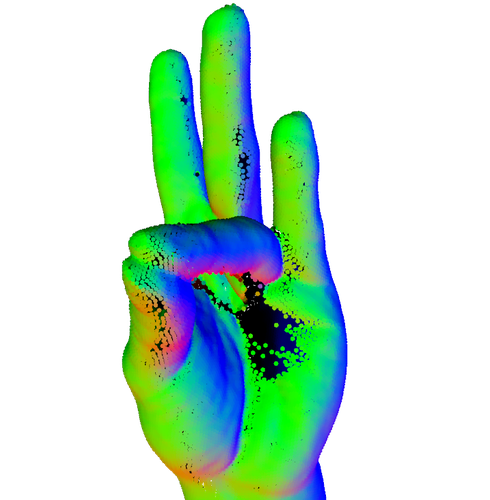}};
    \node (px1) at(px0.south) [anchor=north, yshift=-\yoff] {\includegraphics[width=\imw]{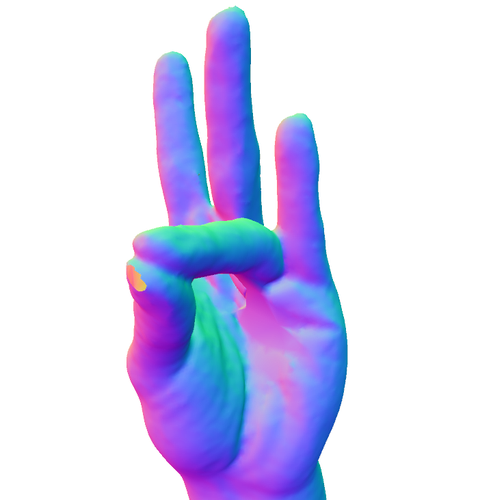}};
    \node (px2) at(px1.south) [anchor=north, yshift=-\yoff] {\includegraphics[width=\imw]{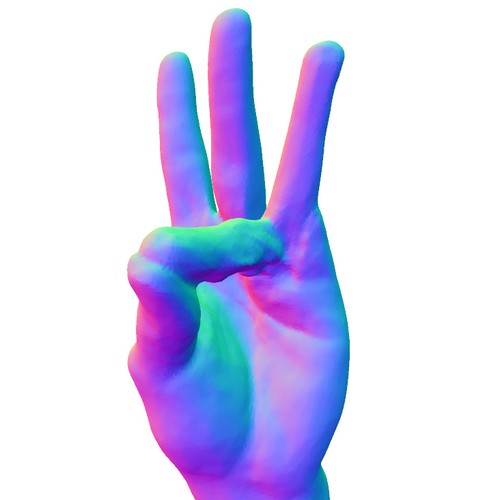}};
    \node (px3) at(px2.south) [anchor=north, yshift=-\yoff] {\includegraphics[width=\imw]{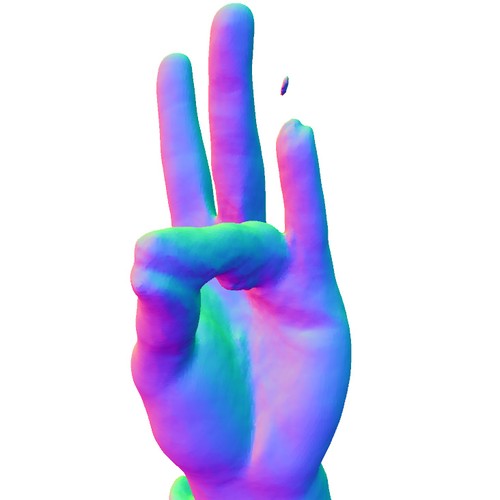}};
    \node (px4) at(px3.south) [anchor=north, yshift=-\yoff] {\includegraphics[width=\imw]{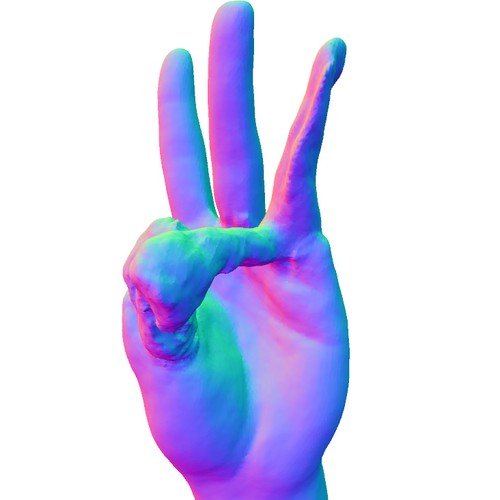}};
    \node (px5) at(px4.south) [anchor=north, yshift=-\yoff] {\includegraphics[width=\imw]{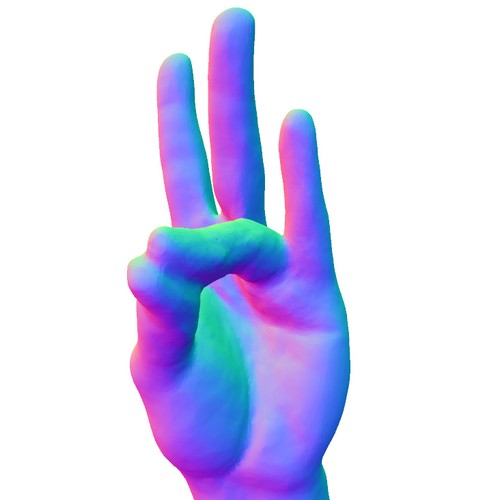}};

    \node (p00) at(px0.east) [anchor=west, xshift=-3ex]{\includegraphics[width=\imw]{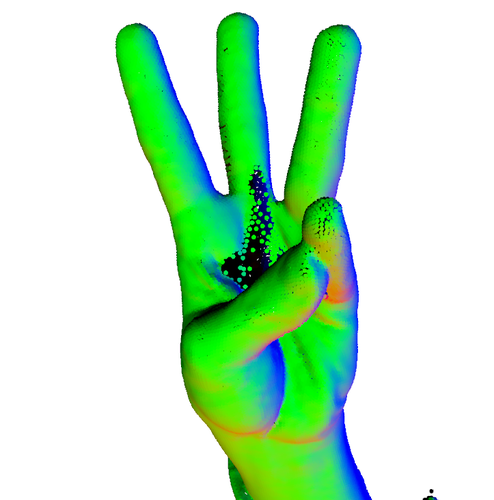}};
    \node (p01) at(p00.south) [anchor=north, yshift=-\yoff] {\includegraphics[width=\imw]{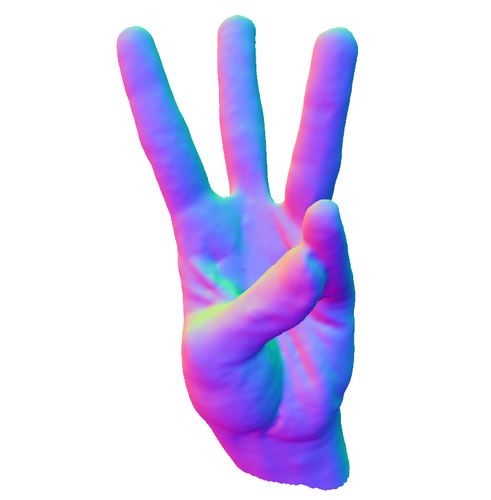}};
    \node (p02) at(p01.south) [anchor=north, yshift=-\yoff] {\includegraphics[width=\imw]{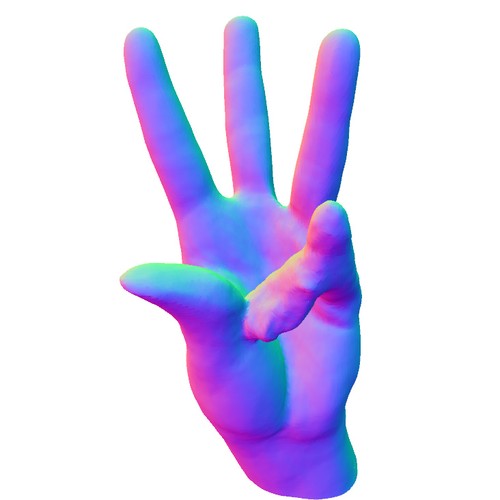}};
    \node (p03) at(p02.south) [anchor=north, yshift=-\yoff] {\includegraphics[width=\imw]{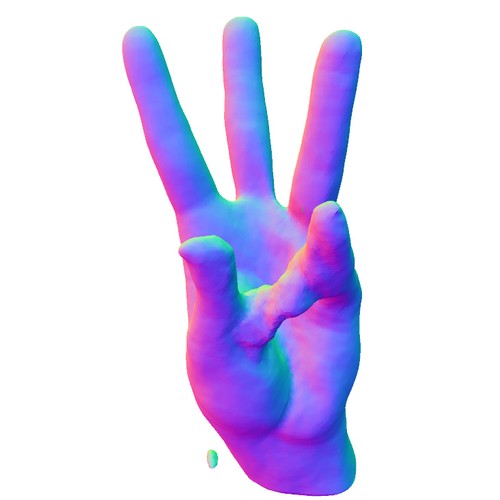}};
    \node (p04) at(p03.south) [anchor=north, yshift=-\yoff] {\includegraphics[width=\imw]{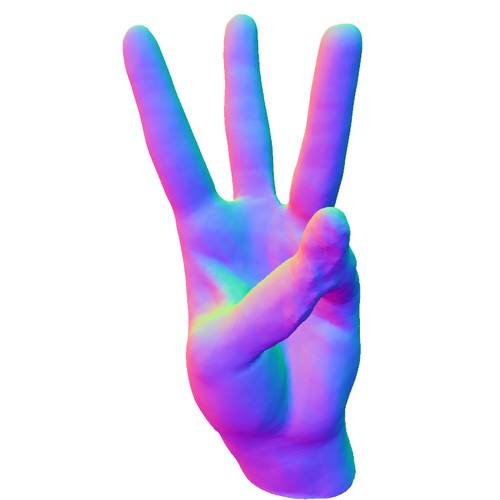}};
    \node (p05) at(p04.south) [anchor=north, yshift=-\yoff] {\includegraphics[width=\imw]{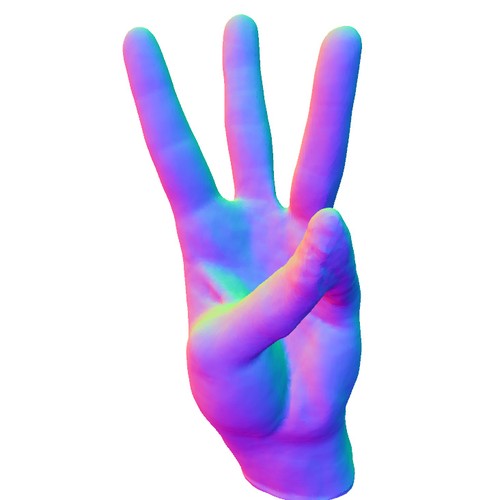}};

    \node (p10) at(p00.east)[anchor=west, xshift=-1ex] {\includegraphics[width=\imw]{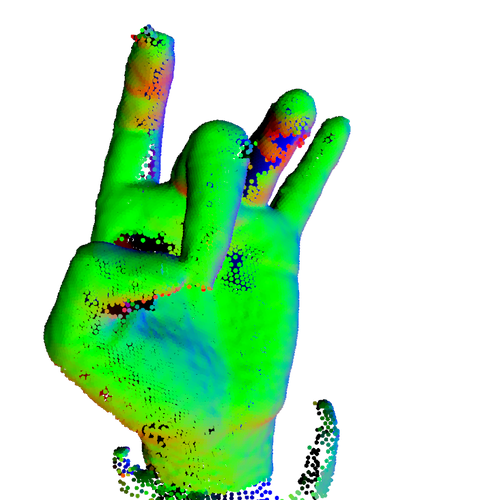}};
    \node (p11) at(p10.south) [anchor=north, yshift=-\yoff] {\includegraphics[width=\imw]{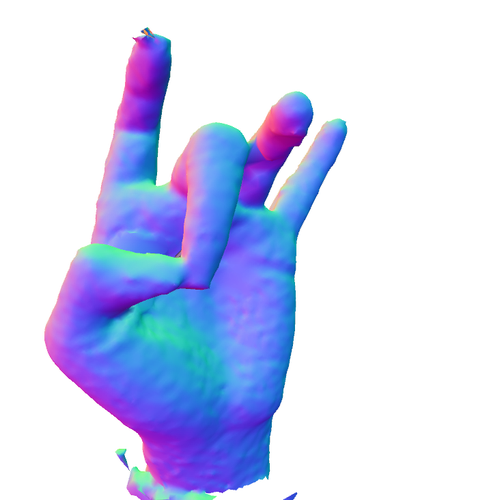}};
    \node (p12) at(p11.south) [anchor=north, yshift=-\yoff] {\includegraphics[width=\imw]{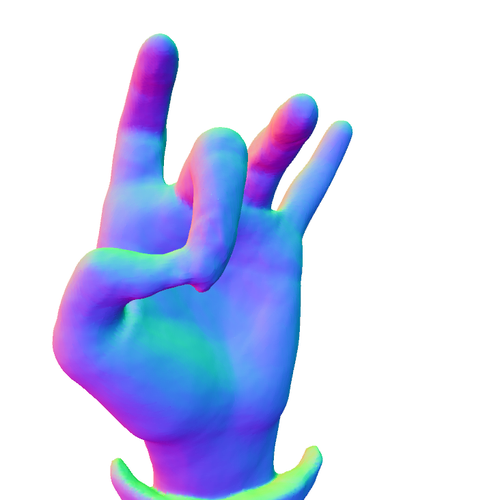}};
    \node (p13) at(p12.south) [anchor=north, yshift=-\yoff] {\includegraphics[width=\imw]{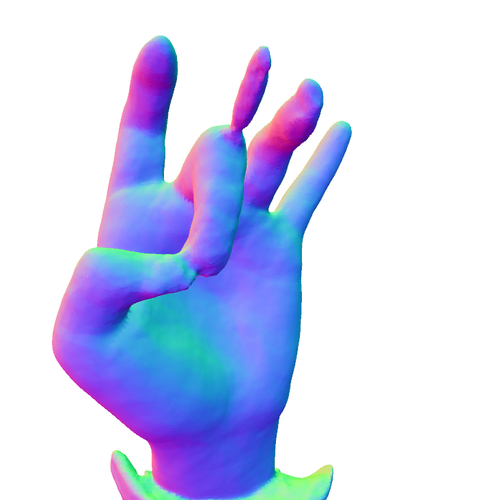}};
    \node (p14) at(p13.south) [anchor=north, yshift=-\yoff] {\includegraphics[width=\imw]{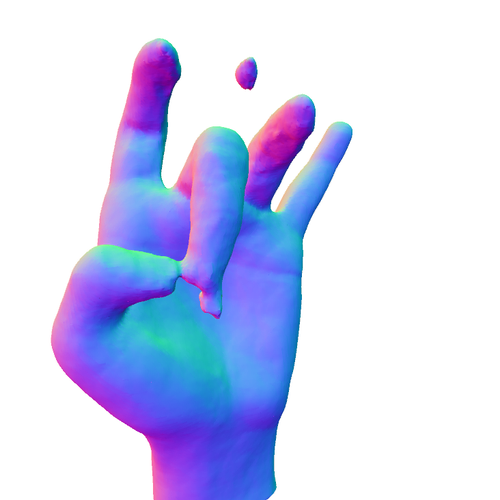}};
    \node (p15) at(p14.south) [anchor=north, yshift=-\yoff] {\includegraphics[width=\imw]{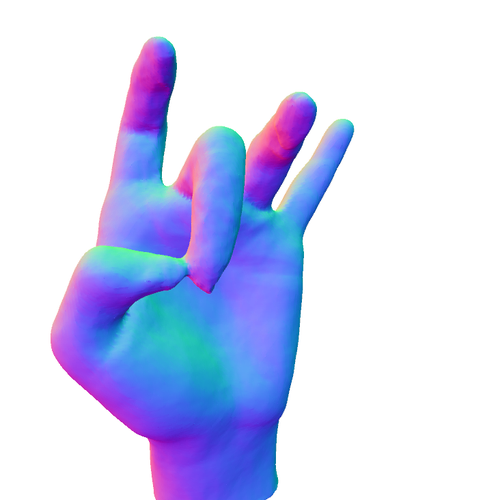}};

    \node (p30) at(p10.east)  [anchor=west, xshift=-3ex]  {\includegraphics[width=\imw]{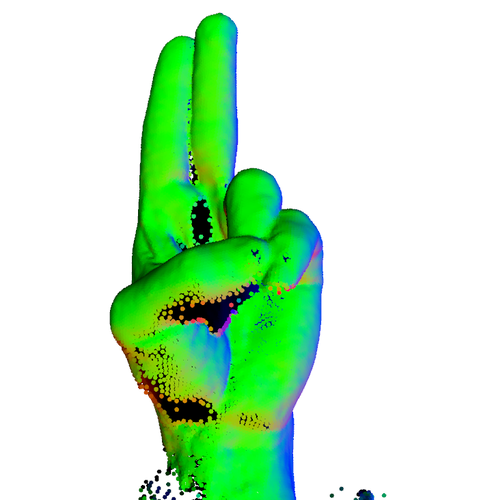}};
    \node (p31) at(p30.south) [anchor=north, yshift=-\yoff] {\includegraphics[width=\imw]{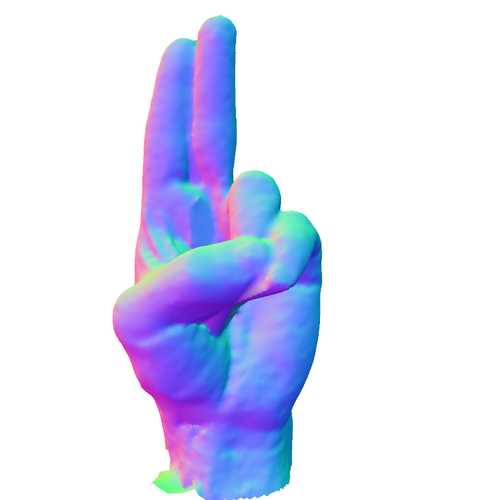}};
    \node (p32) at(p31.south) [anchor=north, yshift=-\yoff] {\includegraphics[width=\imw]{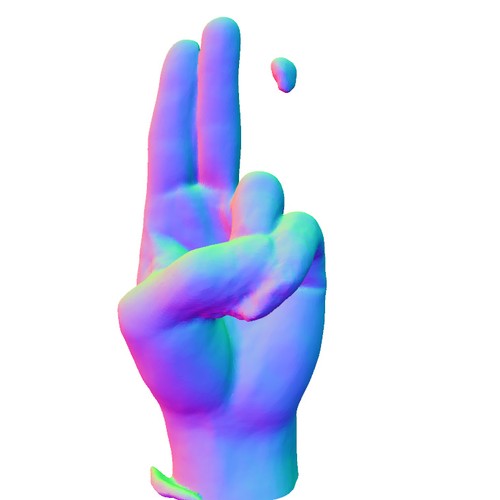}};
    \node (p33) at(p32.south) [anchor=north, yshift=-\yoff] {\includegraphics[width=\imw]{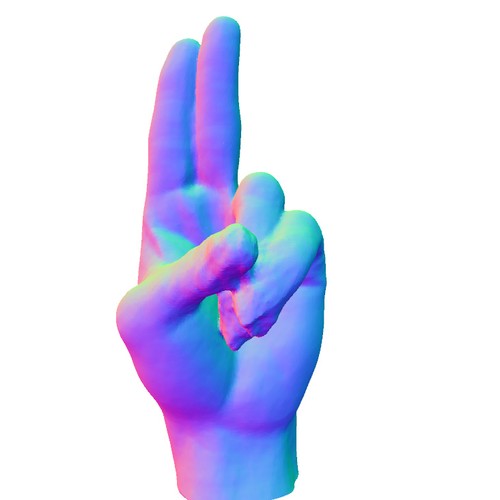}};
    \node (p34) at(p33.south) [anchor=north, yshift=-\yoff] {\includegraphics[width=\imw]{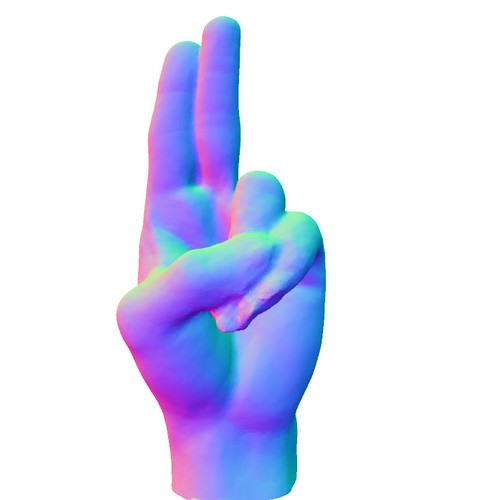}};
    \node (p35) at(p34.south) [anchor=north, yshift=-\yoff] {\includegraphics[width=\imw]{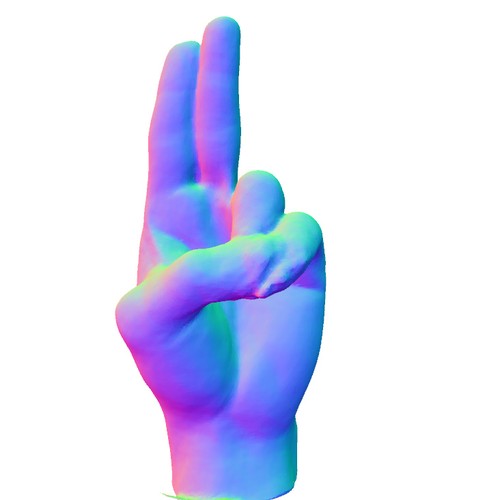}};

    \node (p40) at(p30.east)  [anchor=west,xshift=-3ex]  {\includegraphics[width=\imw]{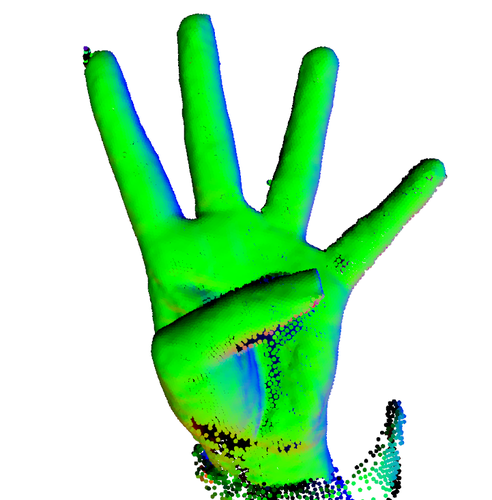}};
    \node (p41) at(p40.south) [anchor=north, yshift=-\yoff] {\includegraphics[width=\imw]{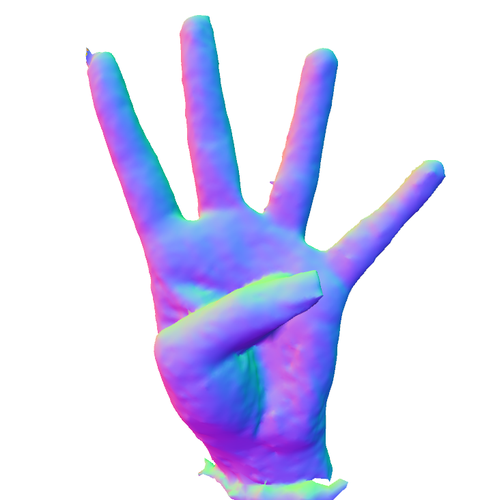}};
    \node (p42) at(p41.south) [anchor=north, yshift=-\yoff] {\includegraphics[width=\imw]{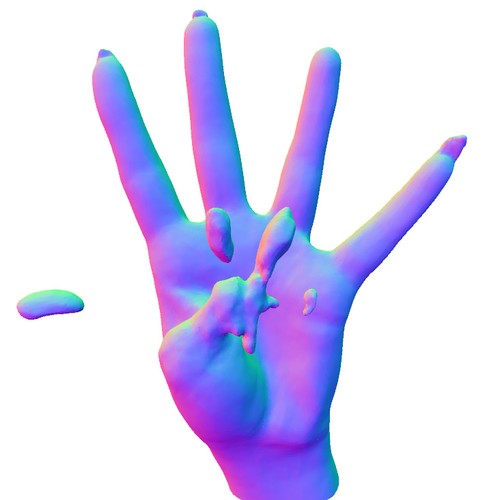}};
    \node (p43) at(p42.south) [anchor=north, yshift=-\yoff] {\includegraphics[width=\imw]{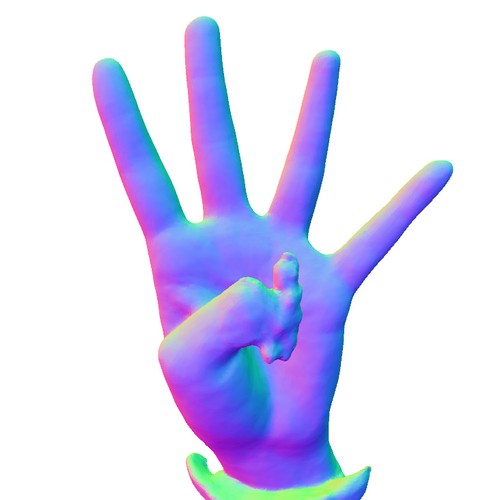}};
    \node (p44) at(p43.south) [anchor=north, yshift=-\yoff] {\includegraphics[width=\imw]{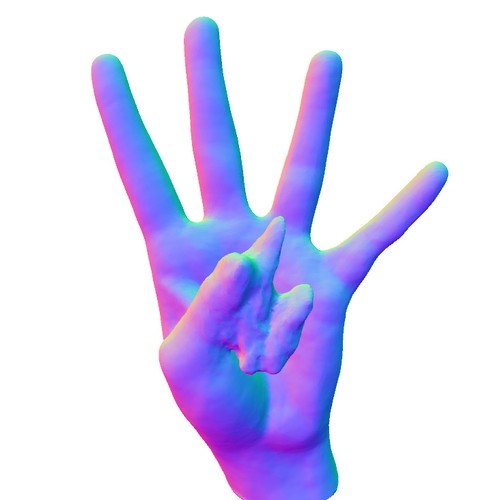}};
    \node (p45) at(p44.south) [anchor=north, yshift=-\yoff] {\includegraphics[width=\imw]{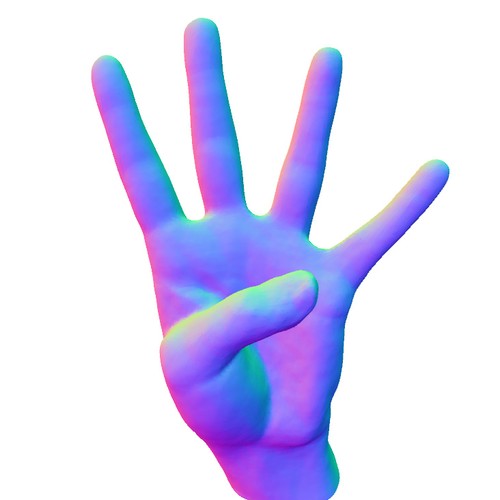}};

    \node (p60) at(p40.east)  [anchor=west,xshift=-1ex]  {\includegraphics[width=\imw]{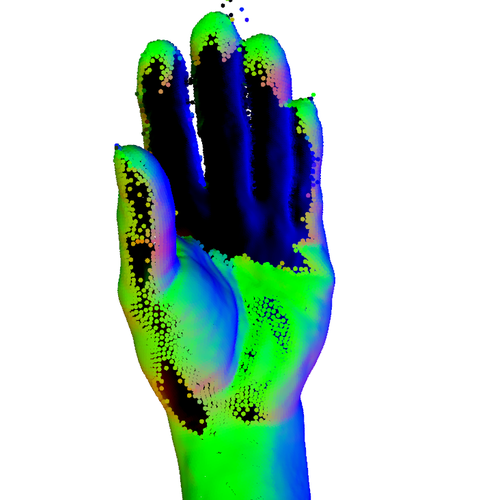}};
    \node (p61) at(p60.south) [anchor=north, yshift=-\yoff] {\includegraphics[width=\imw]{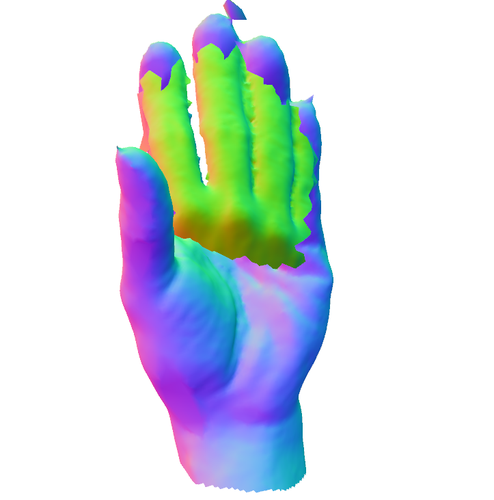}};
    \node (p62) at(p61.south) [anchor=north, yshift=-\yoff] {\includegraphics[width=\imw]{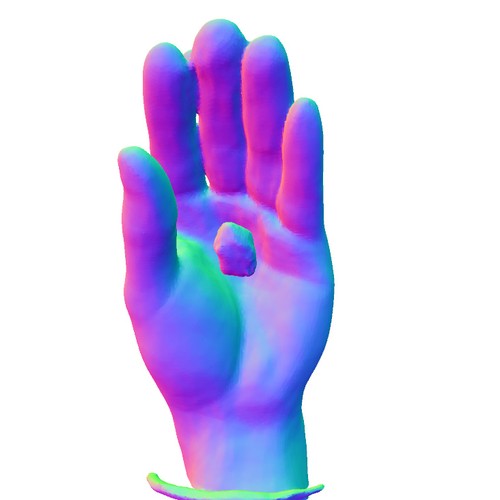}};
    \node (p63) at(p62.south) [anchor=north, yshift=-\yoff] {\includegraphics[width=\imw]{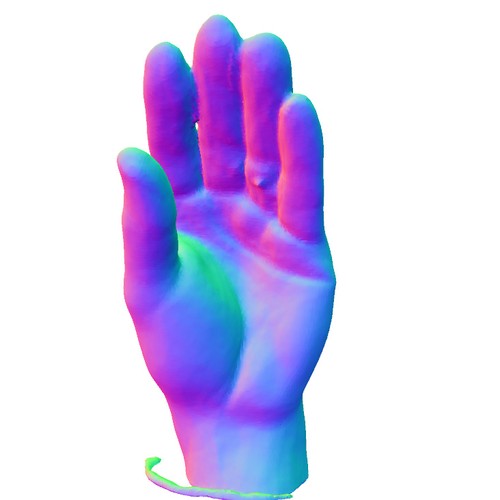}};
    \node (p64) at(p63.south) [anchor=north, yshift=-\yoff] {\includegraphics[width=\imw]{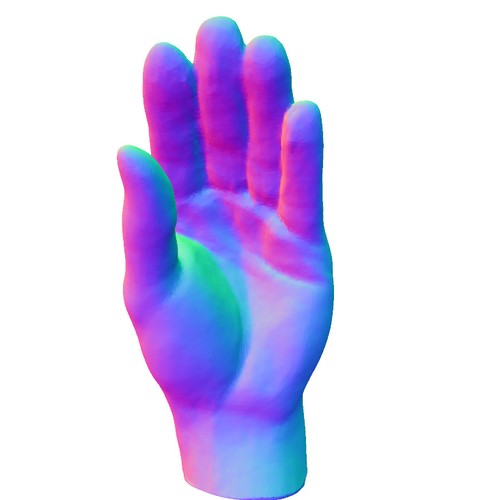}};
    \node (p65) at(p64.south) [anchor=north, yshift=-\yoff] {\includegraphics[width=\imw]{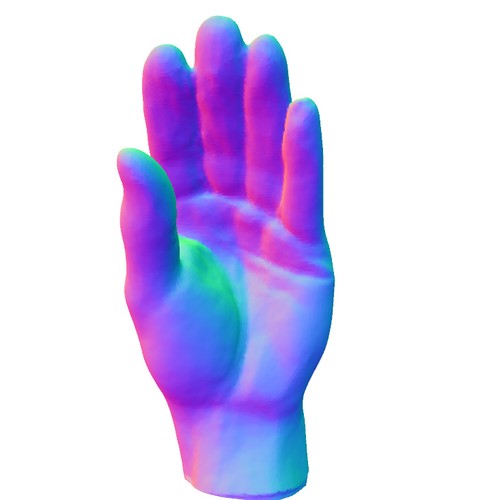}};

    \node (p70) at(p60.east)  [anchor=west,xshift=-2.ex]  {\includegraphics[width=\imw]{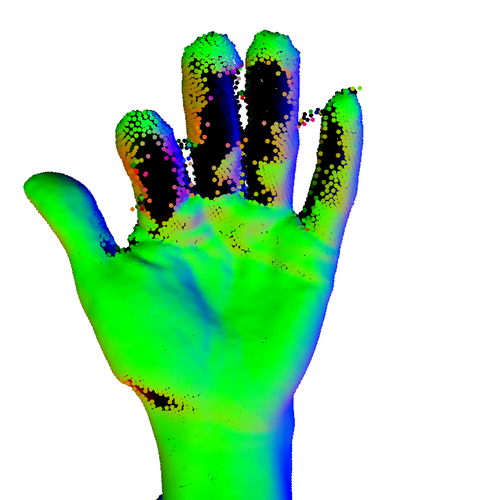}};
    \node (p71) at(p70.south) [anchor=north, yshift=-\yoff] {\includegraphics[width=\imw]{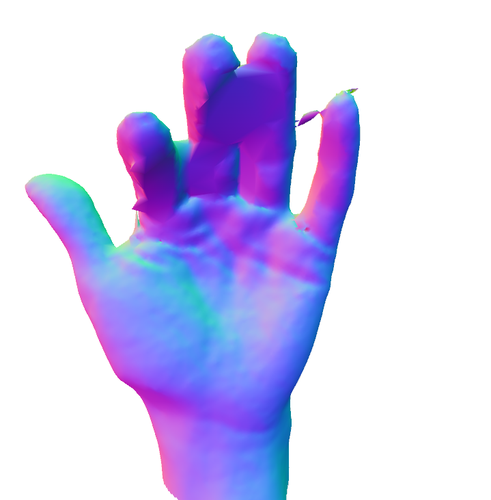}};
    \node (p72) at(p71.south) [anchor=north, yshift=-\yoff] {\includegraphics[width=\imw]{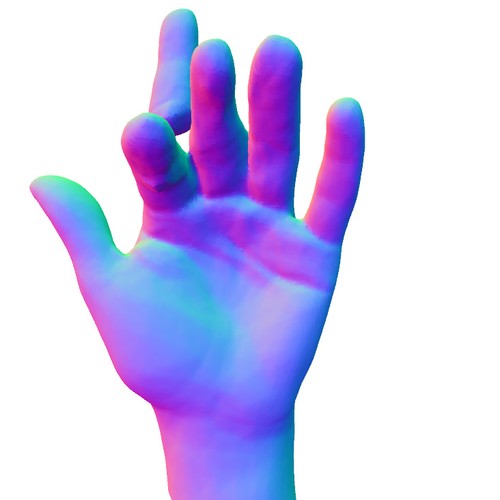}};
    \node (p73) at(p72.south) [anchor=north, yshift=-\yoff] {\includegraphics[width=\imw]{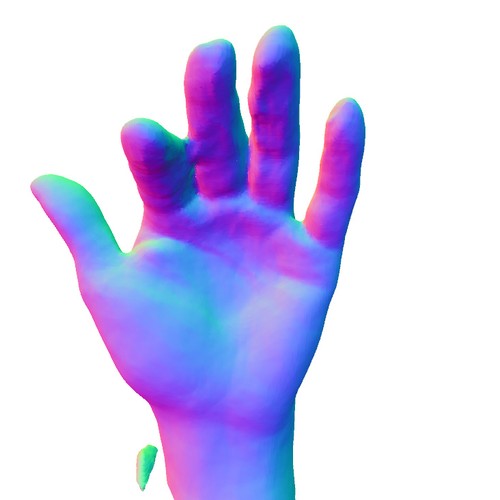}};
    \node (p74) at(p73.south) [anchor=north, yshift=-\yoff] {\includegraphics[width=\imw]{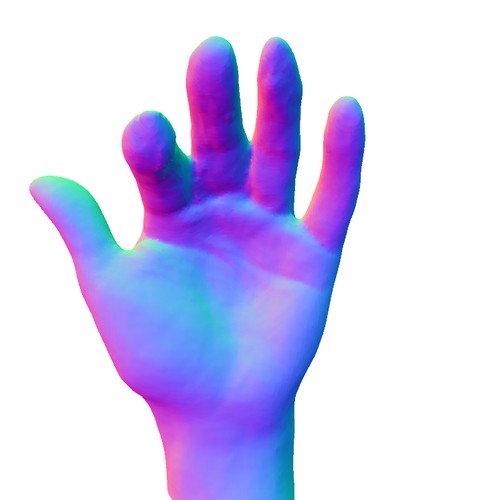}};
    \node (p75) at(p74.south) [anchor=north, yshift=-\yoff] {\includegraphics[width=\imw]{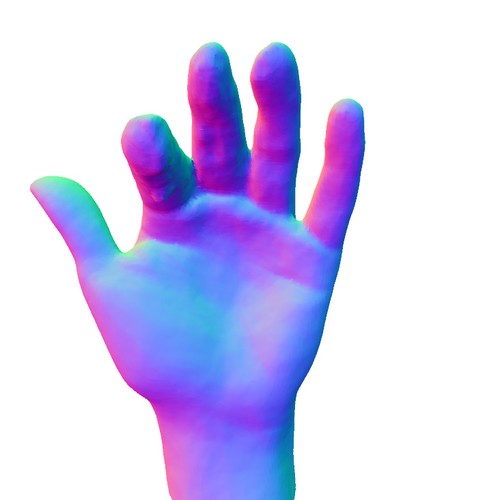}};

    \node at(px0.west)[anchor=south, xshift=1ex, rotate=90, font=\scriptsize] {point cloud};
    \node at(px1.west)[anchor=south, xshift=1ex, rotate=90, font=\scriptsize] {ground truth};
    \node at(px2.west)[anchor=south, xshift=1ex, rotate=90, font=\scriptsize] {IGR};
    \node at(px3.west)[anchor=south, xshift=1ex, rotate=90, font=\scriptsize] {IGR-PE};
    \node at(px4.west)[anchor=south, xshift=1ex, rotate=90, font=\scriptsize][align=center] {\textbf{DiForm-S}};
    \node at(px5.west)[anchor=south, xshift=1ex, rotate=90, font=\scriptsize][align=center] {\textbf{DiForm-J}};

    \end{tikzpicture}
    \caption{Qualitative comparison on the \dataset test set. All methods use point clouds for reconstruction. The ground truth meshes only serve as a visual reference. }
    \vspace{-1em}
    \label{fig:3dmd}
    \label{fig:reconstruct_mano}
\end{figure}

%% file: section/exp-4dscan.tex
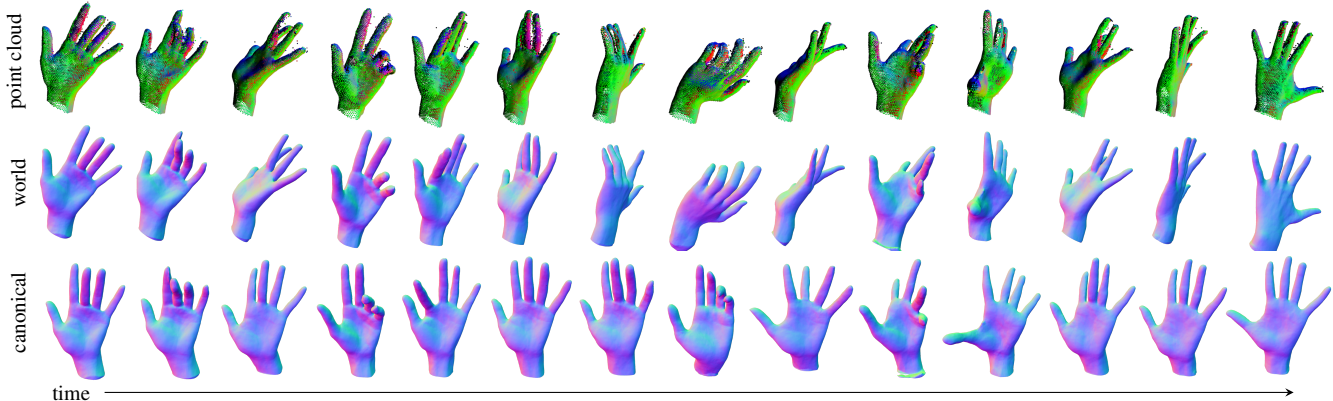
\begin{figure*}
	\def\iw{5.15em}	
	\def\yoff{6pt}	
	\centering
	\begin{tikzpicture}[inner sep=0pt]
		\node(p0)at (0,0)    [anchor=west]{\includegraphics[width=\iw]{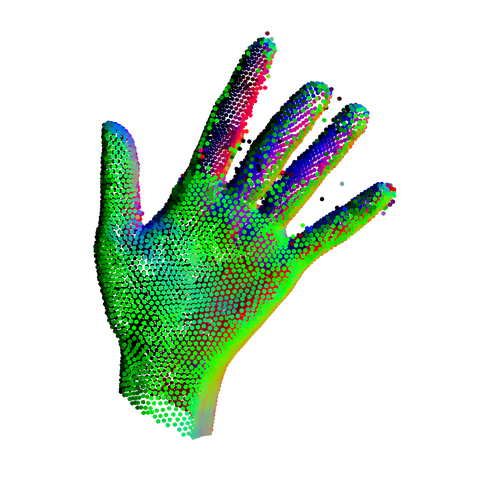}};
		\node(p1)at (p0.east)[anchor=west,xshift=-1.7em]{\includegraphics[width=\iw]{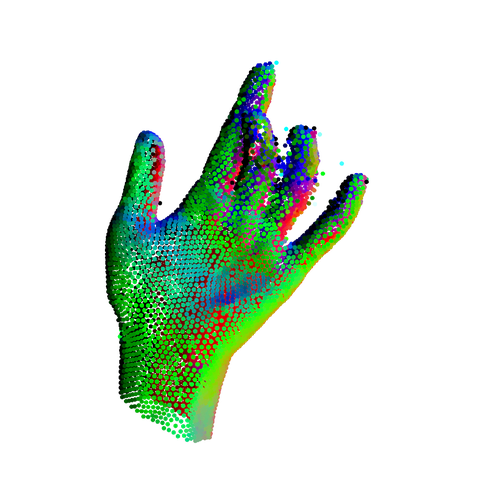}};
		\node(p2)at (p1.east)[anchor=west,xshift=-2em]{\includegraphics[width=\iw]{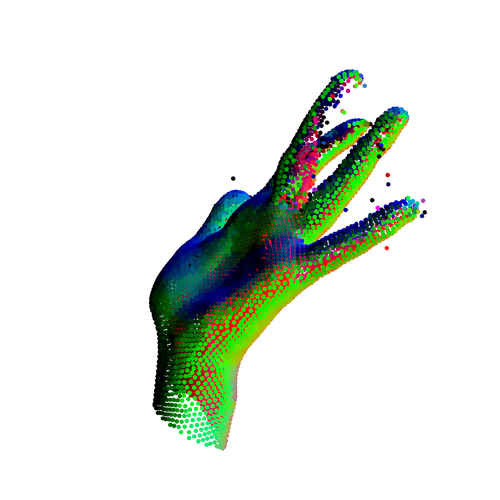}};
		\node(p3)at (p2.east)[anchor=west,xshift=-1.5em]{\includegraphics[width=\iw]{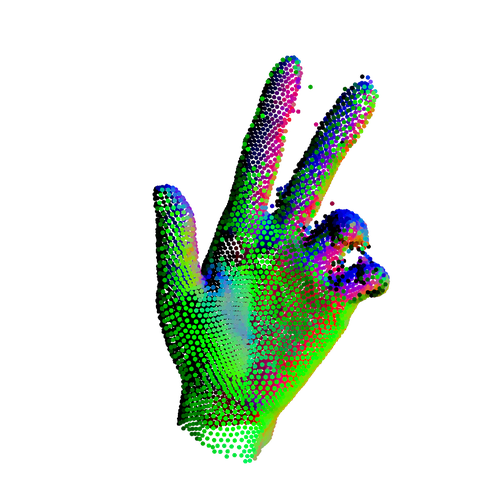}};
		\node(p4)at (p3.east)[anchor=west,xshift=-2.07em]{\includegraphics[width=\iw]{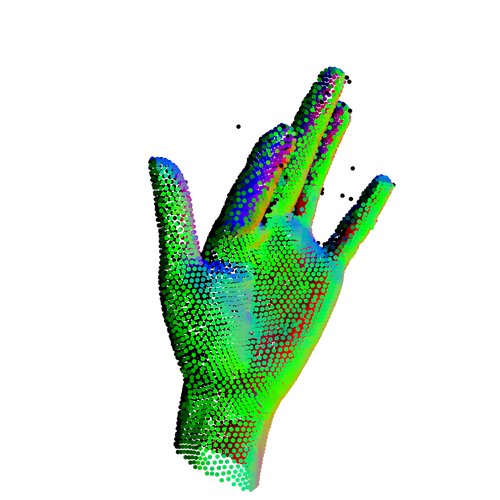}};
		\node(p43)at (p4.east)[anchor=west,xshift=-1.9em]{\includegraphics[width=\iw]{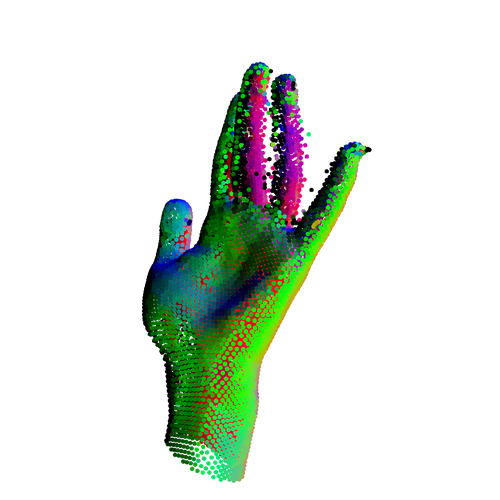}};
		\node(p45)at (p43.east)[anchor=west,xshift=-1.8em]{\includegraphics[width=\iw]{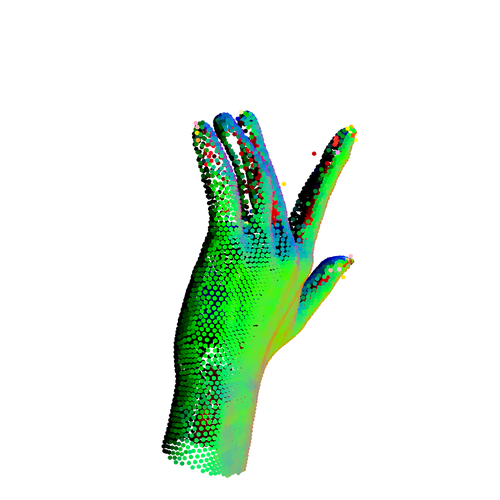}};
		\node(p47)at (p45.east)[anchor=west,xshift=-2em]{\includegraphics[width=\iw]{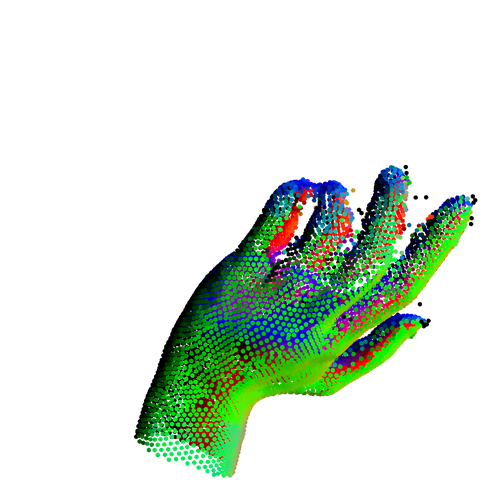}};
		\node(p5)at (p47.east)[anchor=west,xshift=-1.4em]{\includegraphics[width=\iw]{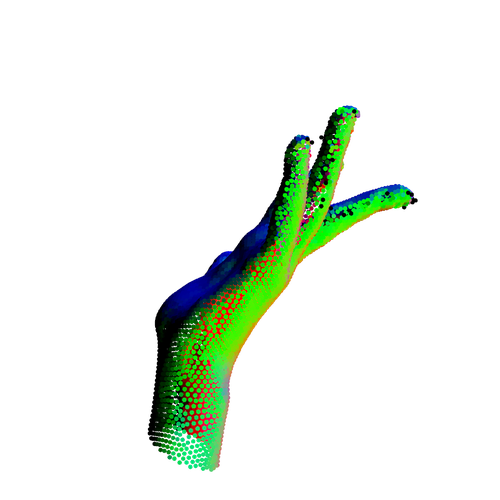}};
		\node(p6)at (p5.east)[anchor=west,xshift=-1.4em]{\includegraphics[width=\iw]{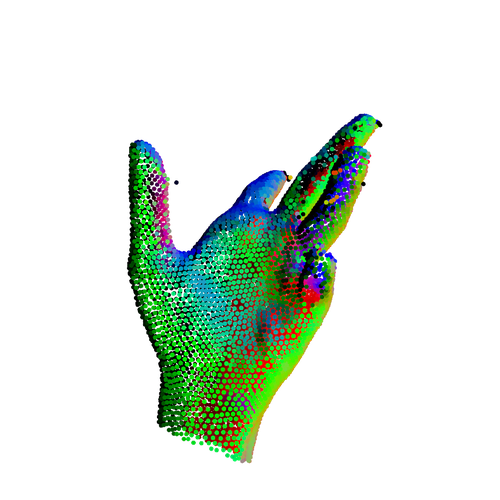}};
		\node(p7)at (p6.east)[anchor=west,xshift=-1.5em]{\includegraphics[width=\iw]{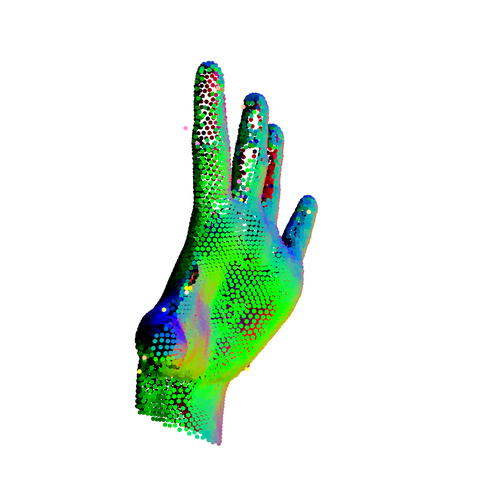}};
		\node(p8)at (p7.east)[anchor=west,xshift=-1.8em]{\includegraphics[width=\iw]{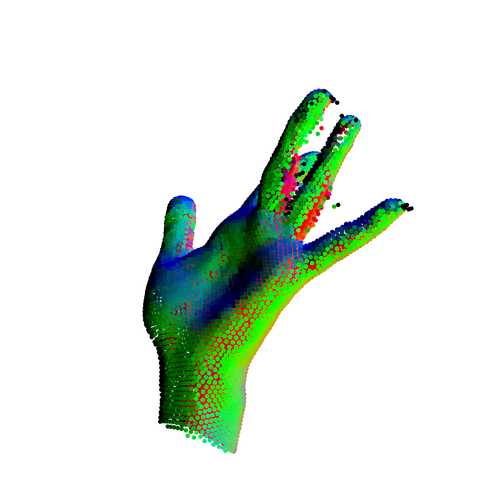}};
		\node(p85)at (p8.east)[anchor=west,xshift=-1.8em]{\includegraphics[width=\iw]{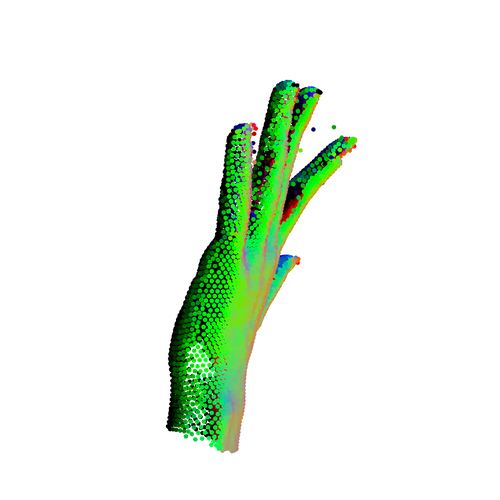}};
		\node(p9)at (p85.east)[anchor=west,xshift=-1.4em]{\includegraphics[width=\iw]{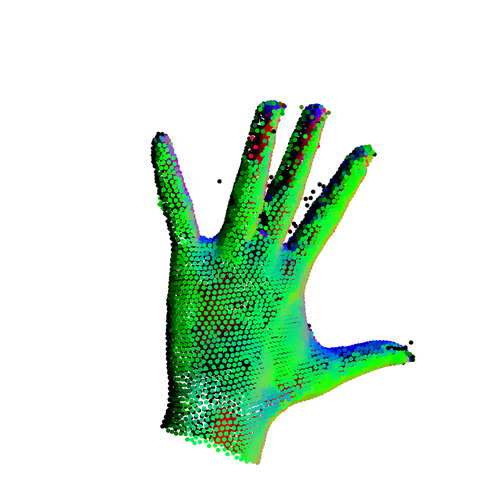}};
		
		\node(r0)at (p0.south)[anchor=north, yshift=\yoff]{\includegraphics[width=\iw]{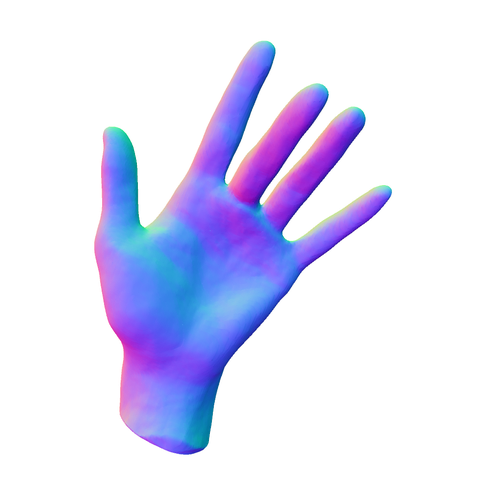}};
		\node(r1)at (p1.south)[anchor=north, yshift=\yoff]{\includegraphics[width=\iw]{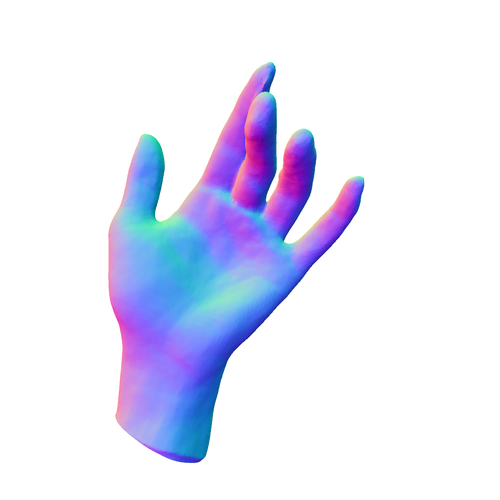}};
		\node(r2)at (p2.south)[anchor=north, yshift=\yoff]{\includegraphics[width=\iw]{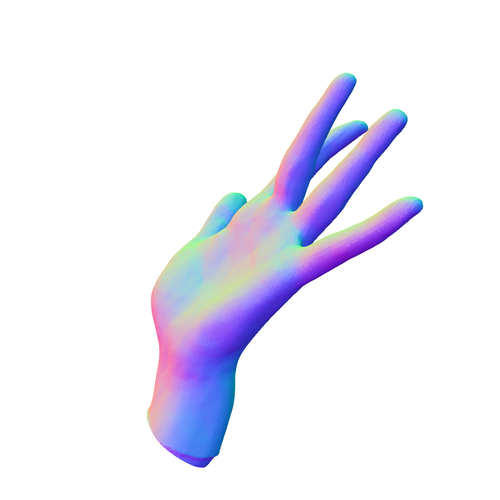}};
		\node(r3)at (p3.south)[anchor=north, yshift=\yoff]{\includegraphics[width=\iw]{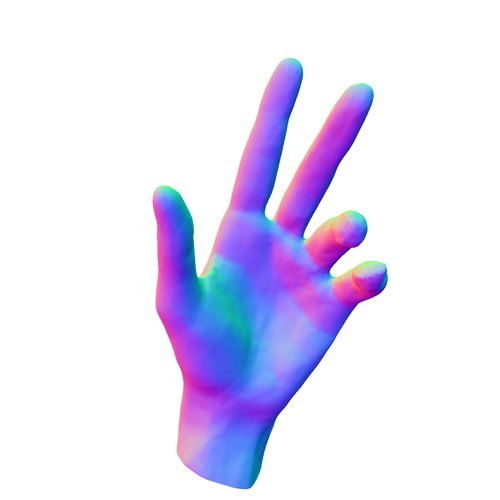}};
		\node(r4)at (p4.south)[anchor=north, yshift=\yoff]{\includegraphics[width=\iw]{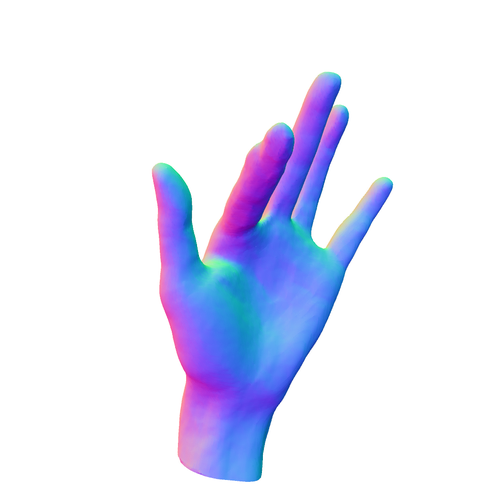}};
		\node(r43)at (p43.south)[anchor=north, yshift=\yoff]{\includegraphics[width=\iw]{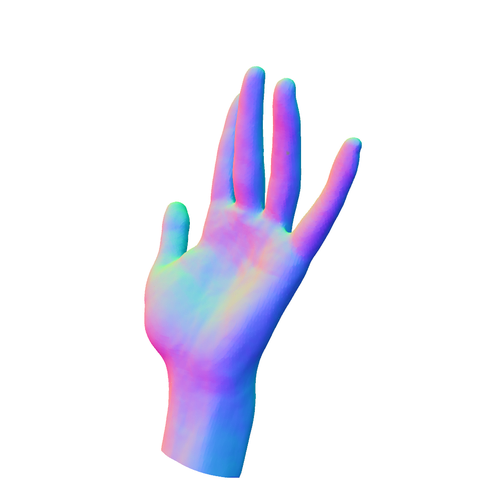}};
		\node(r45)at (p45.south)[anchor=north, yshift=\yoff]{\includegraphics[width=\iw]{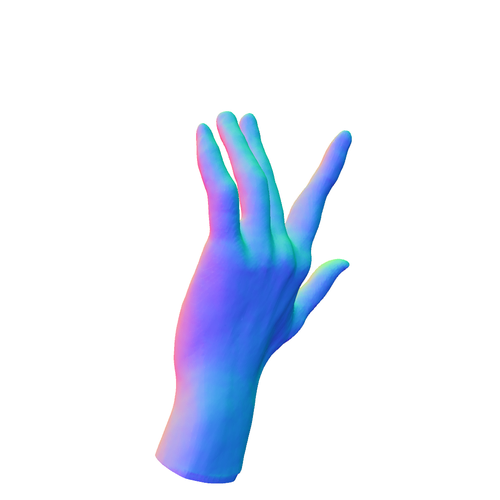}};
		\node(r47)at (p47.south)[anchor=north, yshift=\yoff]{\includegraphics[width=\iw]{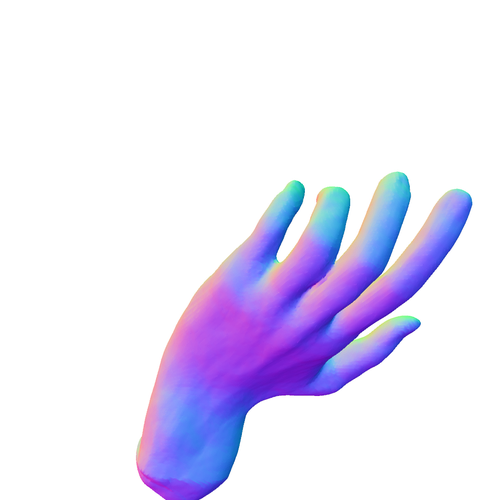}};
		\node(r5)at (p5.south)[anchor=north, yshift=\yoff]{\includegraphics[width=\iw]{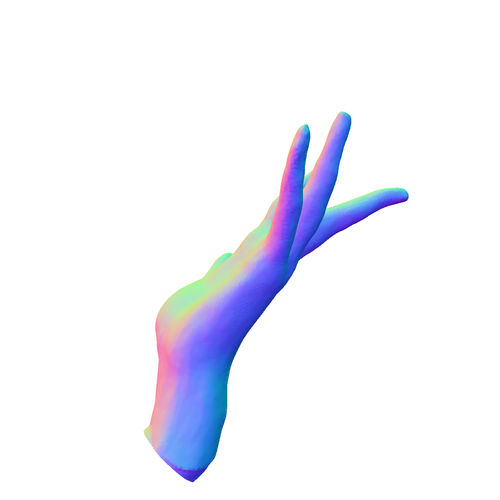}};
		\node(r6)at (p6.south)[anchor=north, yshift=\yoff]{\includegraphics[width=\iw]{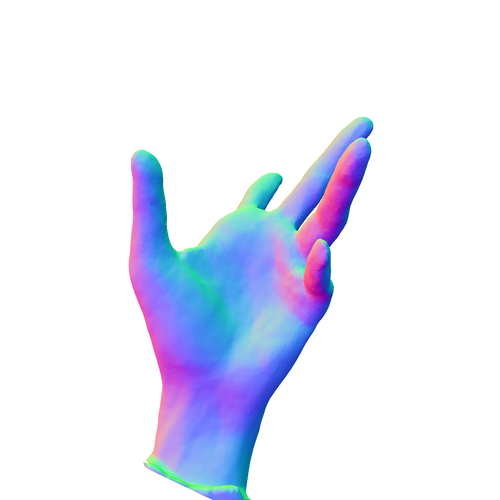}};
		\node(r7)at (p7.south)[anchor=north, yshift=\yoff]{\includegraphics[width=\iw]{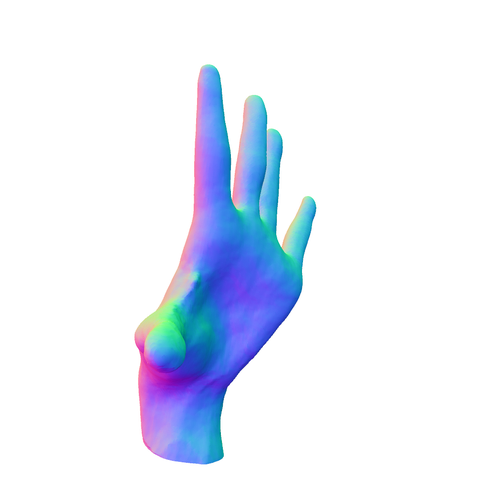}};
		\node(r8)at (p8.south)[anchor=north, yshift=\yoff]{\includegraphics[width=\iw]{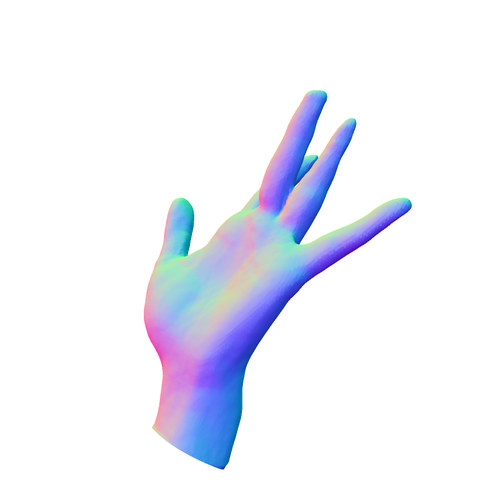}};
		\node(r85)at (p85.south)[anchor=north, yshift=\yoff]{\includegraphics[width=\iw]{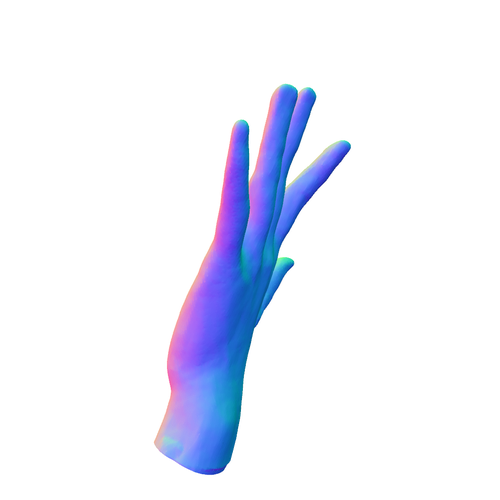}};
		\node(r9)at (p9.south)[anchor=north, yshift=\yoff]{\includegraphics[width=\iw]{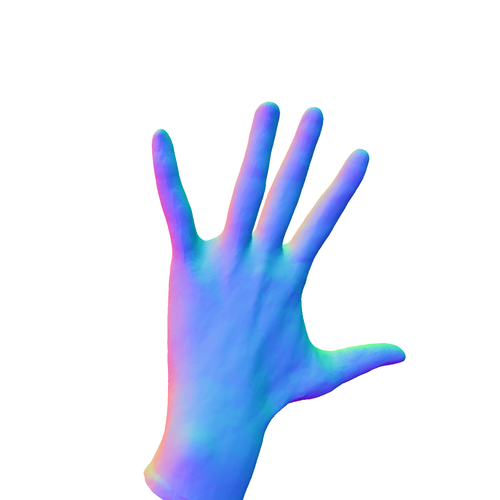}};

		\node(c0)at (r0.south)[anchor=north,   yshift=3pt]{\includegraphics[width=\iw]{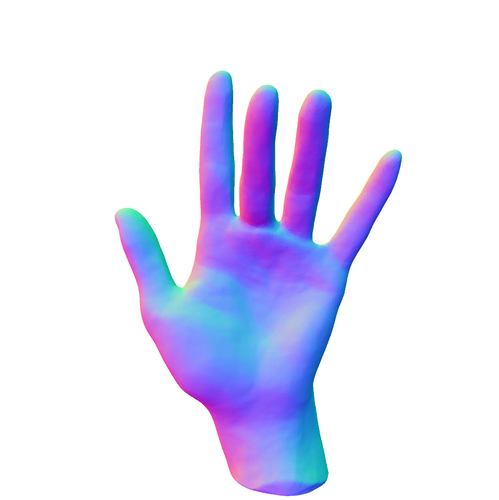}};
		\node(c1)at (r1.south)[anchor=north,   yshift=3pt]{\includegraphics[width=\iw]{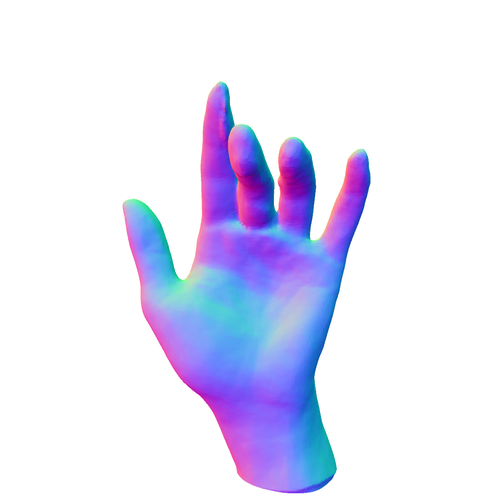}};
		\node(c2)at (r2.south)[anchor=north,   yshift=3pt]{\includegraphics[width=\iw]{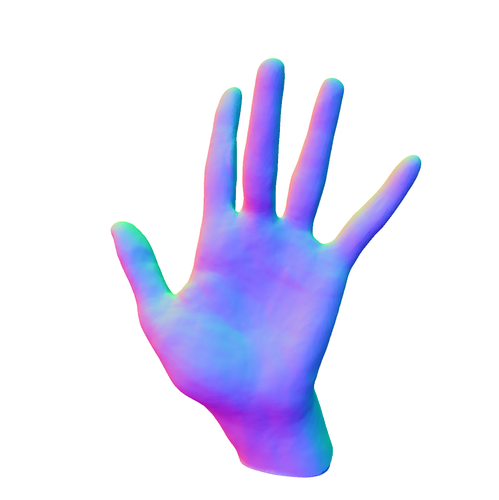}};
		\node(c3)at (r3.south)[anchor=north,   yshift=3pt]{\includegraphics[width=\iw]{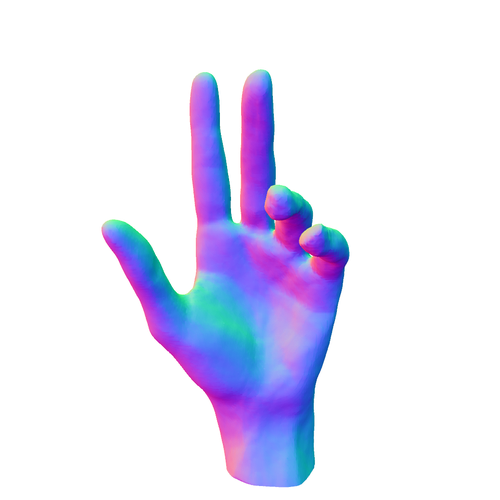}};
		\node(c4)at (r4.south)[anchor=north,   yshift=3pt]{\includegraphics[width=\iw]{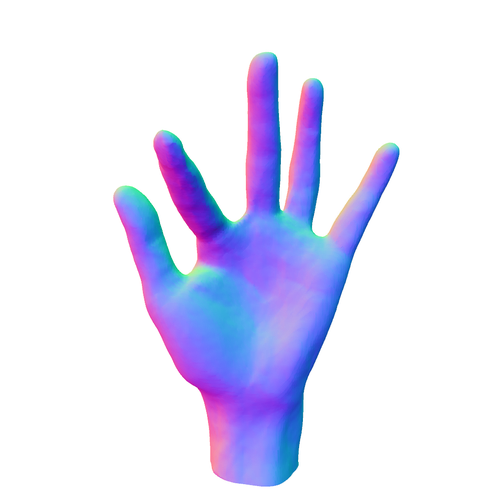}};
		\node(c43)at (r43.south)[anchor=north, yshift=3pt]{\includegraphics[width=\iw]{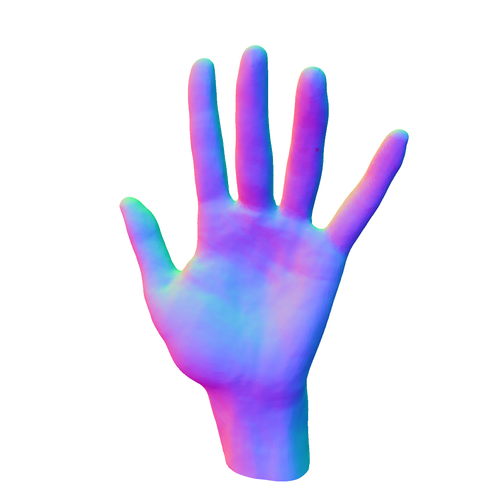}};
		\node(c45)at (r45.south)[anchor=north, yshift=3pt]{\includegraphics[width=\iw]{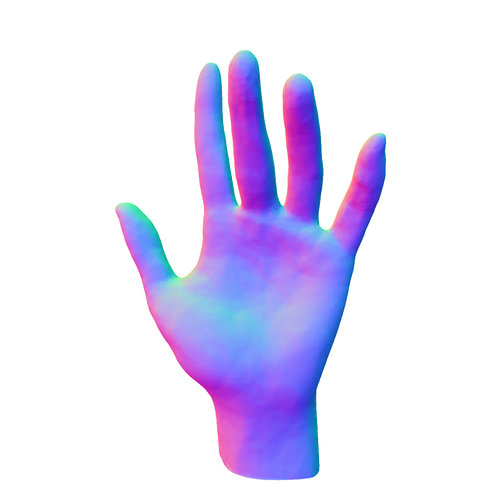}};
		\node(c47)at (r47.south)[anchor=north, yshift=3pt,xshift=0.2em]{\includegraphics[width=\iw]{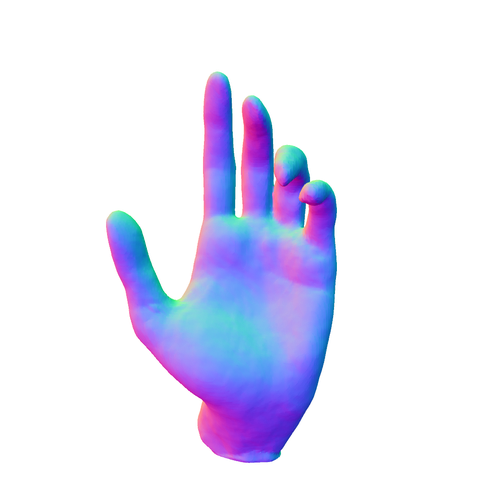}};
		\node(c5)at (r5.south)[anchor=north,   yshift=3pt]{\includegraphics[width=\iw]{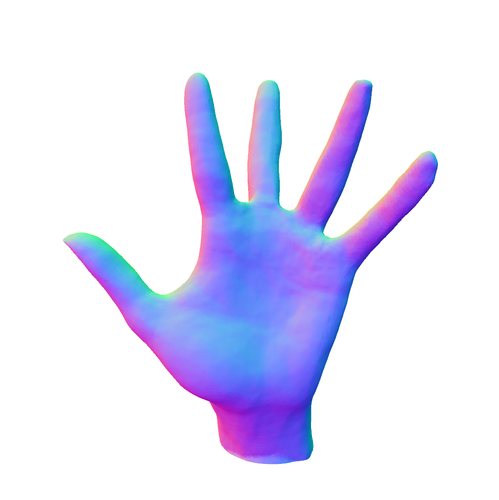}};
		\node(c6)at (r6.south)[anchor=north,   yshift=3pt]{\includegraphics[width=\iw]{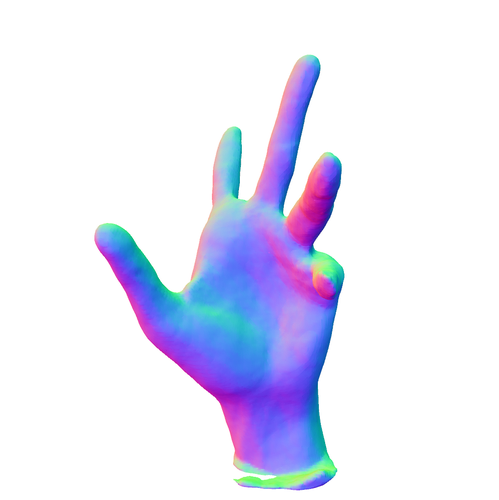}};
		\node(c7)at (r7.south)[anchor=north,   yshift=3pt]{\includegraphics[width=\iw]{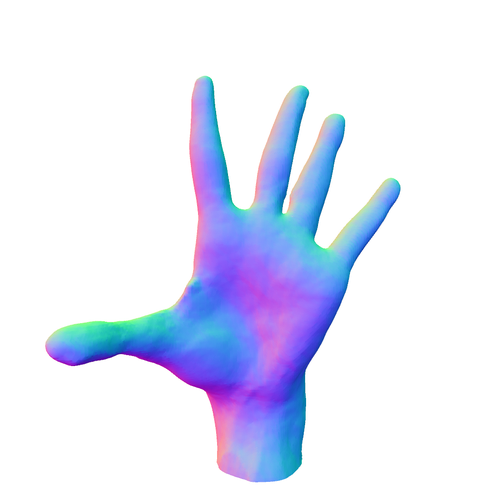}};
		\node(c8)at (r8.south)[anchor=north,   yshift=3pt]{\includegraphics[width=\iw]{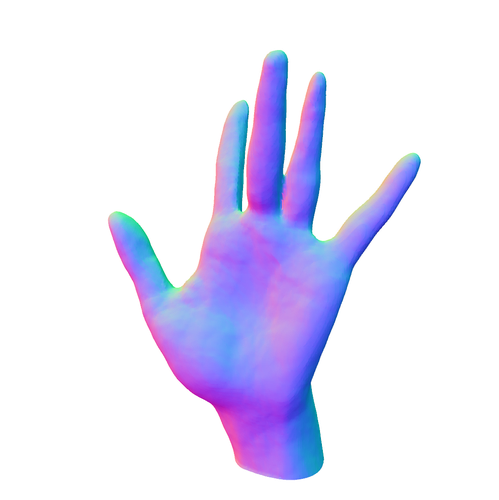}};
		\node(c85)at (r85.south)[anchor=north, yshift=3pt]{\includegraphics[width=\iw]{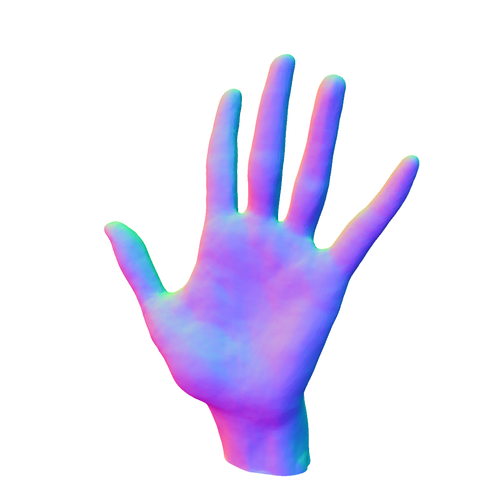}};
		\node(c9)at (r9.south)[anchor=north,   yshift=3pt]{\includegraphics[width=\iw]{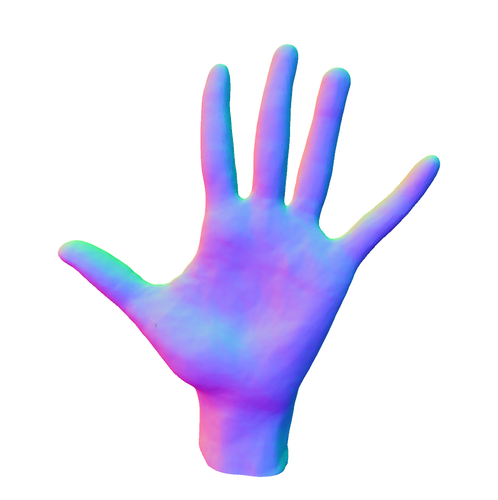}};
		
 		\node at(p0.west)[anchor=south, xshift=1ex, rotate=90, font=\footnotesize] {point cloud};
 		\node at(r0.west)[anchor=south, xshift=1ex, rotate=90, font=\footnotesize] {world};
 		\node at(c0.west)[anchor=south, xshift=1ex, rotate=90, font=\footnotesize] {canonical};

    \node at(0.5,-4.4)[anchor=west, font=\footnotesize] {time};
    \draw[->, >=stealth] [thin] (1.2,-4.4) -- ++(15.8, 0);
	\end{tikzpicture}
  \caption{Dynamic reconstruction of 4D point cloud in the world (top), with DiForm output in the world (mid) and in the canonical (bottom). We direct viewers to the supplementary video for better inspection.}
	\label{fig:4dscan}
\end{figure*}

%% file: supp/fig_failure_case.tex
\begin{figure}
\centering
\def\imw{9.5ex}
\def\yoff{2pt}
\begin{tikzpicture}[inner sep=0pt]
    \node (p00) {\includegraphics[width=\imw]{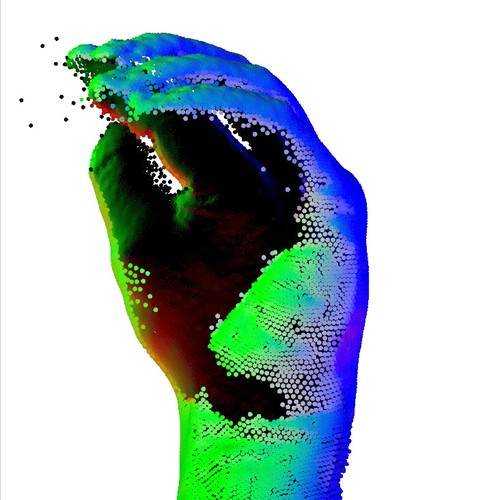}};
    \node (p10) at(p00.east) [anchor=west, xshift=-3ex] {\includegraphics[width=\imw]{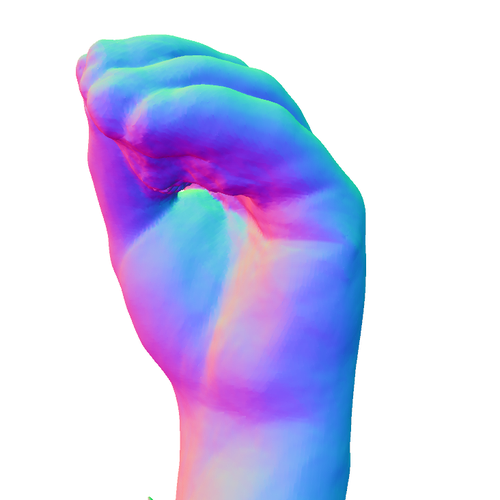}};

    \node (p02) at(p10.east) [anchor=west, xshift=-1ex]{\includegraphics[width=\imw]{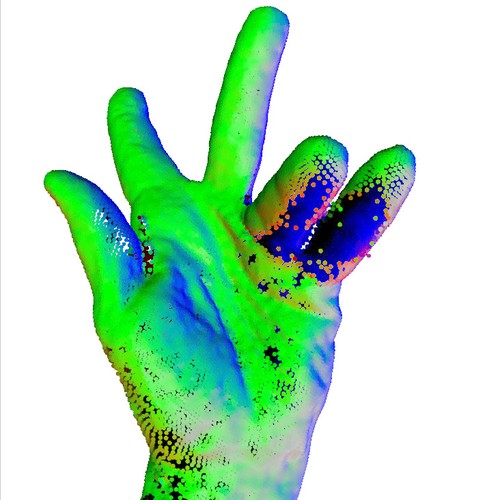}};
    \node (p12) at(p02.east) [anchor=west, xshift=-1ex] {\includegraphics[width=\imw]{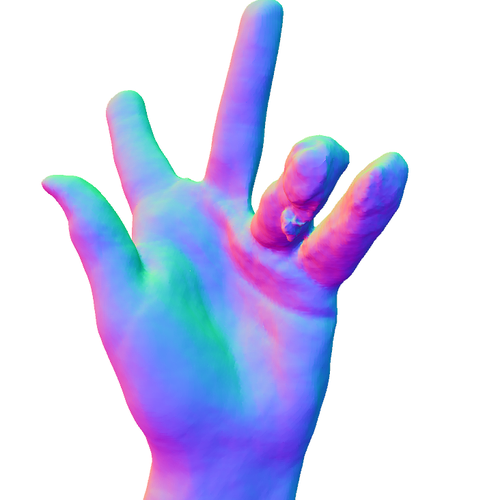}}; %
    
    \node (p03) at(p12.east) [anchor=west, xshift=-1ex]{\includegraphics[width=\imw]{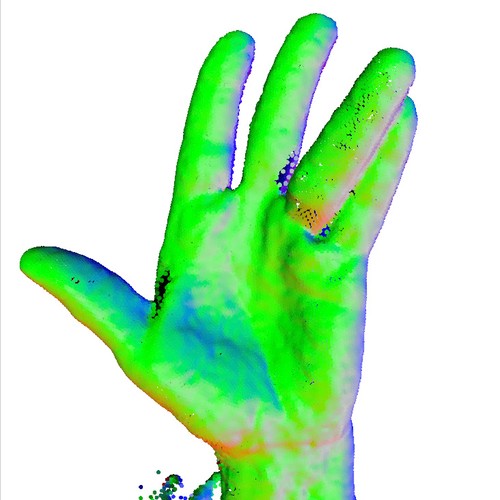}};
    \node (p13) at(p03.east) [anchor=west, xshift=-1ex] {\includegraphics[width=\imw]{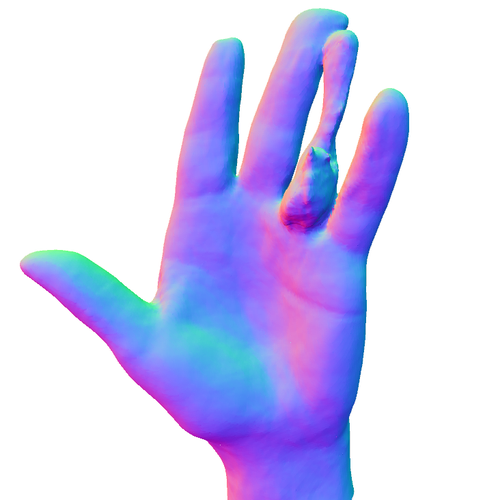}};
    
\end{tikzpicture}
\caption{Failure cases, due to missing significant observations or rare shapes too different from training. }
 	\vspace{-1em}
\label{fig:failurecase}
\end{figure}

%% file: section/conclusions.tex
\section{Conclusion}
This paper proposes \method, a neural deformable model with disentangled identity to reconstruct SDF from dynamic 3D point clouds in arbitrary coordinates. For this purpose, we explicitly represent deformable shapes with two embedding spaces, one to describe the identity related information and the other to describe the generic deformation among all shape instances. The experiments show DiForm achieves the separation between identity and deformation with a good evidence. The deformation codes can drive similar deformation for different identity embedding. The identity code can be solved with a few observations, and then be conditioned for various deformations. We further developed an end-to-end solution to reconstruct observations in arbitrary coordinates using the learnt embedding. Our algorithm is evaluated on a large 3D hand scans, outperforming the state-of-the-art hand modeling methods. 
Dispite not yet having all the answers, DiForm is a valuable exploration to the solution space and show new possibilities with a purely implicit representation. Future direction could be to explore hybrid approaches that combine with template models.

%% file: supp/dataset.tex
\section{\dataset Dataset}\label{sec:dataset}
In this paper, we proposed an algorithm to learn an identity-disentangled neural embedding for deformable shapes that can be used to reconstruct dynamic meshes. To evaluate our algorithm, we require a substantial 3D dataset that contains both inter-identity and intra-identity shape variations.  One challenging test case is the space of human hands.  First, hand shape varies significantly between people.  Second, a single individual's hand shape when posed deforms significantly due to both the articulation of the hand bones and the complicated soft tissue dynamics of the muscles and tendons.  However, the similarities in anatomical structure between individuals mean they should deform in a similar (though not identical) manner given the same pose inputs.   We therefore collected a large-scale 3D scans of hands using a commercial 3DMD scanner to support this study. We use 3DMD's proprietary software to fuse the outputs of five RGB-D cameras to reconstruct a sequence of 3D meshes. %

\begin{figure*}
  \centering
  \includegraphics[width=0.98\textwidth]{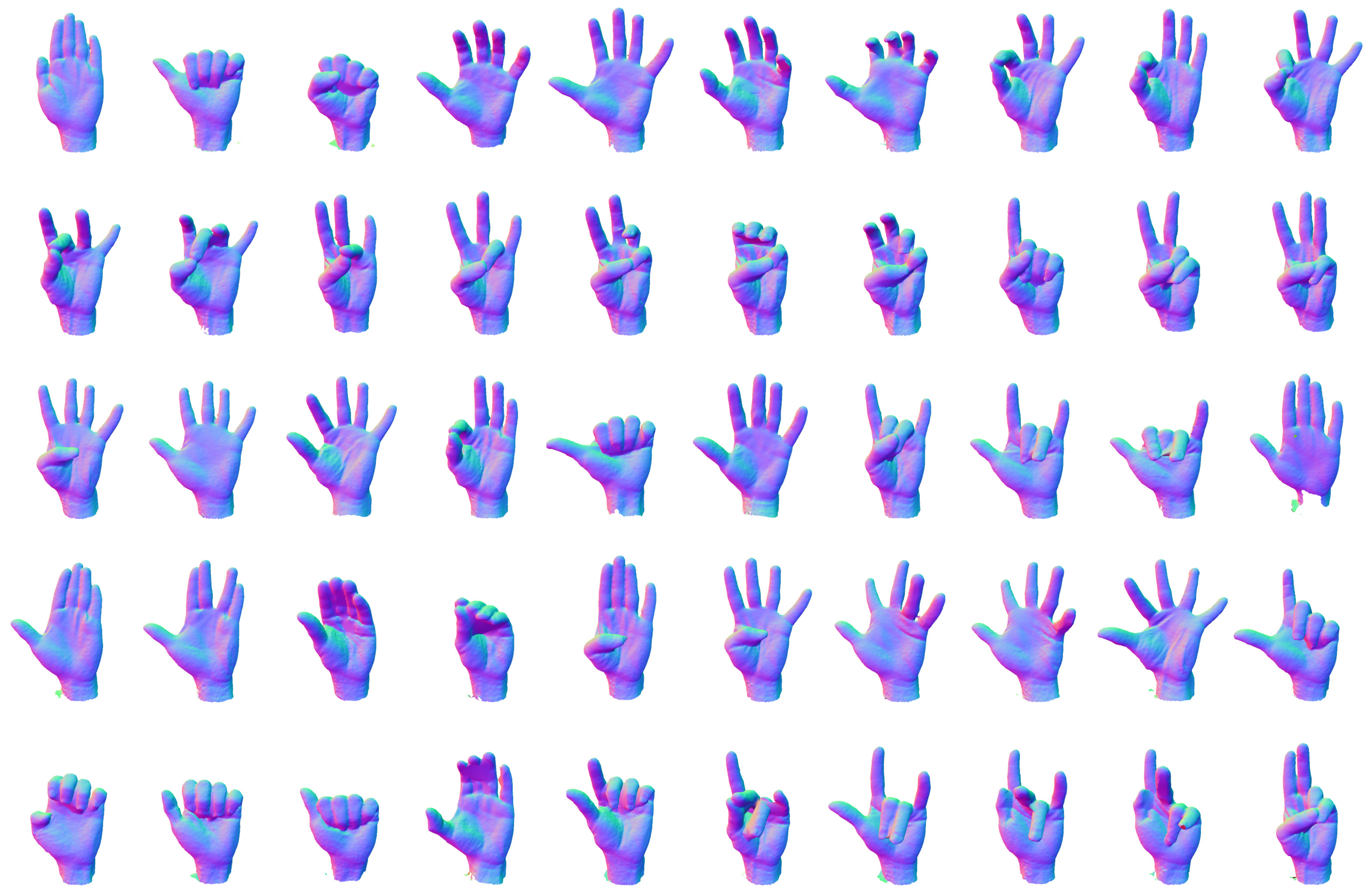}
  \caption{An example 3D capture of the same left hand performing the predefined gestures.}
  \label{fig:6014}
\end{figure*}

\begin{figure*}
  \centering
    \begin{tikzpicture}[inner sep=0pt, font=\small]
      \node(p0) {\includegraphics[width=0.98\textwidth]{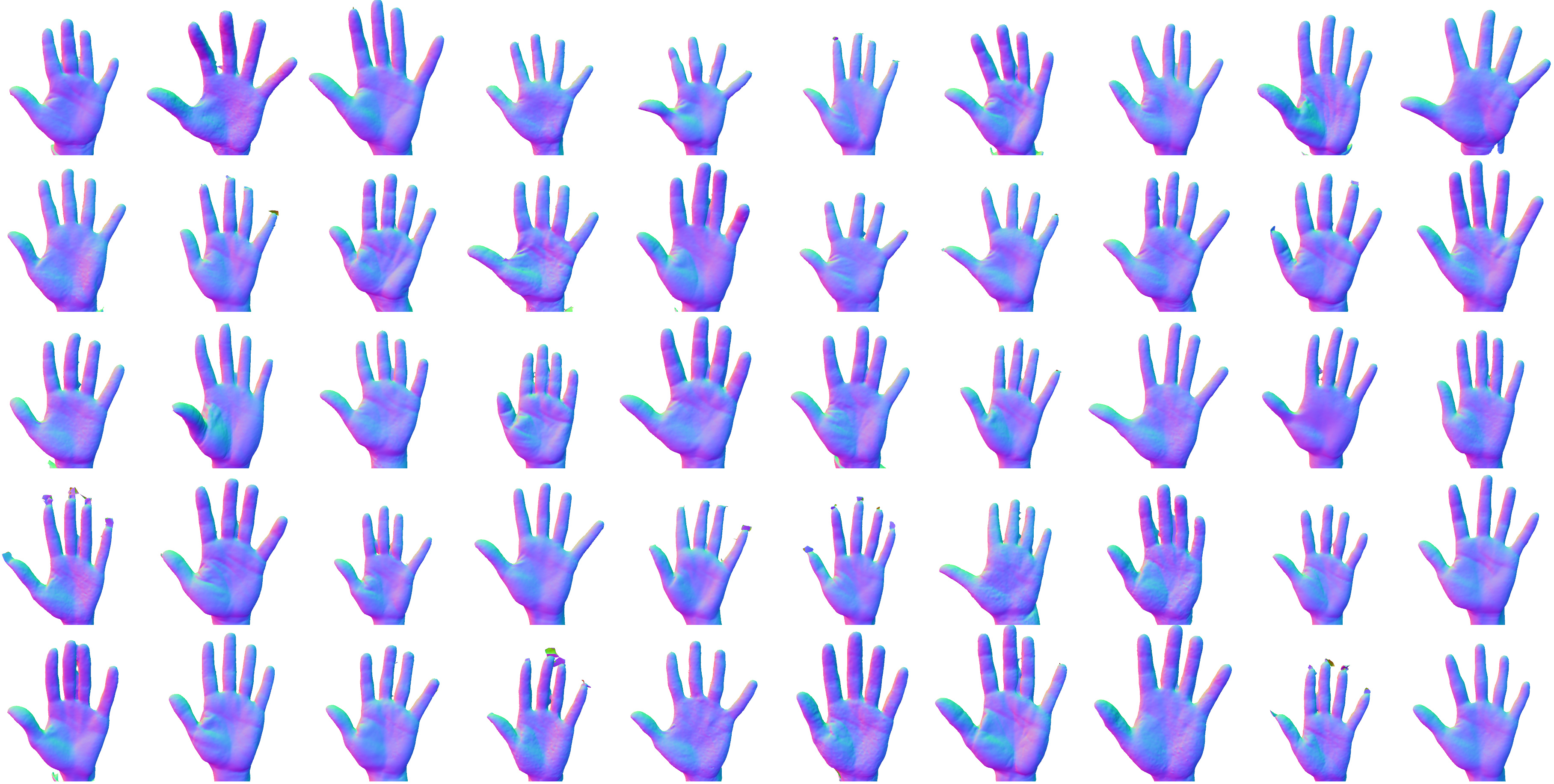}};
      \node(t0) at(p0.south)[anchor=north, yshift=-2ex] {(a) An randomly selected 50 hands performing the similar gesture.};
      \node(p1)at(t0.south)[anchor=north, yshift=-2ex] {\includegraphics[width=0.98\textwidth]{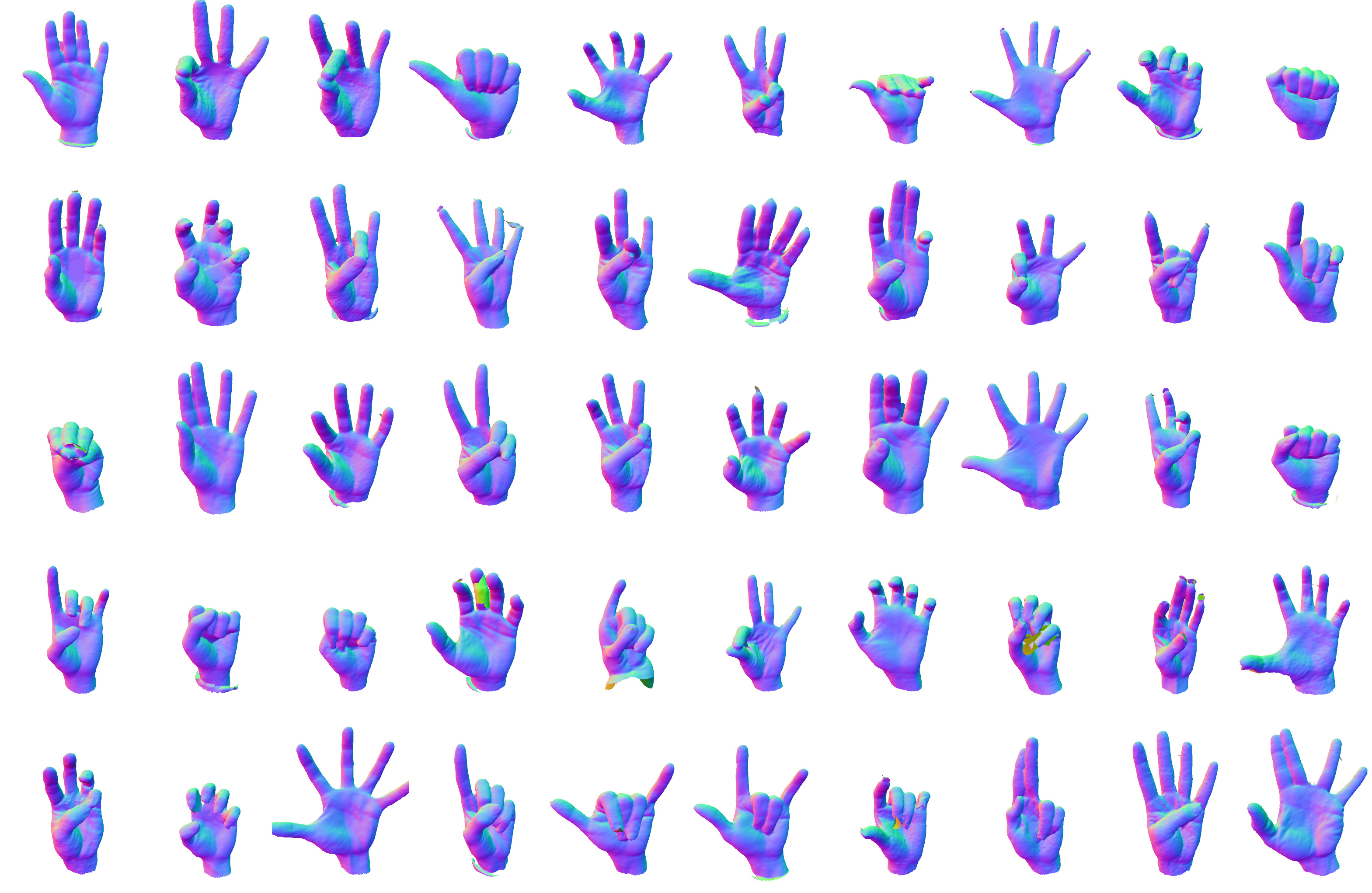}};
      \node(t1) at(p1.south)[anchor=north, yshift=-2ex] {(b) An randomly selected 50 hands performing a random gesture.};
    \end{tikzpicture}
  \caption{An overview of 50 different hands from our \dataset dataset. The dataset contains in total 160 different hands. }
  \label{fig:dataset-overview}
\end{figure*}

\paragraph{Data capture protocol.} 
In total we capture two sets of data. First, the \dataset dataset contains 183 different left hands. During capture, each participant performs a predefined series of gestures with their left hands, including counting, pointing, grabbing etc (c.f. \cref{fig:6014}). \cref{fig:dataset-overview}(a) shows 50 different hands perform a similar gesture and \cref{fig:dataset-overview}(b) visualizes a subset of 50 different hands with a random selected gesture per hand. During the capture, all participants rest their left hands on a fixed plaform. Consequently, all captured meshes are roughly oriented and aligned.  Each captured sequence is tagged with the hand identity, but no other annotations for the gesture identity or correspondences are included. Note that the data capture is a continuous process, so the transition gestures from one predefined gesture to the next are also captured. 

For the second collection, 3DH-4D dataset, we remove the arm rest and ask participant to perform random left hand motion freely in the space, without resting it on the fixed handle bar. As a result, these captures do not align in the canonical coordinates with the \dataset dataset. A total of 13 sequences from 5 people are captured, who are not included in the previous data capture.

\paragraph{Comparison to MANO.} Our capture setting is similar to the one used in MANO dataset~\cite{Romero:etal:TOG2017}. However, our dataset contains 6 times as many hand shapes than MANO and captures significantly more meshes per hand. The comparison of the two datasets is given in \cref{tab:dataset}. Note that MANO dataset contains both left and right hands, where half of the meshes capture hands holding an object. Our dataset only contains variations of left hands without any objects. Besides, the captured meshes in our dataset are pre-aligned and thus can be directly used to train shape embedding. We have also provided continuous 4D sequences for evaluating 4D reconstruction performance in the wild.

\begin{table}[tb]
\begin{threeparttable}
\newcommand{\xmark}{\ding{55}}
\newcommand{\cmark}{\ding{51}}
    \centering
    \footnotesize
    \caption{Statistic comparison of the MANO dataset~\cite{Romero:etal:TOG2017} and the \dataset dataset (ours).}
    \label{tab:dataset}
    \setlength{\tabcolsep}{1.2pt}
    \begin{tabular}{R{8ex} R{10ex} R{10ex} R{10ex} R{16ex} R{8ex} }
    	\toprule
		dataset &hands & meshes &left hand &avg. mesh/hand &aligned \\
	    \midrule
    	MANO &31 &1554 &296 \tnote{*}  &11.4$\pm$6.2 &\xmark\\ %
    	\dataset &183 &14854 &14854 &91.5$\pm$28.3 &\cmark\\
        \bottomrule
    \end{tabular}
\begin{tablenotes}
	\item[*]: The MANO datasset contains in total 659 meshes of left hands, but only 296 meshes are without objects presence.
\end{tablenotes}
\end{threeparttable}
\end{table}

%% file: supp/implement.tex
\section{Implementation Details}\label{sect:impl}
This section provides additional implementation details excluded in the main submission due to the page limit. 

\begin{figure}[t]
	\centering
	\includegraphics[width=\columnwidth]{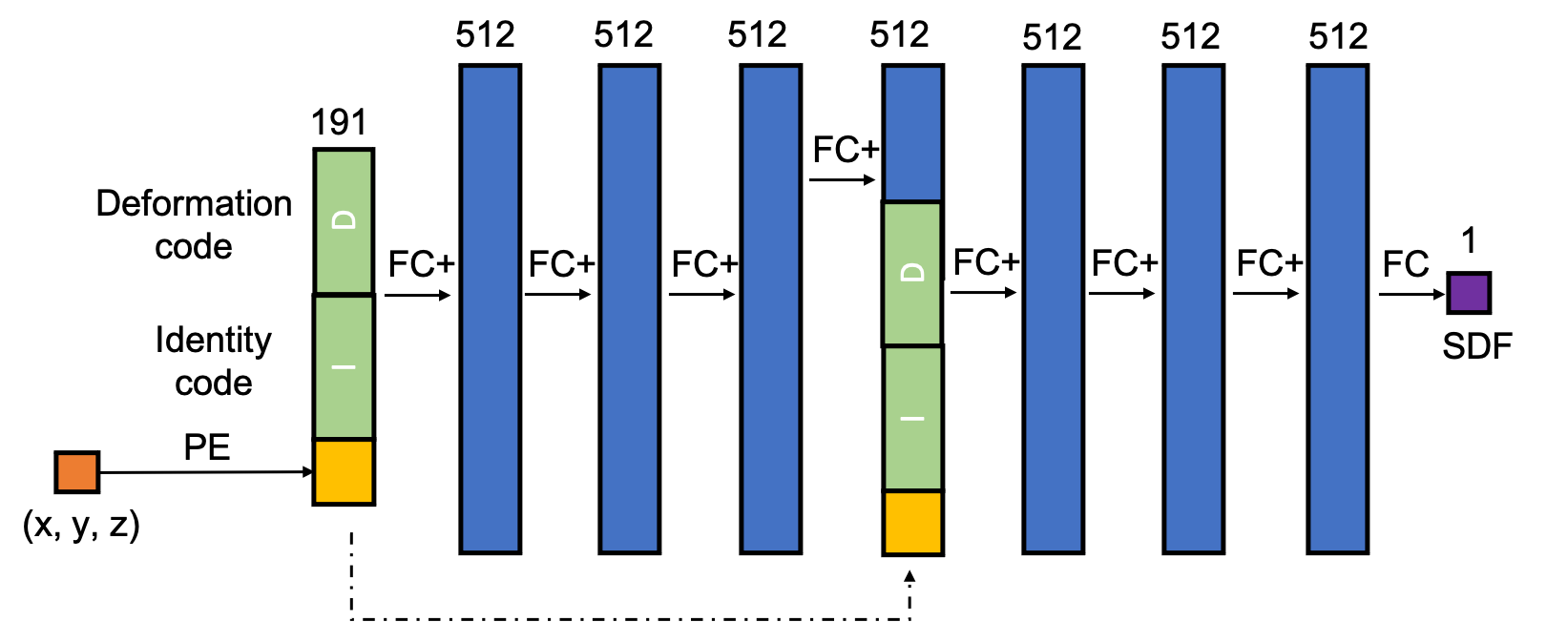}%
	\caption{The architecture of the shape embedding network. For each 3D point $(x, y, z)$, we apply positional encoding (PE) to decompose it to 10-frequency bands. This 63-dimensional vector is then concatenated with a 64-dimensional deformation code and a 64-dimensional identity code to yield a 191-dimension input vector. The input vector is fed through a decoder, which contains eight fully-connected (FC) layers with one skip connection. FC+ denotes a FC with a following softplus activation and the last FC layer output a single SDF value.}
	\label{fig:deephandnet_arch}
\end{figure}

\paragraph{Shape Embedding Network}
We base the shape embedding network on the auto-decoder architecture proposed in DeepSDF~\cite{Park:etal:CVPR2019} (c.f. \cref{fig:deephandnet_arch} for the network architecture). The network is composed of eight fully-connected layers, where a skip connection is used to shortcut the input to the fourth layer. We add two modifications. First, positional encoding (PE)~\cite{Mildenhall:etal:ECCV2020} with 10 frequency bands is applied to map the input 3D location into a 63-dimensional vector. Second, the encoded input is concatenated with two latent codes, one for deformation and another for identity. To train with this explicit code space separation, each training shape has its own deformation code, whereas all shapes belonging to the same person share one identity code. In our experiments, the identity and the deformation codes both have 64 dimensions to model hands.

To reconstruct a mesh given a point cloud in the canonical space at inference time, we follow the same procedure as proposed in DeepSDF~\cite{Park:etal:CVPR2019}. First, the latent codes are optimized using Equation (2) as defined in the main submission. Once the latent codes are found, we reconstruct a 3D SDF volume by passing the coordinates of each voxel center through the decoder to compute a SDF value. Finally, the mesh is extracted from the SDF volume using the standard Marching Cubes algorithm~\cite{Lorensen:Cline:SIGGRAPH1987}. In all experiments, we randomly sample 16K points on surface and 16K points off surface. %
To generate the off-surface points, 8K points are sampled around the surface by Gaussian distribution and the remaining 8K are randomly sampled in the volume. To extract a mesh, the SDF volume is set to be at 256-voxel resolution in each dimension.

In all our training and testing, we use the same hyper-parameters in the loss weights. $\lambda_{s}$, $\lambda_{n}$,  $\lambda_{g}$, $\lambda_{i}$, $\lambda_z$, $a$, are set to be 30, 1, 0.1, 1, 0.0001, and 100, respectively.

\paragraph{Pose Prediction Network}
\cref{fig:deephandnet_arch} illustrates the architecture of the pose prediction network (PN). Given an observed point cloud in world coordinates, we use PN to estimate a 6DoF transformation taking it to the canonical space. The idea is to take a reference shape in the canonical coordinates and predict the relative transformation between the target input and the reference. During training, we use the shape corresponds to the $\boldsymbol{0}$-code as reference, with a small random perturbation added to the $\boldsymbol{0}$ code at each iteration. At inference, we always use the $\boldsymbol{0}$-code shape as reference.
\begin{figure}[t]
	\centering
	\begin{subfigure}[b]{0.5\textwidth}
         \centering
	\includegraphics[width=0.8\textwidth]{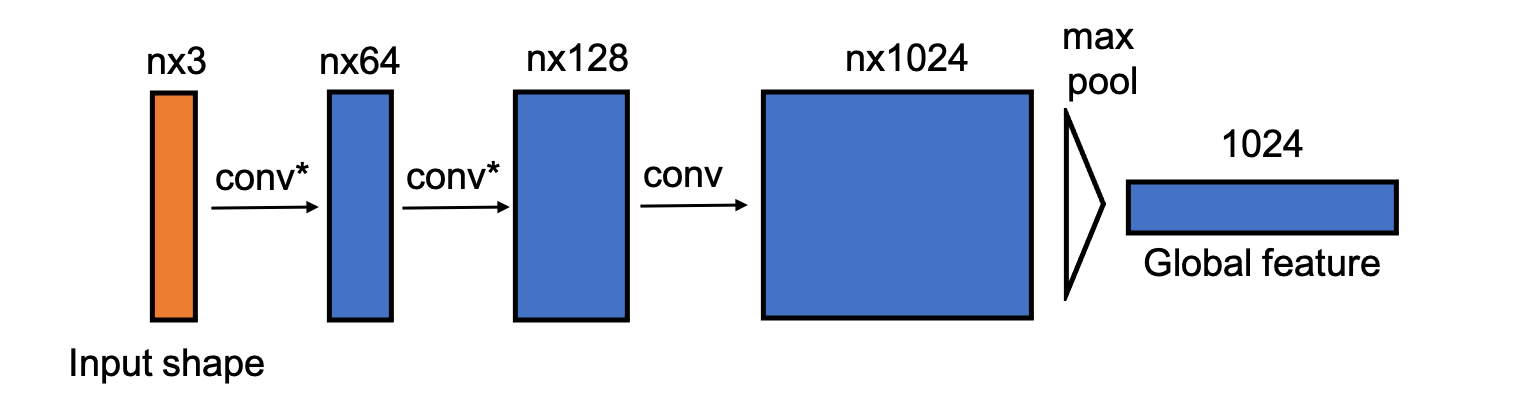}%
         \caption{The feature encoder based on PointNet~\cite{Qi:etal:CVPR2017} to extract a feature vector from a point cloud. First, a point cloud is fed through three convolutional layers to obtain point-wise feature vectors. Then all point features are aggregated via max pooling to output a 1024-dimensional global feature. We use Conv to denote 1-dimensional convolutional and Conv* to denote a Conv layer with Leaky-ReLU activation.}
         \label{fig:posenet_feature}
     \end{subfigure}
     
     \begin{subfigure}[b]{0.5\textwidth}
         \centering
	\includegraphics[width=\textwidth]{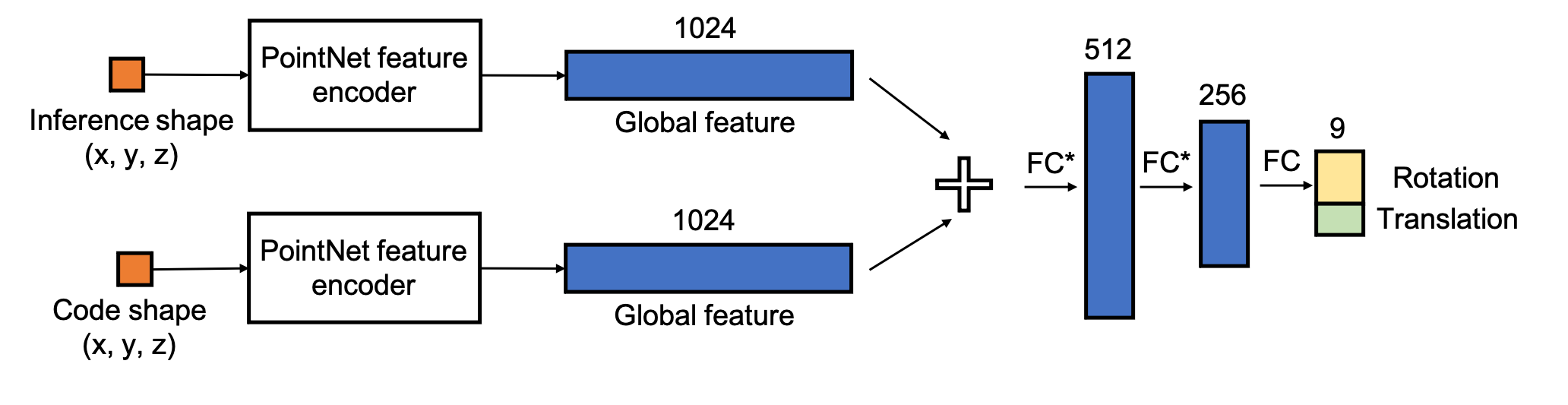}%
         \caption{The architecture of the pose prediction network (PN). We first feed the target shape and the reference shape through the weight-sharing Siamese feature encoder (\cref{fig:posenet_feature}). The output features from each pair of shapes are concatenated to be filtered by three fully-connected layers, where the last layer outputs the 9-dimensional transformation parameters. The first six dimensions represents rotation parameters as proposed in~\cite{Zhou:6drot:2019cvpr} and the last three dimensions form the translation. FC is a fully-connected linear layer and FC* is a fully-connected layer with a following Leaky-ReLU activation.}
         \label{fig:posenet_pose}
     \end{subfigure}
     
	\caption{The architecture of the pose prediction network to estimate the 6DoF transformation from the world coordinate to the canonical space.}
	\label{fig:posenet}
\end{figure}

\paragraph{Policy gradient}
In the policy gradient initialization, we uniformly sample 2000 rotations as the action space. At each iteration, we sample 400 rotation hypothesis from the action space and then evaluate the loss from Equation (2) after 10 shape reconstruction optimization steps. We use 250 sampled surface points to evaluate each hypothesis. We use ADAM to optimize the policy gradient with a constant learning rate 0.01.

%% file: supp/experiments.tex
\section{Additional Experiment Results}\label{sec:exp}

\subsection{Interpolation and Deformation Transfer}
In \cref{fig:interp} we show two sets of latent code interpolation. To generate the results, we randomly select some deformation codes and identity codes from training to be the anchors. For the first row, we linearly interpolate the anchor identity codes while keeping the deformation codes fixed. Similarly, we interpolate the deformation code along the first column, where the deformation codes are interpolated while the identity codes are fixed. To populate the remaining entries, the identity code is held the same along each column, which is combined with the deformation codes from the first column per row. These results indicate the latent space of both identity and deformation is smooth and deformation can be properly transferred from from one identity to another. 
We also notice interpolation artifacts, such as the 3rd row from the bottom in \cref{fig:interp}(b) and \cref{fig:linear_interpolation}. The artifacts occur when interpolating two relatively distant deformation. This suggests the latent space is not always Euclidean smooth, causing some linear interpolated values to diviate from the embedding space. Notice that our training data is dominant by a set of predefined hand poses. With a more diverse training set, the latent space is expected to be better learnt.

\input{supp/fig_interpolation.tex}

\input{supp/tsne}

\subsection{Ablation Study of the Code Dimension}
We design an ablation study to evaluate the impact of latent code dimensions on reconstruction quality. To this end, we densely paired 16, 32 and 64 dimensions for the identity and deformation and report the reconstruction performance on \dataset by DiForm-S in \cref{tab:code_dim}. It can be seen that the accuracy improves with the dimension of the identity and deformation increases. The performance gap slows down as both the identity and the deformation reaches 64 dimension.

\begin{table}[tb]
	\centering
	\footnotesize
	\caption{The impact of the latent code dimensions of identity (I) and deformation (D) on the reconstruction quality. Chamfer Distance (CD) is reported in millimeter. }
	\label{tab:code_dim}
	\setlength{\tabcolsep}{3.0pt}
	\begin{tabular}{lccccccc}
		\toprule
		Methods &$\mu(\text{CD})$ &$\sigma(\text{CD})$ &$\mu\left(^r_g\text{CD}\right)$ &$\sigma\left(^r_g\text{CD}\right)$ &$\mu\left(^g_r\text{CD}\right)$ &$\sigma\left(^g_r\text{CD}\right)$ \\
		\midrule
		I16-D16 & 1.7359& 0.4407& 2.0605& 0.6984& 1.4125& 0.232 \\
		I16-D32 & 1.5016& 0.3452& 1.8129& 0.5828& 1.1914& 0.1352 \\
		I16-D64 &1.4015& 0.3197& 1.6952& 0.5514& 1.1078& 0.1122 \\
		I32-D16 & 1.6204& 0.4162& 1.9460& 0.6857& 1.2948& 0.1869 \\
		I32-D32 & 1.4466& 0.3503& 1.7447& 0.6045& 1.1474& 0.1257 \\
		I32-D64 &1.3806 &0.3195 &1.6743  &0.5645 &1.0857 &0.1027 \\
		I64-D16 &1.5676& 0.4287& 1.9306& 0.7412& 1.2046& 0.1556 \\
		I64-D32 &1.4136 &0.3238& 1.7227&0.5806& 1.1056 &\textbf{0.0920} \\
		I64-D64 &\textbf{1.3542}&\textbf{0.2954}& \textbf{1.6413} &\textbf{0.5165}& \textbf{1.0669} &0.1056 \\
		\bottomrule
	\end{tabular}
\end{table}

\subsection{Ablation Study on Pose Prediction}
In Section 4.2 of the main submission, we evaluate the 6DoF prediction and report the quantitative results.%
In this section, we further report the ablation studies conducted behind the design of our pose prediction component.
To recap: we proposed two methods to find the 6DoF transformation that aligns the observed point cloud to the canonical space. The first solution uses Policy Gradient (PG) to optimize the distribution of pose hypotheses such that the highest probability pose is the correct one. The second solution directly regresses the pose using the pose prediction network (PN). Given a large hypothesis pool, PG guarantees finding a good solution. However, it is expensive to compute at runtime in comparison to the PN solution. The drawback of the PN is it must be trained for a particular class of objects and so a network trained for hands will not generalize to other domains. 

\cref{tab:poseinit-ablation} presents further ablation studies for pose initialization methods. The same benchmark data is used as in the main submission, which contains 136 shapes with a random 6DoF transformation applied to the canonical space (c.f. \cref{tab:se3stat} for the pose statistics).

We first reported an ablation study on the case where we do not perform a pose initialization step. We denote it as `identity' in the table. It can be seen in the \cref{tab:poseinit-ablation-after} that pose estimation and shape reconstruction both have very high error before and even after the joint optimization. Therefore, it is important to initialize the transformation from the input world coordinate to the canonical space properly. As a comparison, we also conduct an experiment where the ground-truth transformation is used to initialize the joint optimization, which is denoted as `GT pose'. It can be seen in the \cref{tab:poseinit-ablation-after} that this initialization gives a very low CD in terms of the shape reconstruction accuracy after the optimization. %

For PG, in each policy gradient step, we evaluate the loss from Equation (2) after 10 shape reconstruction optimization steps. We denote this version as `PG'. As a comparison, we also implemented a version which follows the standard reinforcement learning approach to evaluate the loss after each shape optimization step. We denote this version as `PG-std' and we run 10000 policy gradient steps for a fair comparison.  The remaining hyper-parameters are kept the same. As shown in the \cref{tab:poseinit-ablation}, our version with 10 shape reconstruction optimization steps shows more stable and better results in pose estimation and shape reconstruction before and after the joint optimization.

For PN, we compare a version that predicts an absolute pose directly without a reference shape as the input. We denote it as `PN-abs' where we only feed the global feature vector from the target shape without concatenating it with another feature vector from the reference shape. As a comparison, it can be seen that our version that predicts relative pose transformation shows much better results in the pose estimation and the shape reconstruction before and after the joint optimization. In the relative pose prediction part, we also conducted two ablation studies on the reference shape input. One is to use the shape generated by the $0$-code as the reference and the other is to use the same input shape represented in the canonical space as the reference. We denote them as `PN-can' and `PN-zero', respectively. It can be seen that PN-can has slightly better reconstruction results in the canonical space before optimization. However,  PN-zero and PN have better performance than PN-can after the joint optimization. By comparing PN-zero and PN, it can also be noticed that our PN version outperforms PN-zero after the joint optimization.

\begin{table}
	\centering
	\footnotesize
	\caption{Statistics of ground-truth poses used in the pose evaluation benchmark. The test data contain 136 shapes of random 6DoF transformation. }
	\label{tab:se3stat}
	\setlength{\tabcolsep}{3pt}
	\begin{tabular}{L{16ex} R{10ex} R{10ex} R{10ex} R{10ex} }
		\toprule
		  & average & std & min & max \\
		\midrule
		rotation (\textdegree) &127.3685 &  39.8378 &  30.8595 & 179.3358 \\
		translation (mm) & 54.7840 & 16.0392 & 23.0687 & 89.9536 \\
		\bottomrule
	\end{tabular}
\end{table}

\begin{table}
 \caption{Quantitative evaluation of ablation studies on pose initialization.  We evaluate pose estimation error in the rotation ($\mbf R$) and translation ($\mbf t$) using the relative pose error (RPE) metrics. We evaluate shape reconstruction error using the chamfer distance (CD) metrics w.r.t the ground truth in the canonical space and in the world space.}
 \label{tab:poseinit-ablation}
\begin{subtable}[]{0.45\textwidth}
 \centering
 \footnotesize
 \caption{Before joint shape and pose optimization (avg $\pm$ std).}
 \label{tab:poseinit-ablation-before}
 \setlength{\tabcolsep}{1pt}
	\begin{tabular}{L{8ex}C{13ex} C{13ex} C{14ex} C{14ex}}
		\toprule
		initial & RPE: $\mbf R$ (\textdegree) & RPE: $\mbf t$ (mm) & CD: canonical (mm) & CD: world (mm) \\ 
		\midrule
		identity & 127.4$\pm$39.84 & 54.78$\pm$16.04 & 7.299$\pm$1.569 & 28.24$\pm$9.192 \\ \midrule
		GT pose & 0.0 & 0.0 & 7.295$\pm$1.570 & 7.295$\pm$1.570 \\ 
		\midrule
		\midrule
		PG-std             &     18.67$\pm$32.18                   &10.23$\pm$7.380  &  7.399$\pm$1.604 &7.011$\pm$\textbf{1.245} \\
		PG             &     6.297$\pm$4.495                   &\textbf{5.229}$\pm$\textbf{3.134}  &  \textbf{7.289}$\pm$\textbf{1.559 }             &       \textbf{6.784}$\pm$1.330 \\ 
		\midrule
		PN-abs &   36.85$\pm$43.79  & 27.36$\pm$8.975 &   7.291$\pm$1.561  & 15.78$\pm$4.473  \\ 
		PN-can &   12.57$\pm$20.68  & 14.73$\pm$6.602 &   \textbf{7.289}$\pm$1.568  & 9.492$\pm$2.494  \\ 
		PN-zero    &   7.088$\pm$5.519  & 9.860$\pm$4.582 &   7.291$\pm$\textbf{1.559}  & 7.795$\pm$1.795   \\ 
		PN                   &   \textbf{5.253}$\pm$\textbf{3.353}  & 5.888$\pm$3.937 &   7.888$\pm$1.826  & 7.295$\pm$1.564 \\ 
		\bottomrule
	\end{tabular}
\end{subtable}

\vspace{3pt}

\begin{subtable}[]{0.45\textwidth}
 \centering
\footnotesize
 \caption{After joint shape and pose optimization (avg $\pm$ std).}
 \label{tab:poseinit-ablation-after}
 \setlength{\tabcolsep}{1pt}
	\begin{tabular}{L{8ex}C{13ex} C{13ex} C{14ex} C{14ex}}
		\toprule
		inital & RPE: $\mbf R$ (\textdegree) & RPE: $\mbf t$ (mm) & CD: canonical (mm) & CD: world (mm) \\
        \midrule
        identity & 125.5$\pm$42.66 & 40.00$\pm$14.61 & 11.61$\pm$3.657 & 11.99$\pm$2.949 \\ \midrule
		GT pose & \textbf{3.319}$\pm$\textbf{1.909} & \textbf{4.403}$\pm$\textbf{2.582} & \textbf{2.653}$\pm$\textbf{0.926} & 1.393$\pm$0.629 \\
		\midrule
		\midrule
		PG-std &15.03$\pm$33.47  &  7.378$\pm$7.438   &4.098$\pm$1.646 &    2.458$\pm$1.086\\
		PG & 5.443$\pm$3.591  &   5.167$\pm$3.370 &     3.096$\pm$\textbf{1.174}    &  1.442$\pm$0.596 \\ \hline
		PN-abs &  33.53$\pm$45.69 & 17.67$\pm$13.67&    6.392$\pm$2.795 & 6.318$\pm$5.216 \\ 
		PN-can &  9.334$\pm$21.06 & 6.546$\pm$6.091 &  3.641$\pm$2.059 & 1.818$\pm$1.760 \\ 
		PN-zero   &  5.172$\pm$4.835 & 5.043$\pm$2.980&    3.043$\pm$1.323 & 1.439$\pm$0.690 \\ 
		PN   &   \textbf{4.532}$\pm$\textbf{3.187} & \textbf{4.640}$\pm$\textbf{2.822}&    \textbf{2.930}$\pm$1.239 & \textbf{1.386}$\pm$\textbf{0.522} \\ 
		\bottomrule
	\end{tabular}
\end{subtable}
\end{table}

\subsection{Evaluation on Dynamic Reconstruction}
In the main submission, we have shown the dynamic mesh reconstruction qualitatively. This section presents the quantitative evaluations on that. %
For quantitative evaluation, we clipped the sequences into 50 frames per clip and compare the end-to-end reconstruction quality for IGR~\cite{Gropp:etal:ICML2020}, IGR with postional encoding (denoted as IGR-PE) and the proposed DiForm. For fair comparison, all methods are initialized with the same 6DoF pose, computed by our pose prediction network. \cref{tab:eval4d} reports the results on 4D reconstruction and it can be seen that our proposed 4D reconstruction pipeline achieves the optimal performance across all the sequences. We direct reviewers to the video supplementary for qualitative visualization.

\input{supp/table_4dscan.tex}

%% file: supp/fig_interpolation.tex
\begin{figure*}
  \def\imw{0.49\textwidth}
  \centering
  \begin{tikzpicture}[inner sep=0pt,font=\footnotesize]
  \begin{scope}
\def\imh{1.54}
\def\xoff{7.}
\def\yoff{0.00}
\def\recz{1.5}
  \node(p00)at(0,0)[anchor=north west] {\includegraphics[width=\imw]{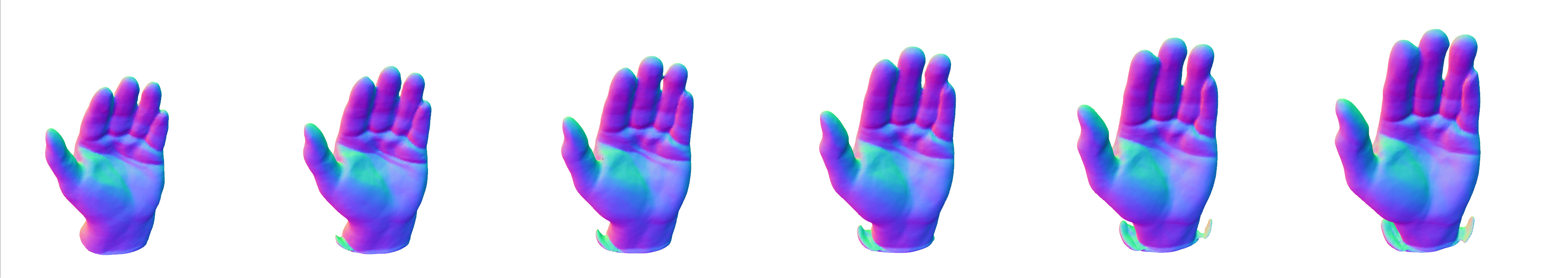}};
  \node(p01)at(p00.south)[anchor=north]{\includegraphics[width=\imw]{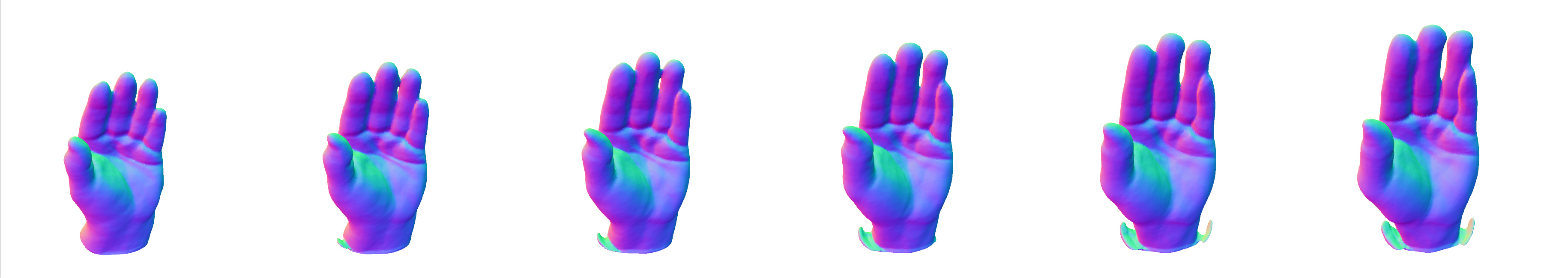}};
  \node(p02)at(p01.south)[anchor=north]{\includegraphics[width=\imw]{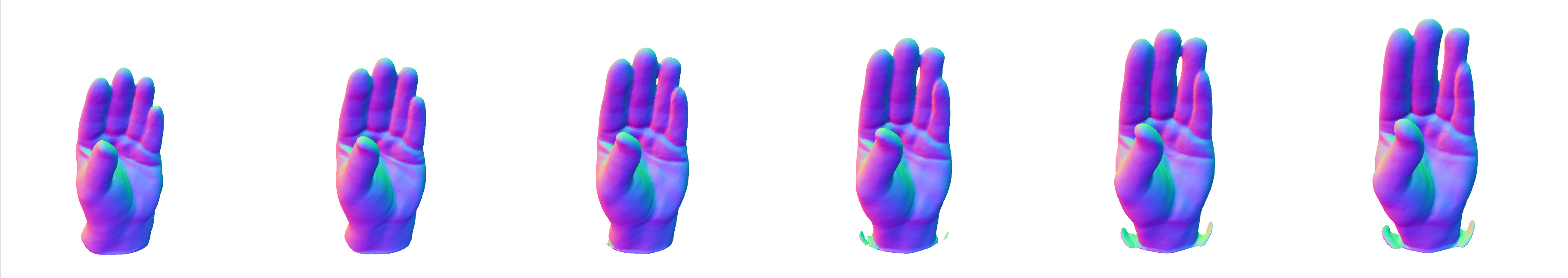}};
  \node(p03)at(p02.south)[anchor=north]{\includegraphics[width=\imw]{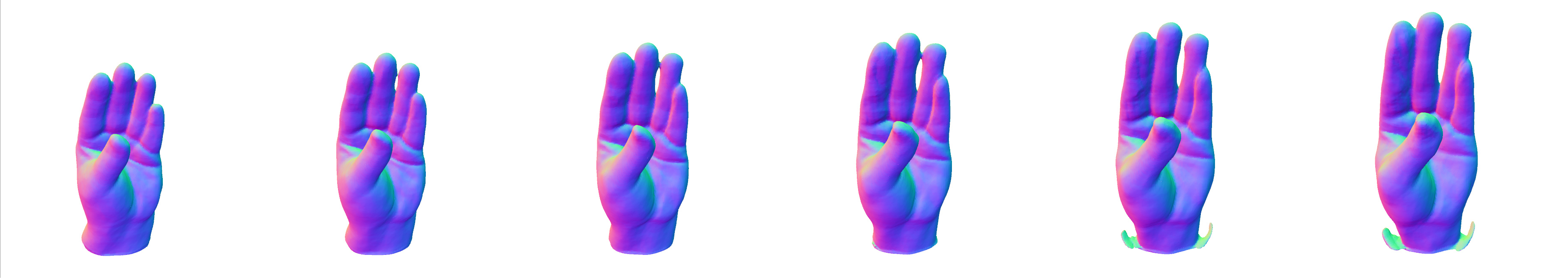}};
  \node(p04)at(p03.south)[anchor=north]{\includegraphics[width=\imw]{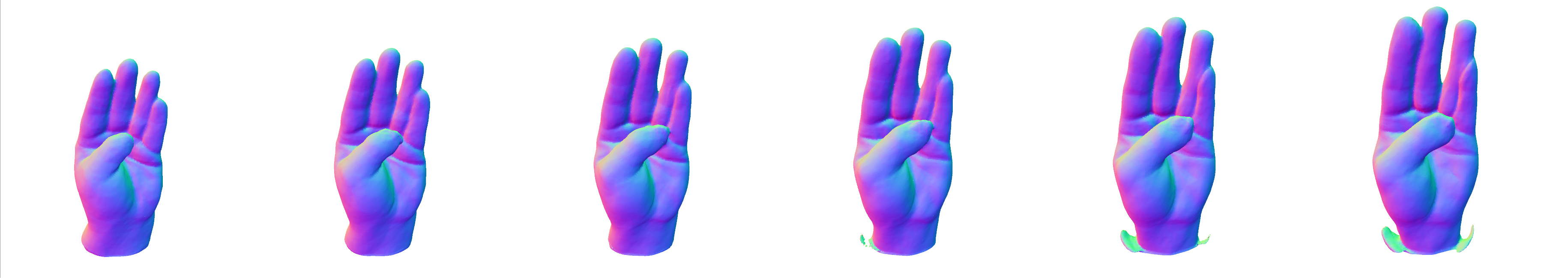}}; %
  \node(p05)at(p04.south)[anchor=north]{\includegraphics[width=\imw]{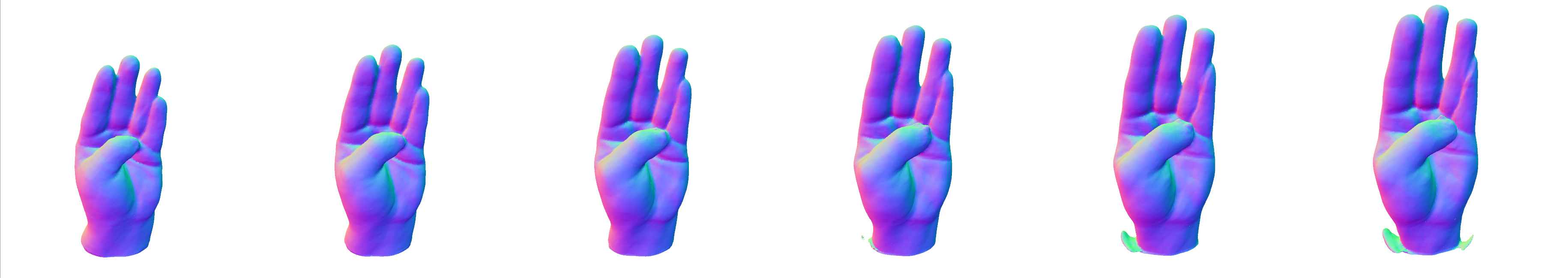}};
  \node(p06)at(p05.south)[anchor=north]{\includegraphics[width=\imw]{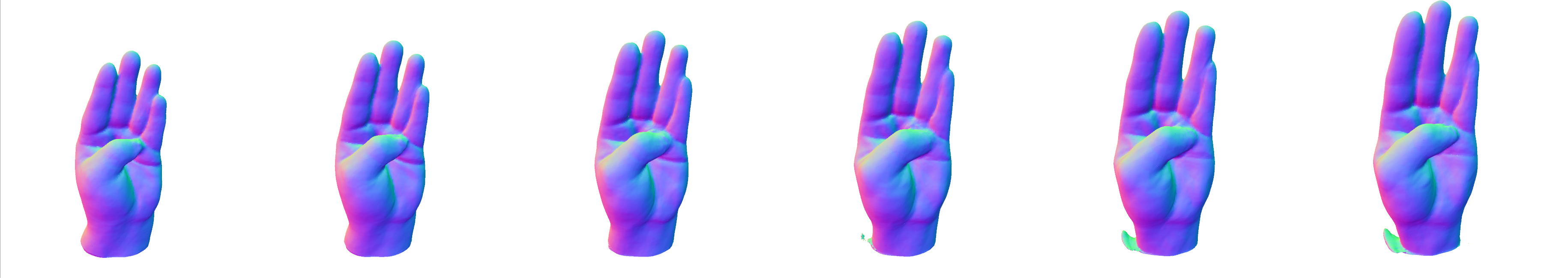}};
  \node(p07)at(p06.south)[anchor=north]{\includegraphics[width=\imw]{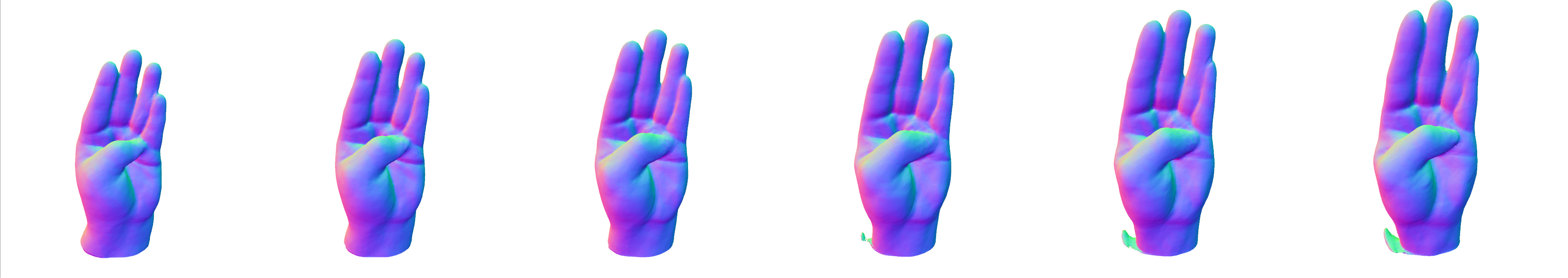}};
  \node(p08)at(p07.south)[anchor=north]{\includegraphics[width=\imw]{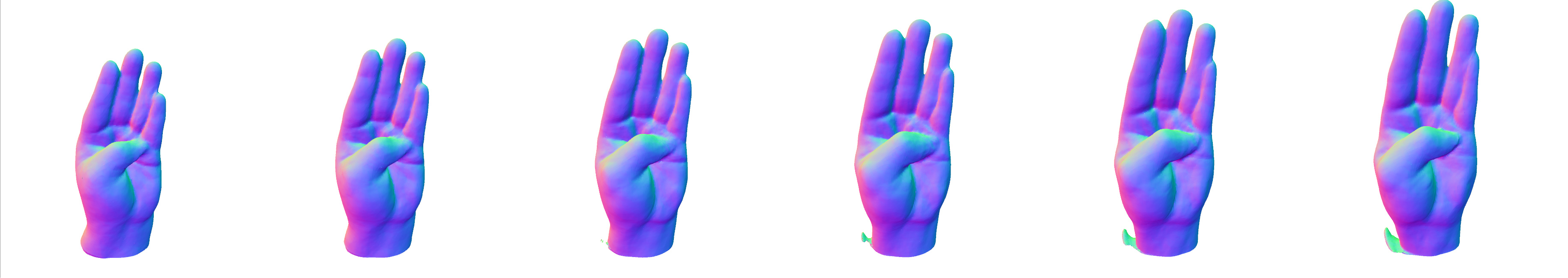}}; %
  \node(p09)at(p08.south)[anchor=north]{\includegraphics[width=\imw]{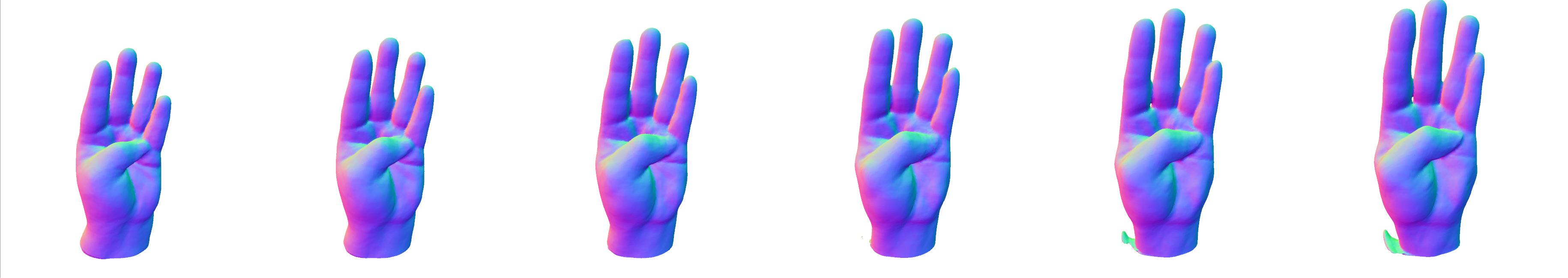}};
  \node(p10)at(p09.south)[anchor=north]{\includegraphics[width=\imw]{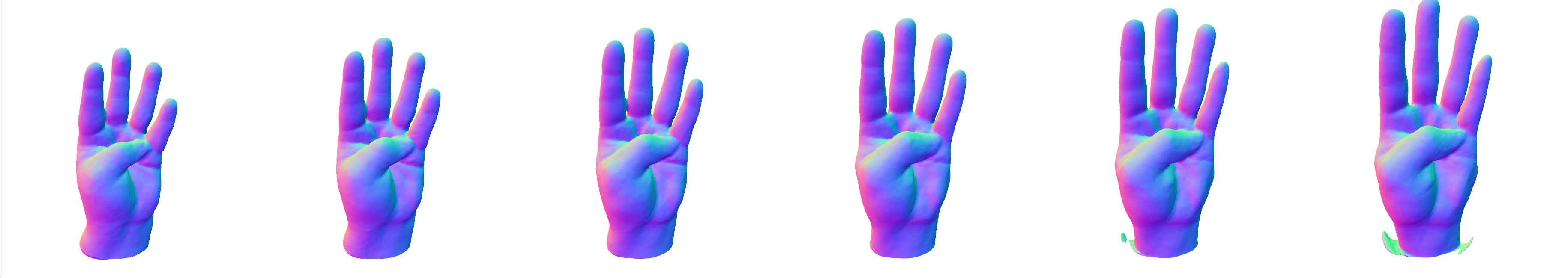}};
  \node(p11)at(p10.south)[anchor=north]{\includegraphics[width=\imw]{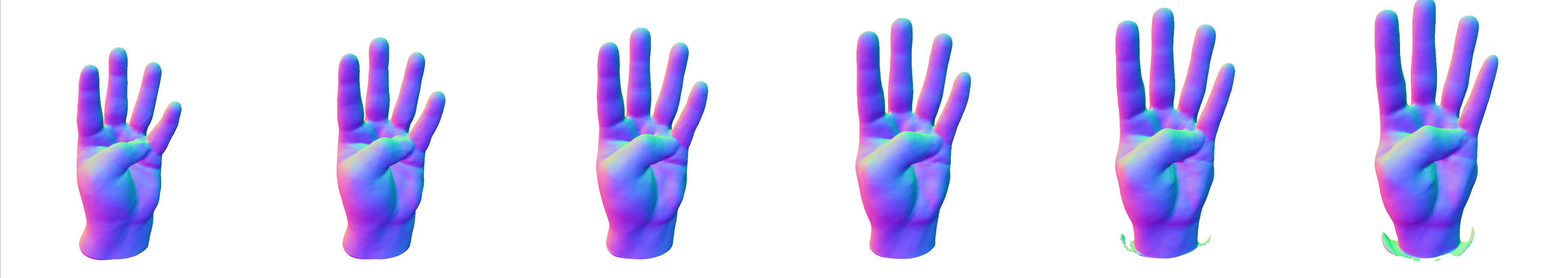}};
  \node(p12)at(p11.south)[anchor=north]{\includegraphics[width=\imw]{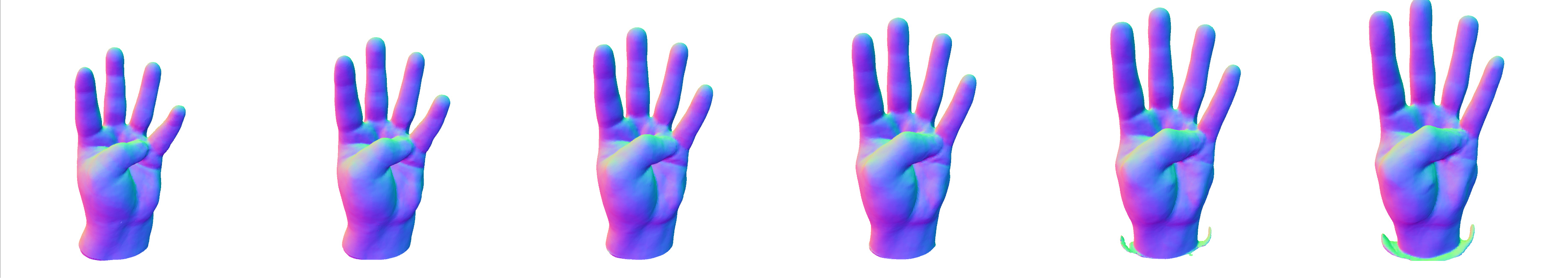}}; %

  \draw[draw=red] (0,-\yoff) rectangle ++(\recz,-\recz);
  \draw[draw=blue] (0.05,-0.05) rectangle ++(\recz-0.1,-\recz+0.1);
  \draw[draw=blue] (\xoff,-\yoff) rectangle ++(\recz,-\recz);
  \draw[draw=red] (0,-4*\imh) rectangle ++(\recz,-\recz);
  \draw[draw=red] (0,-8*\imh) rectangle ++(\recz,-\recz);
  \draw[draw=red] (0,-12*\imh) rectangle ++(\recz,-\recz);
  
  \node at(p12.south)[anchor=north,yshift=-2pt]{(a)};
  \end{scope}
  \begin{scope}[xshift=8.8cm]
\def\imh{1.54}
\def\xoff{7.}
\def\yoff{0.00}
\def\recz{1.5}
  \node(p00)at(0,0)[anchor=north west] {\includegraphics[width=\imw]{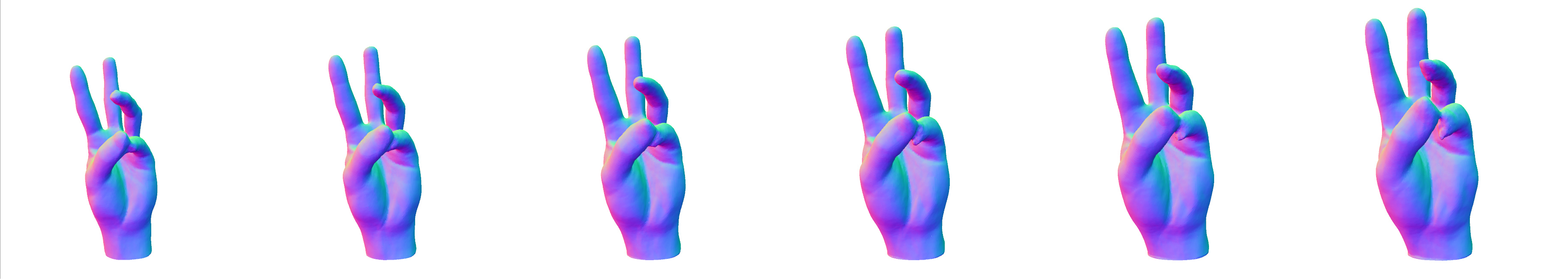}};
  \node(p01)at(p00.south)[anchor=north]{\includegraphics[width=\imw]{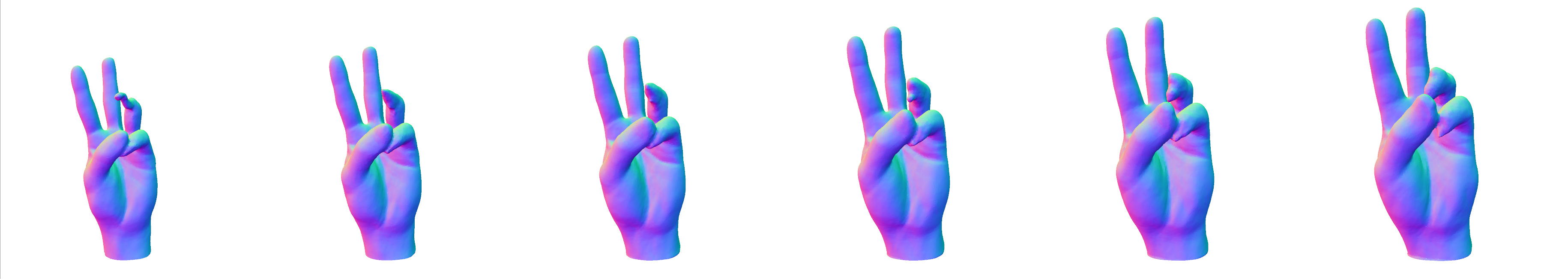}};
  \node(p02)at(p01.south)[anchor=north]{\includegraphics[width=\imw]{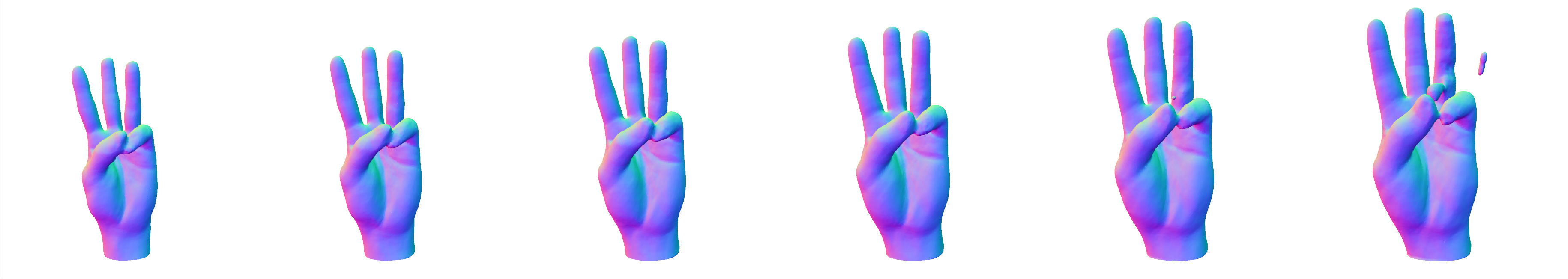}};
  \node(p03)at(p02.south)[anchor=north]{\includegraphics[width=\imw]{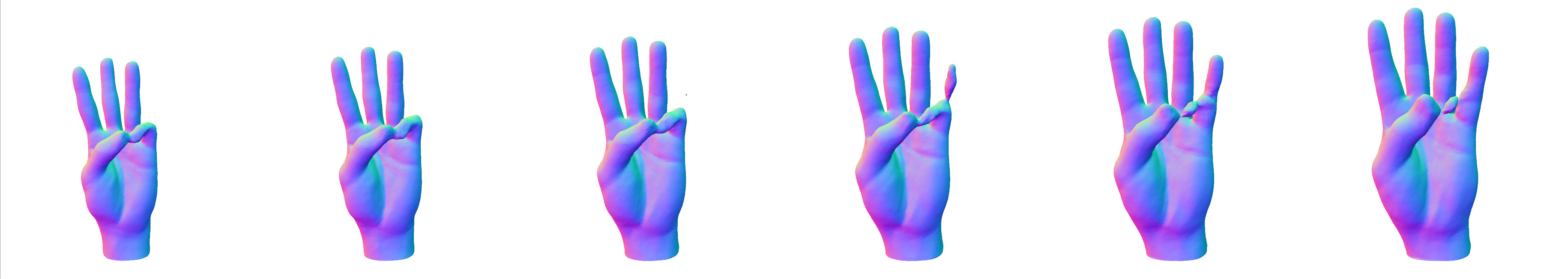}};
  \node(p04)at(p03.south)[anchor=north]{\includegraphics[width=\imw]{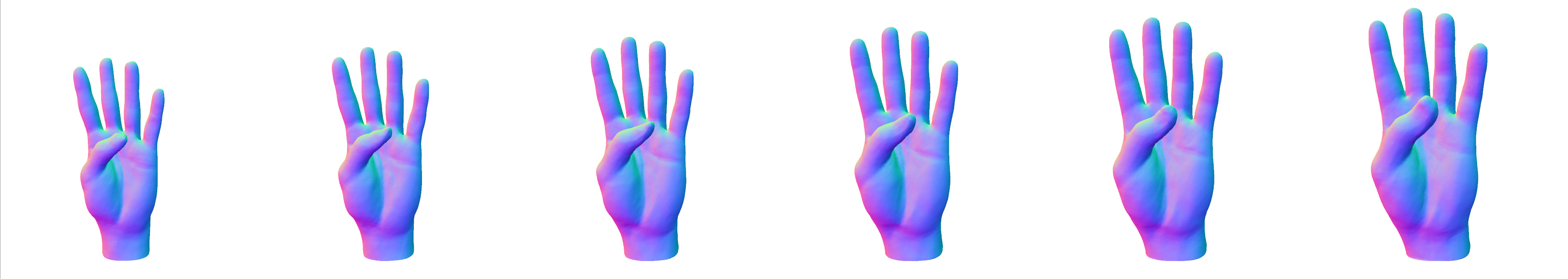}}; %
  \node(p05)at(p04.south)[anchor=north]{\includegraphics[width=\imw]{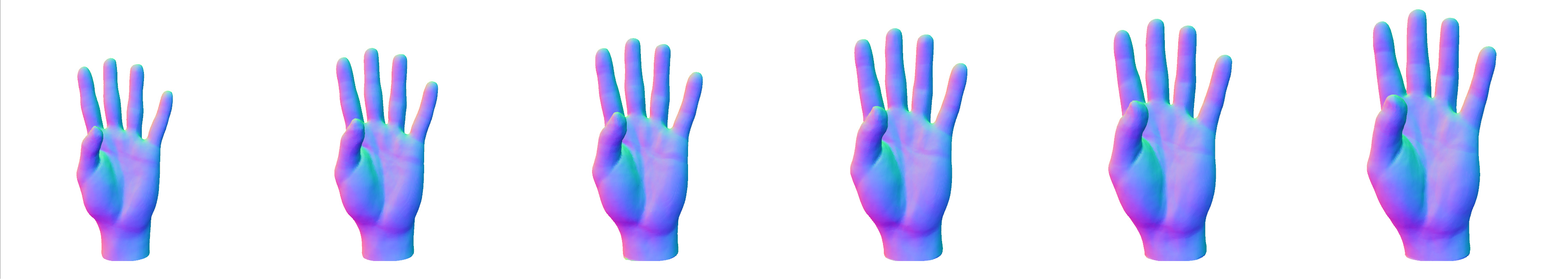}};
  \node(p06)at(p05.south)[anchor=north]{\includegraphics[width=\imw]{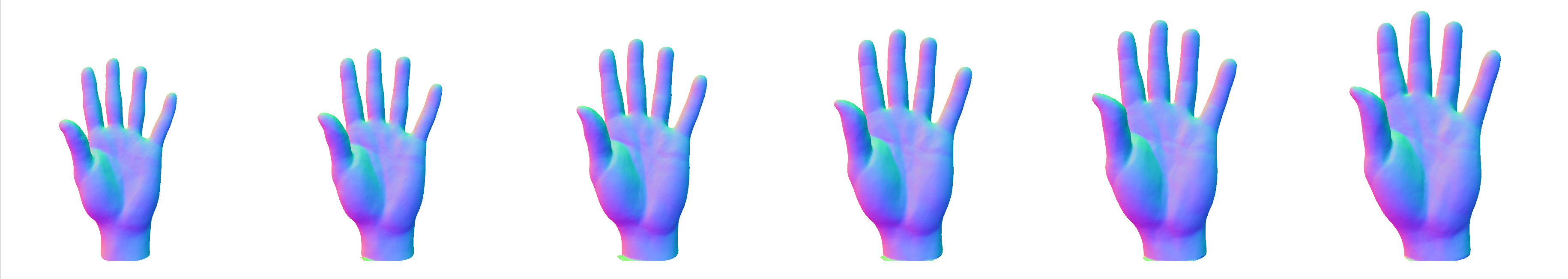}};
  \node(p07)at(p06.south)[anchor=north]{\includegraphics[width=\imw]{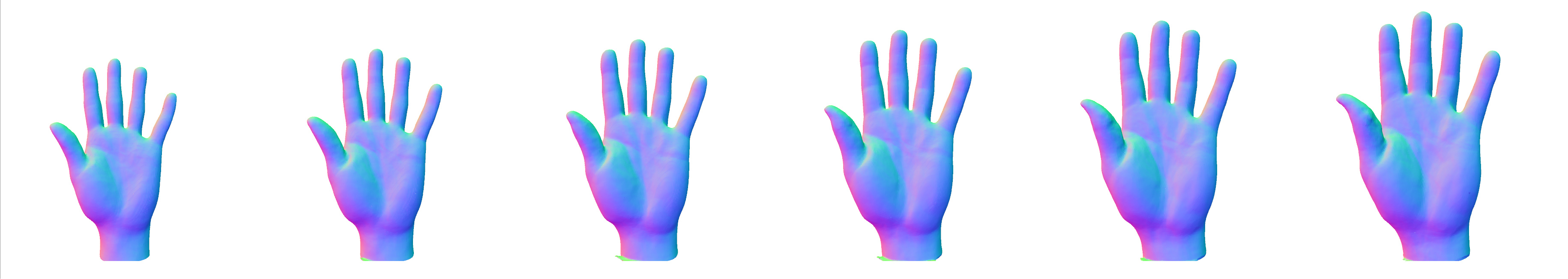}};
  \node(p08)at(p07.south)[anchor=north]{\includegraphics[width=\imw]{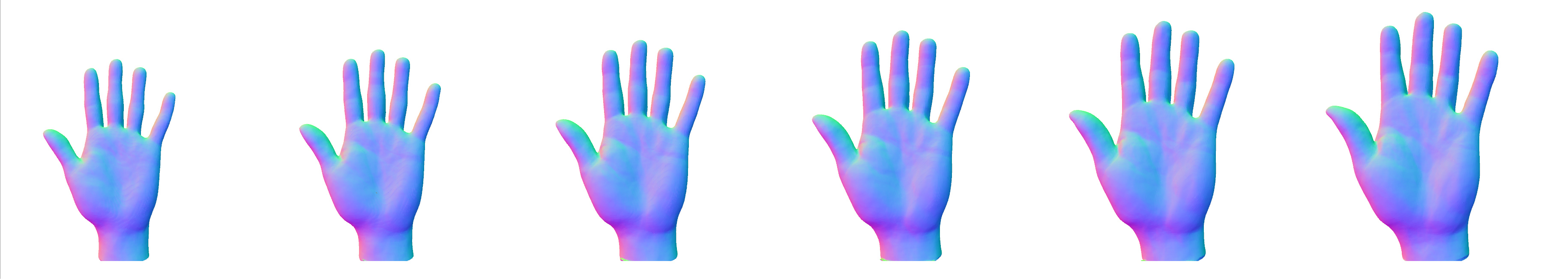}}; %
  \node(p09)at(p08.south)[anchor=north]{\includegraphics[width=\imw]{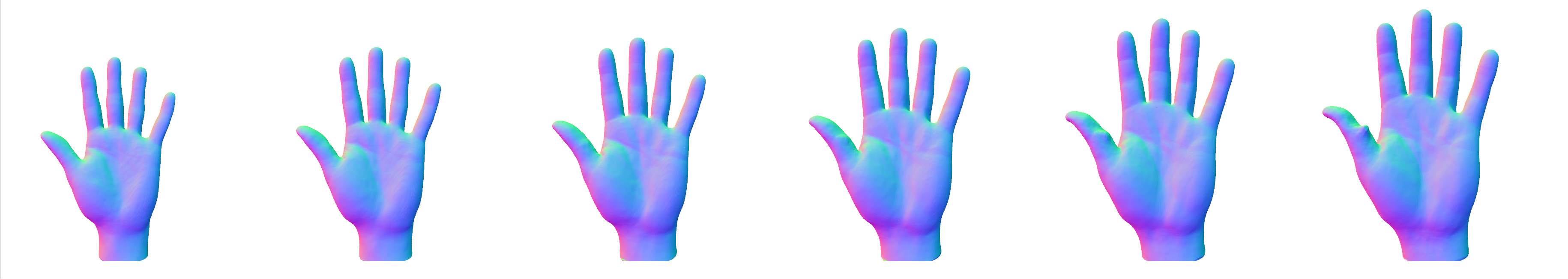}};
  \node(p10)at(p09.south)[anchor=north]{\includegraphics[width=\imw]{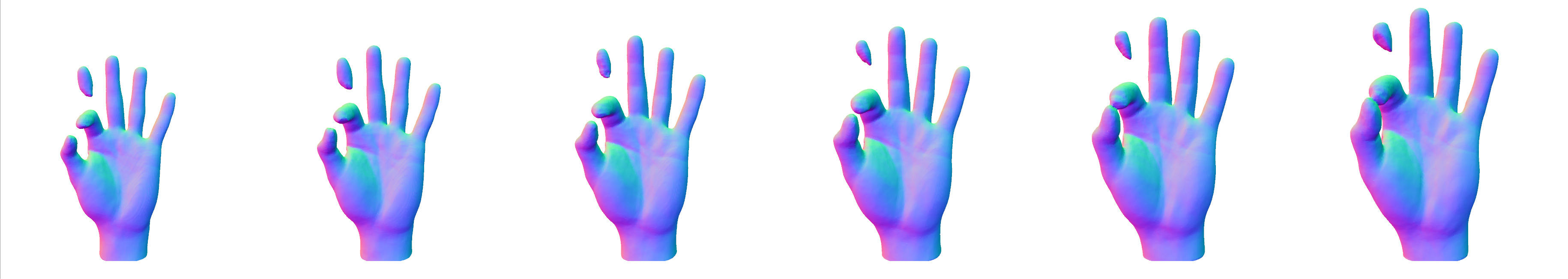}};
  \node(p11)at(p10.south)[anchor=north]{\includegraphics[width=\imw]{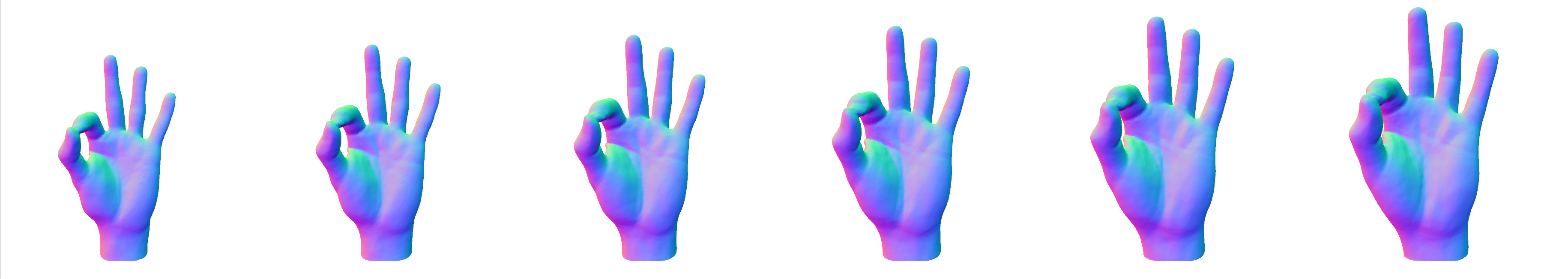}};
  \node(p12)at(p11.south)[anchor=north]{\includegraphics[width=\imw]{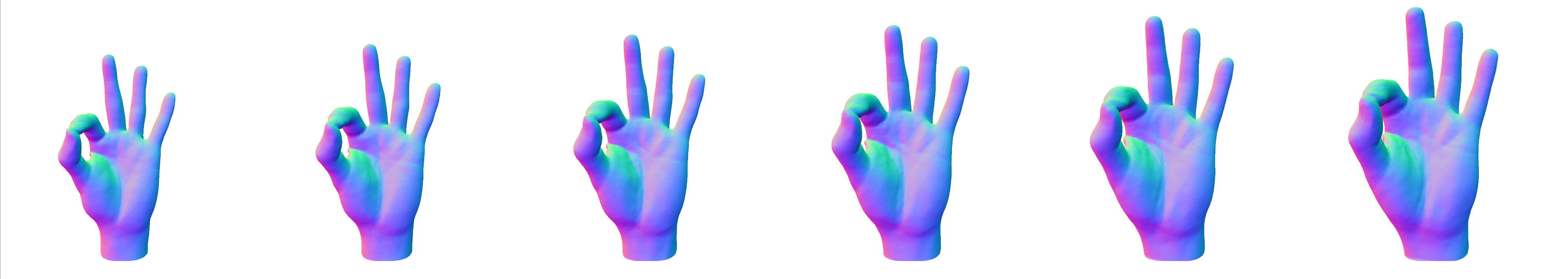}}; %

  \draw[draw=red] (0,-\yoff) rectangle ++(\recz,-\recz);
  \draw[draw=blue] (0.05,-0.05) rectangle ++(\recz-0.1,-\recz+0.1);
  \draw[draw=blue] (\xoff,-\yoff) rectangle ++(\recz,-\recz);
  \draw[draw=red] (0,-4*\imh) rectangle ++(\recz,-\recz);
  \draw[draw=red] (0,-8*\imh) rectangle ++(\recz,-\recz);
  \draw[draw=red] (0,-12*\imh) rectangle ++(\recz,-\recz);
  \node at(p12.south)[anchor=north, yshift=-2pt]{(b)};
  \end{scope}
  \end{tikzpicture}
  \caption{Latent codes interpolation and deformation transfer with two examples. In both examples, the \textcolor{red}{deformation} code is linearly interpolated along the first column with the anchor codes marked in \textcolor{red}{red} box. The \textcolor{blue}{identity} code is linearly interpolated along the first row with anchor codes marked in \textcolor{blue}{blue}. With each column being a fixed identity and each row being a fixed deformation, the figure also shows that deformation is consistently transferred to different identity. }
  \label{fig:interp}
\end{figure*}

\begin{figure*}
	
	\centering
	\def\imw{5em}
	\def\yoff{-5pt}
	\def\recz{1.7}
	\begin{tikzpicture}[inner sep=0pt]
		\node (p00) {\includegraphics[width=\imw]{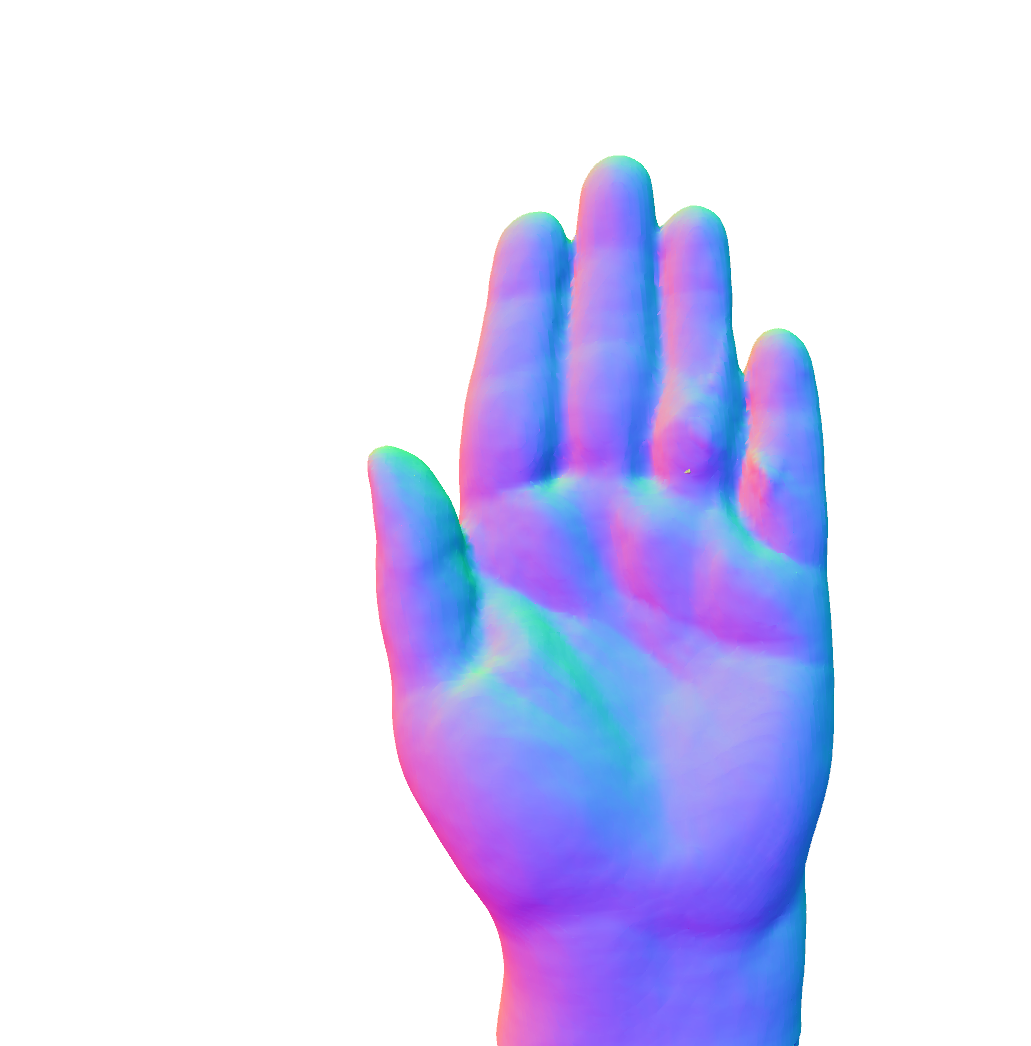}};
		
		\node (p01) at(p00.east) [anchor=west, xshift=\yoff] {\includegraphics[width=\imw]{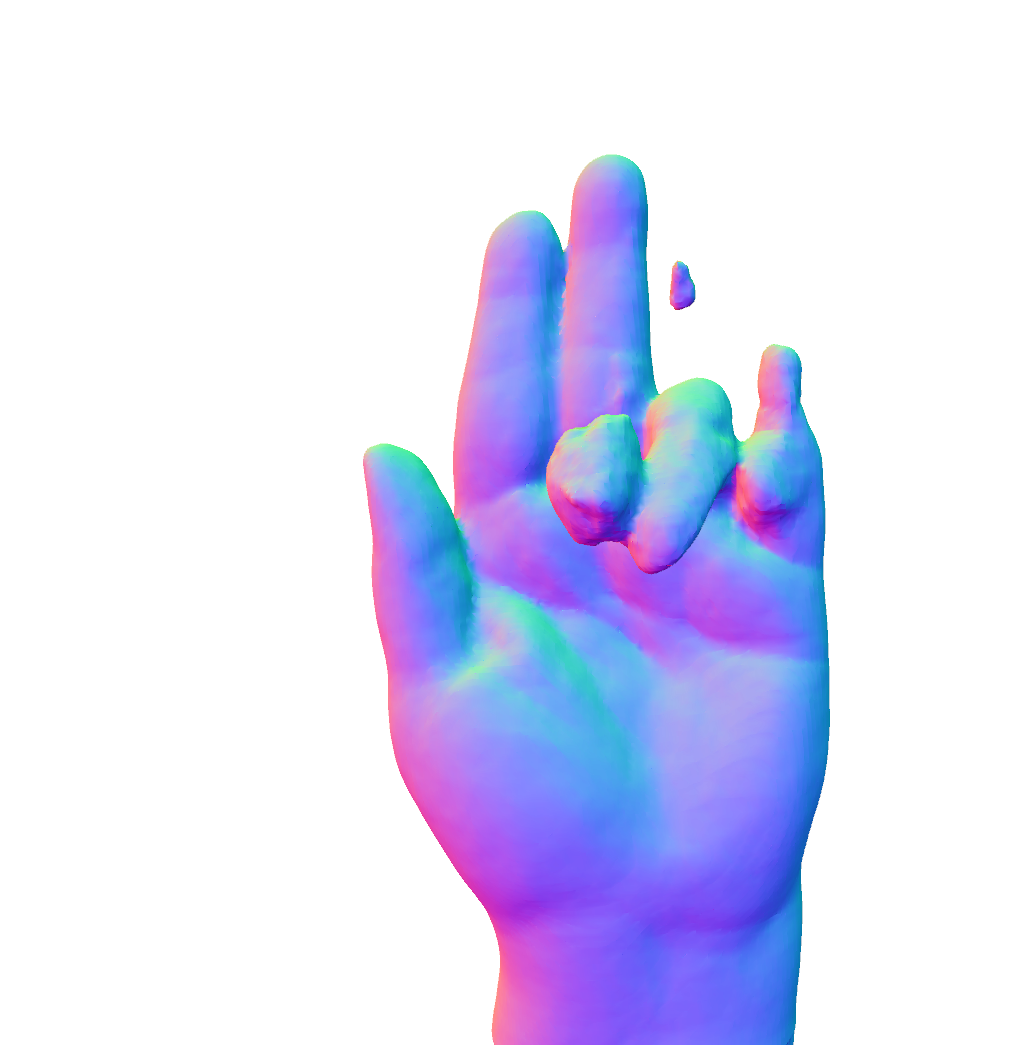}};
		\node (p02) at(p01.east) [anchor=west, xshift=\yoff]{\includegraphics[width=\imw]{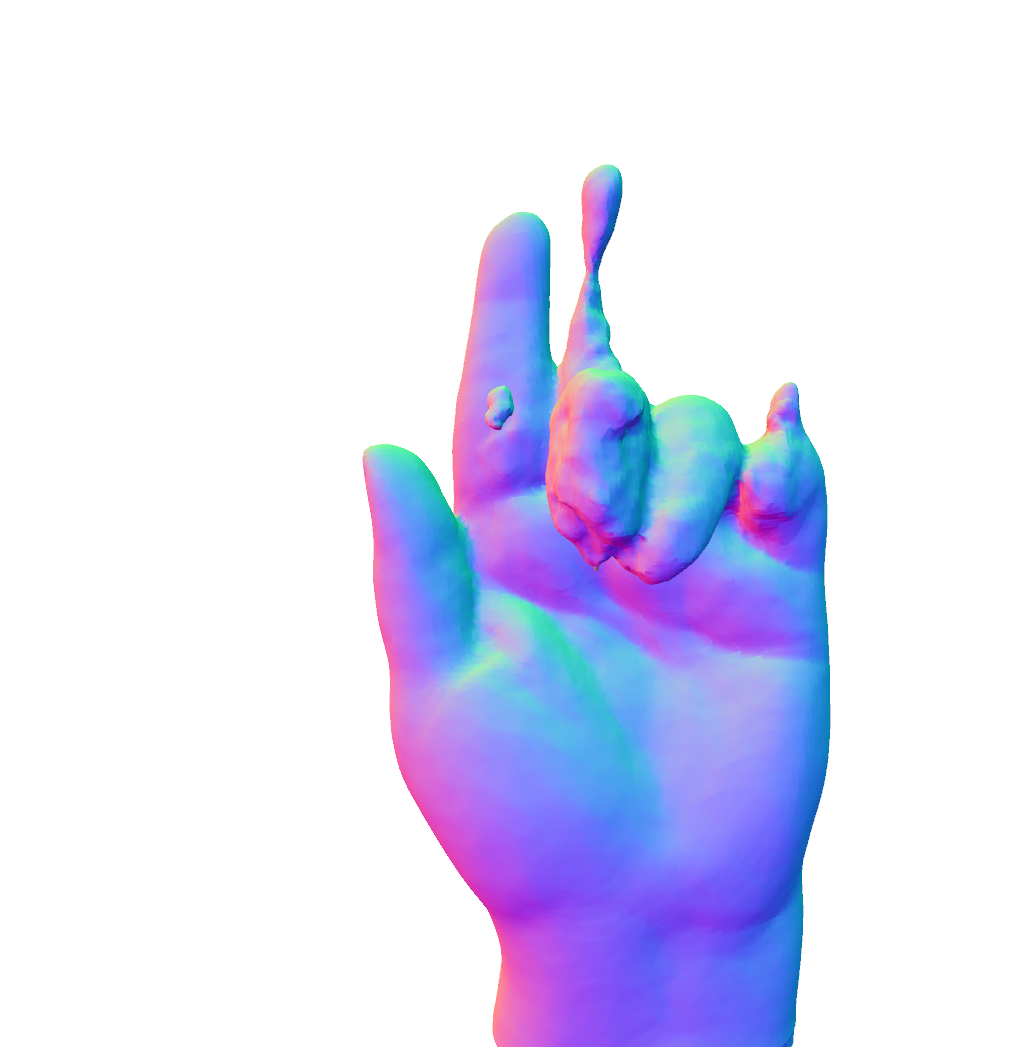}};
		\node (p03) at(p02.east) [anchor=west, xshift=\yoff] {\includegraphics[width=\imw]{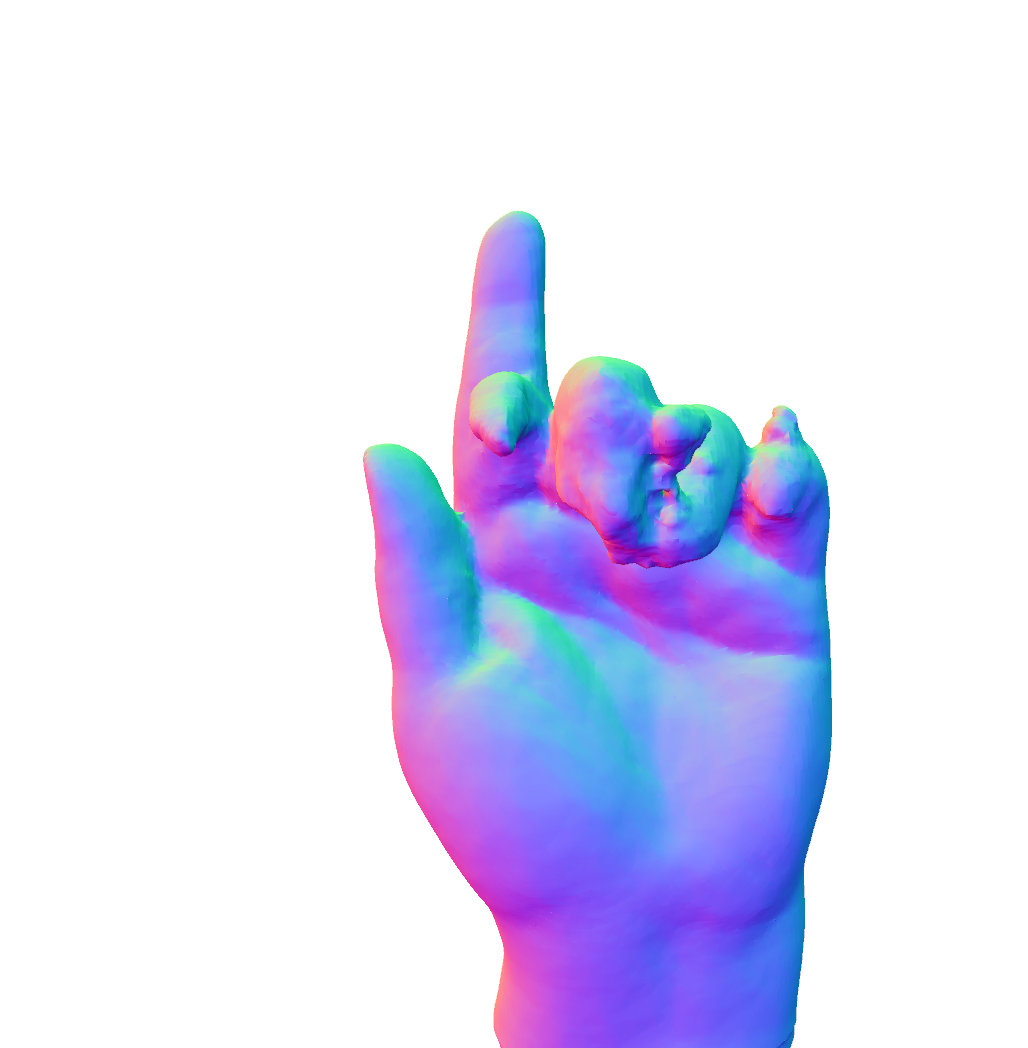}}; 
		
		\node (p04) at(p03.east) [anchor=west, xshift=\yoff]{\includegraphics[width=\imw]{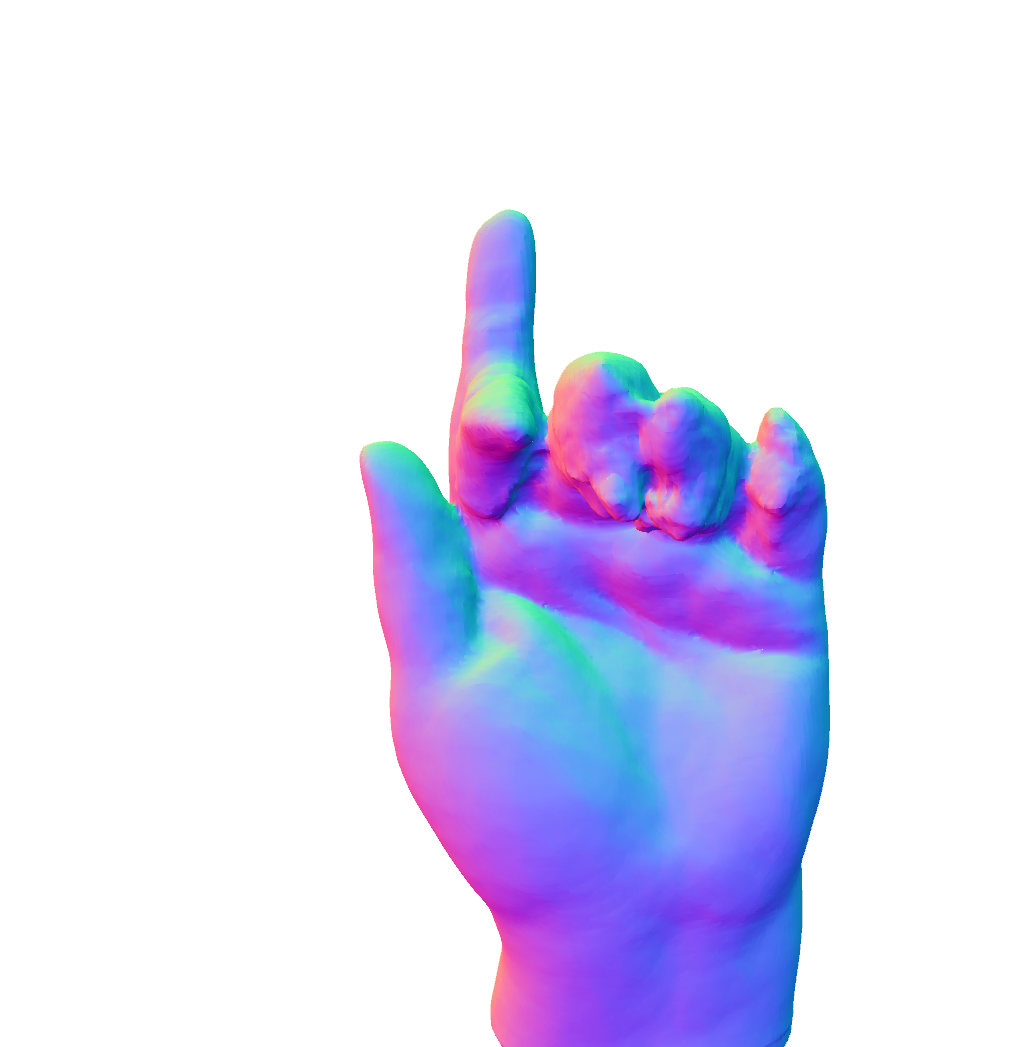}};
		\node (p05) at(p04.east) [anchor=west, xshift=\yoff] {\includegraphics[width=\imw]{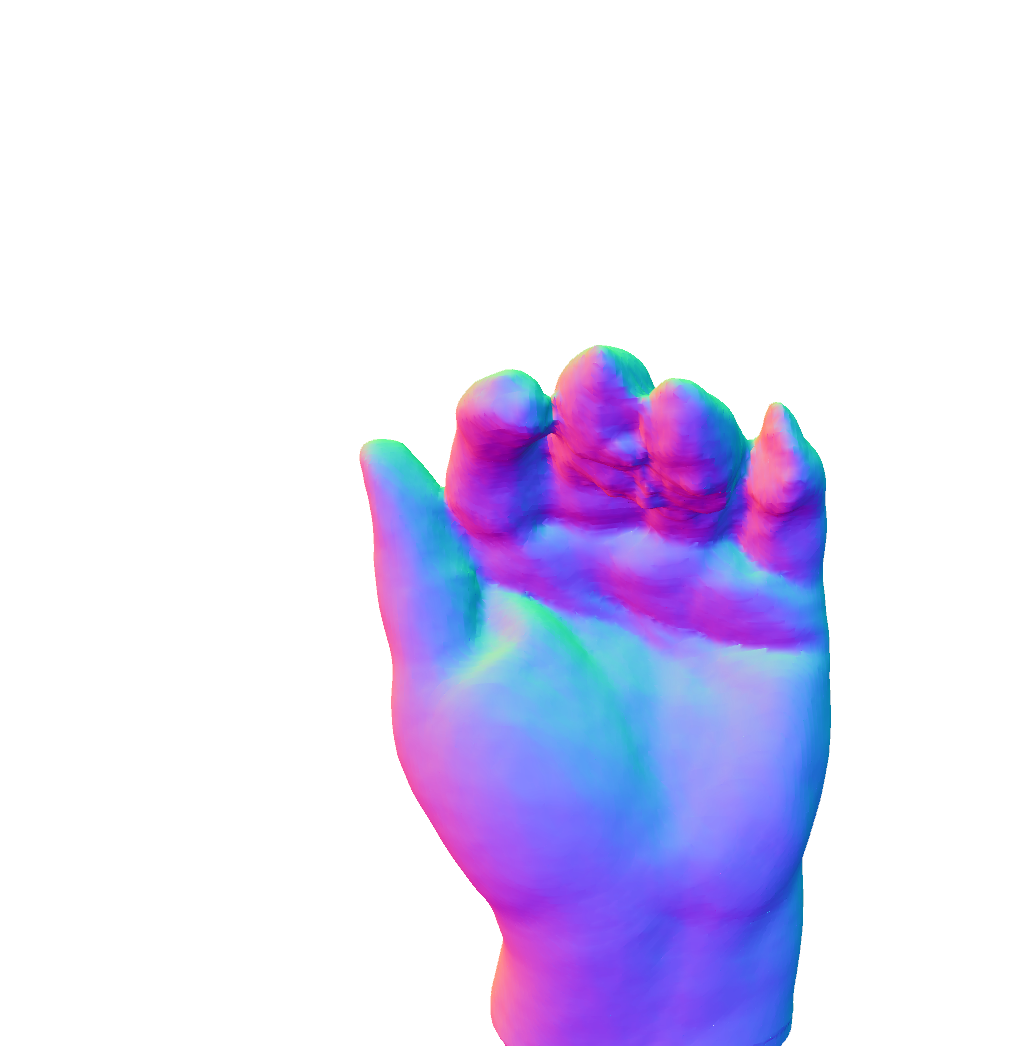}};
		
		\node (p06) at(p05.east) [anchor=west, xshift=\yoff]{\includegraphics[width=\imw]{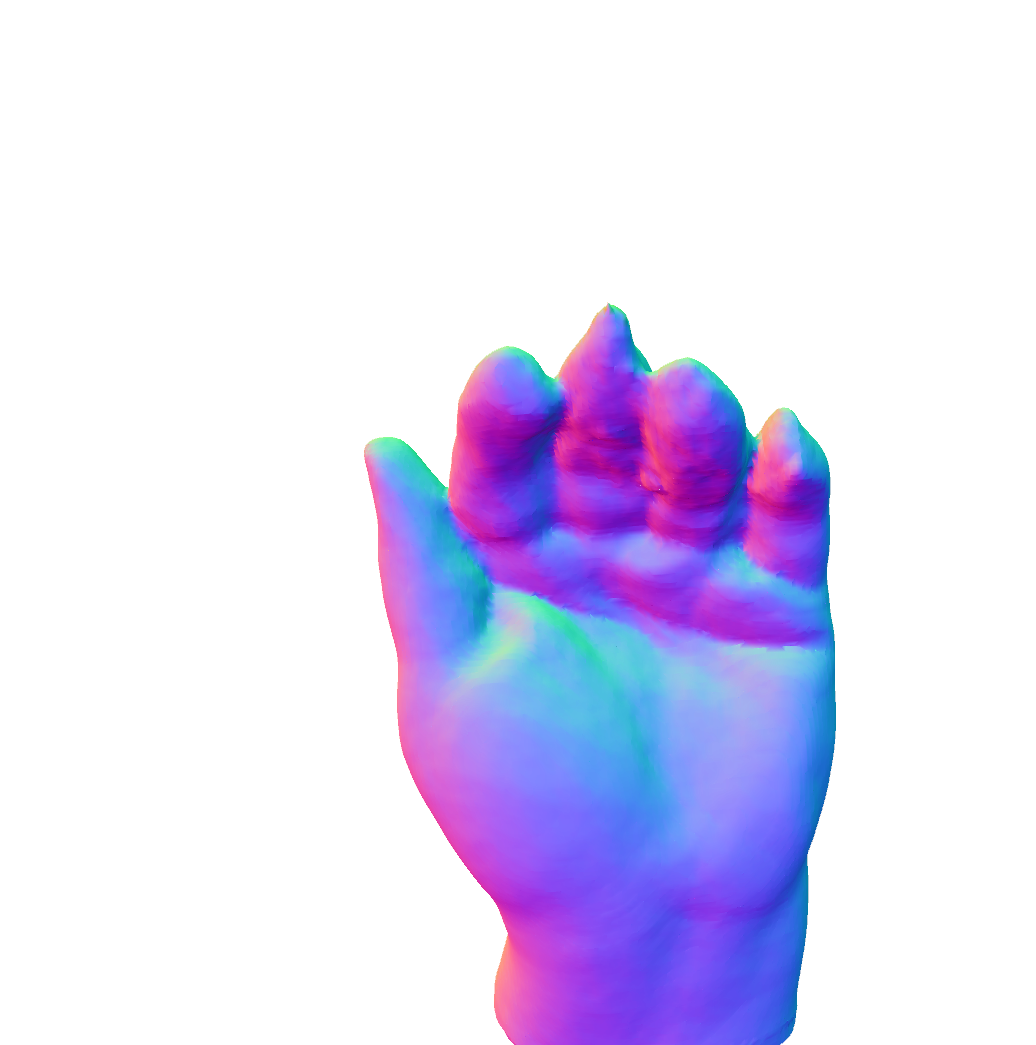}};
		\node (p07) at(p06.east) [anchor=west, xshift=\yoff] {\includegraphics[width=\imw]{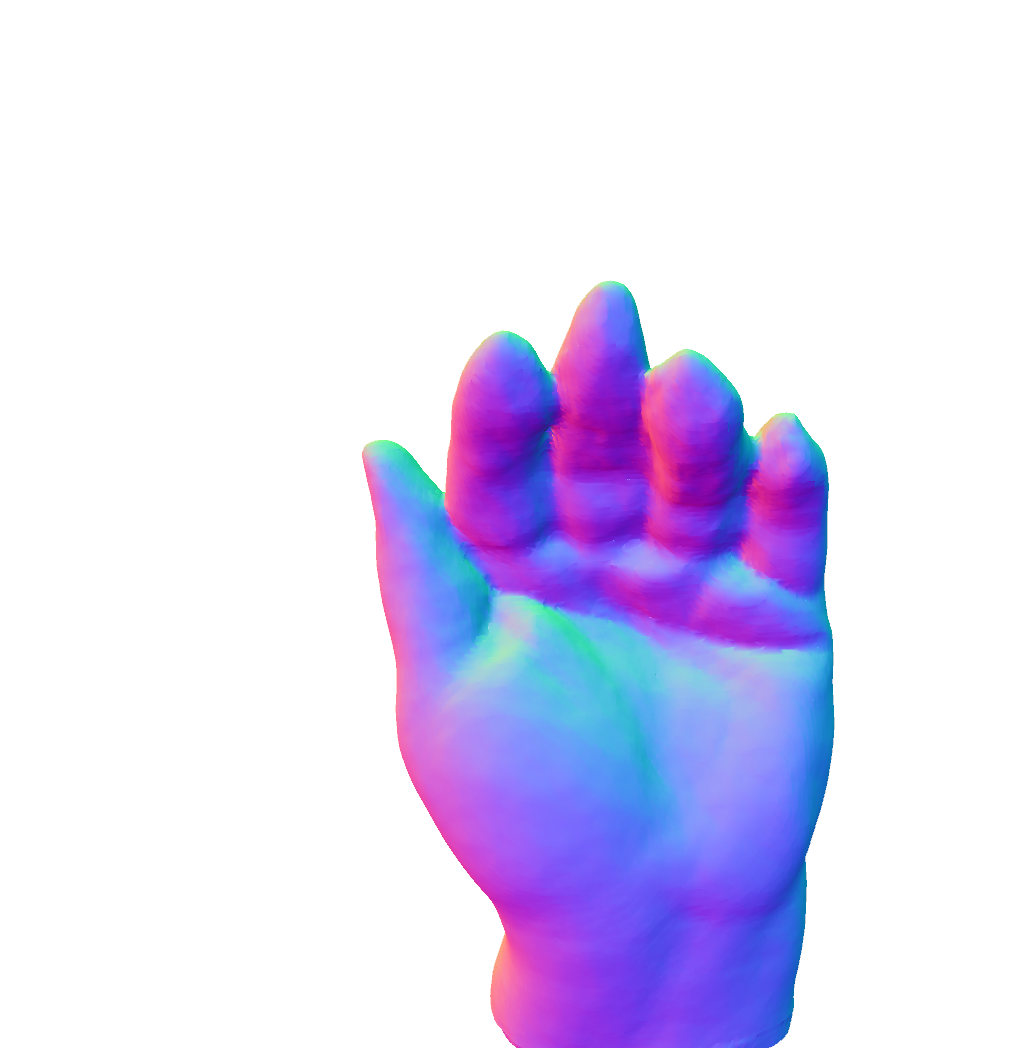}};
		\node (p08) at(p07.east) [anchor=west, xshift=\yoff]{\includegraphics[width=\imw]{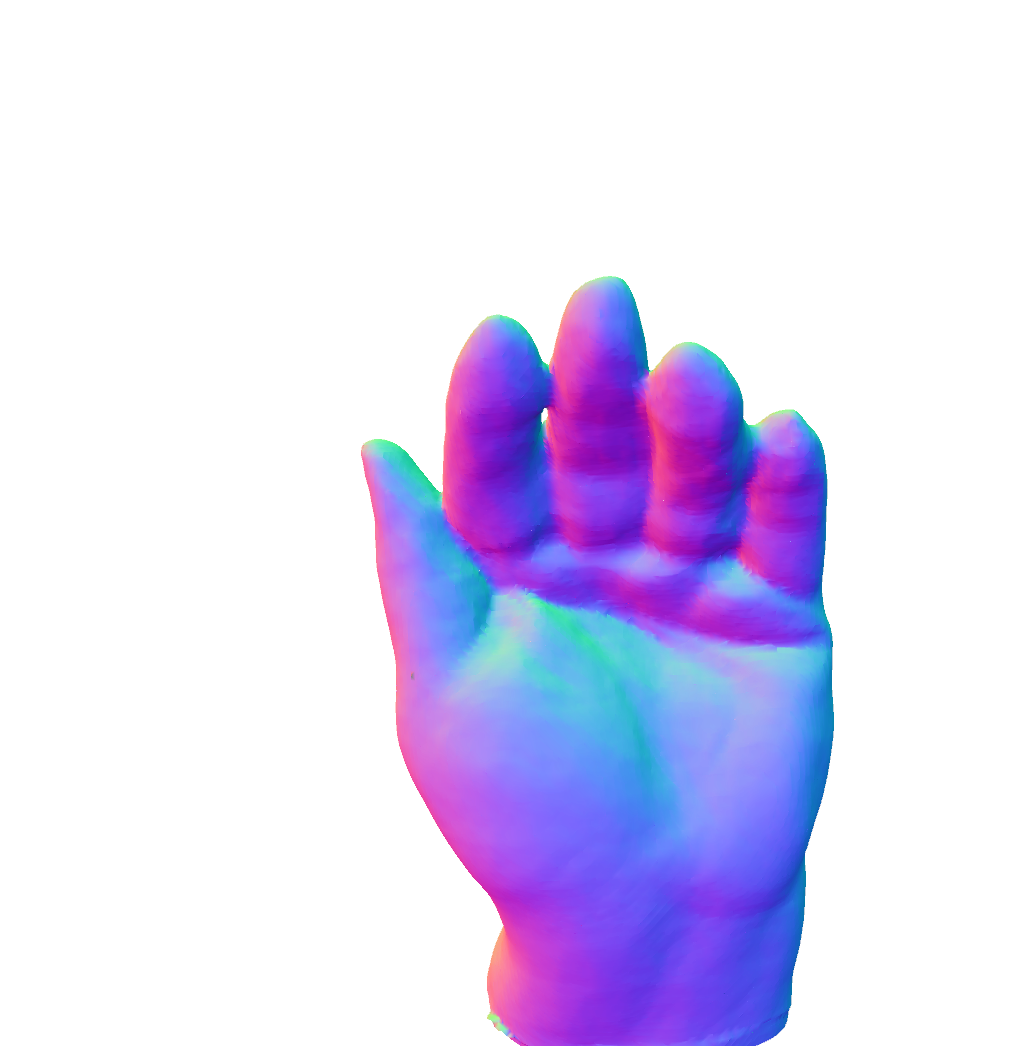}};
		\node (p09) at(p08.east) [anchor=west, xshift=\yoff] {\includegraphics[width=\imw]{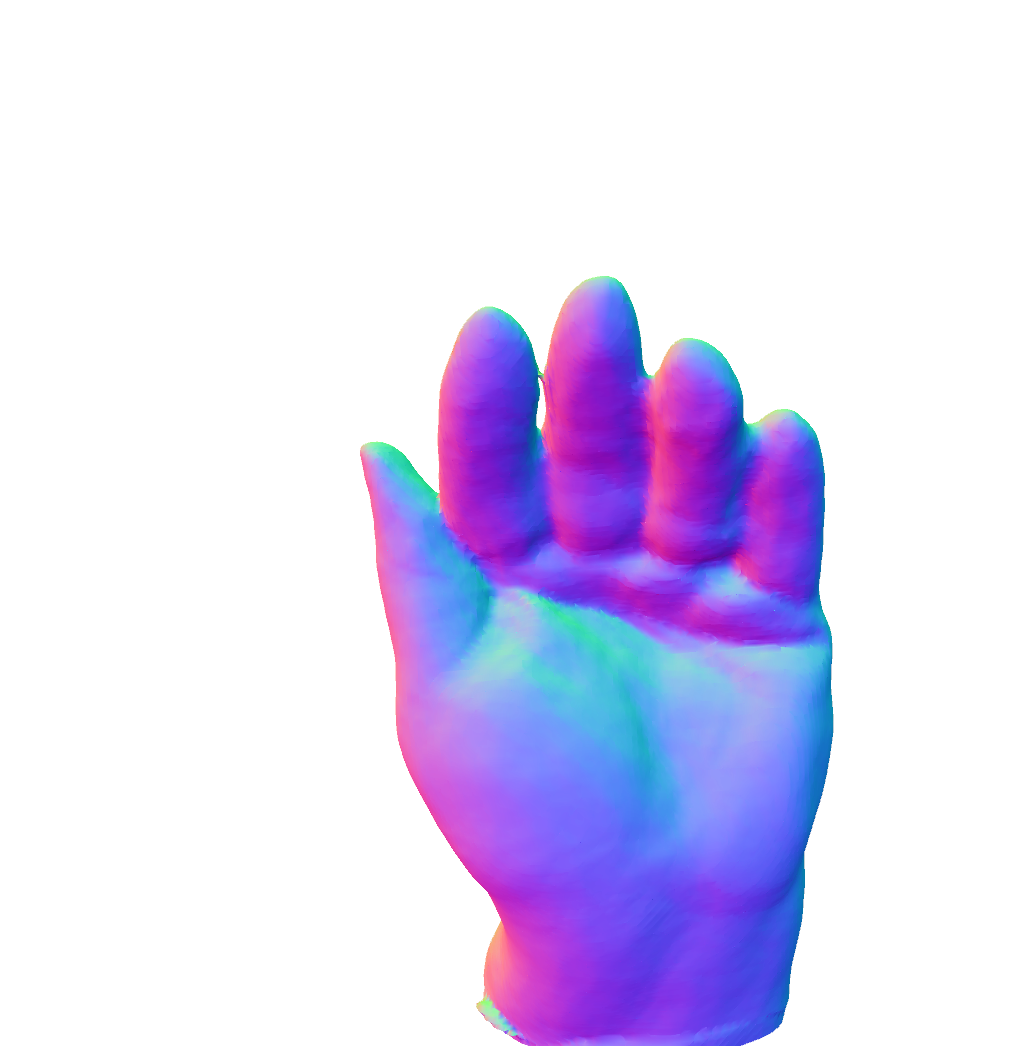}};
		\node (p10) at(p09.east) [anchor=west, xshift=\yoff]{\includegraphics[width=\imw]{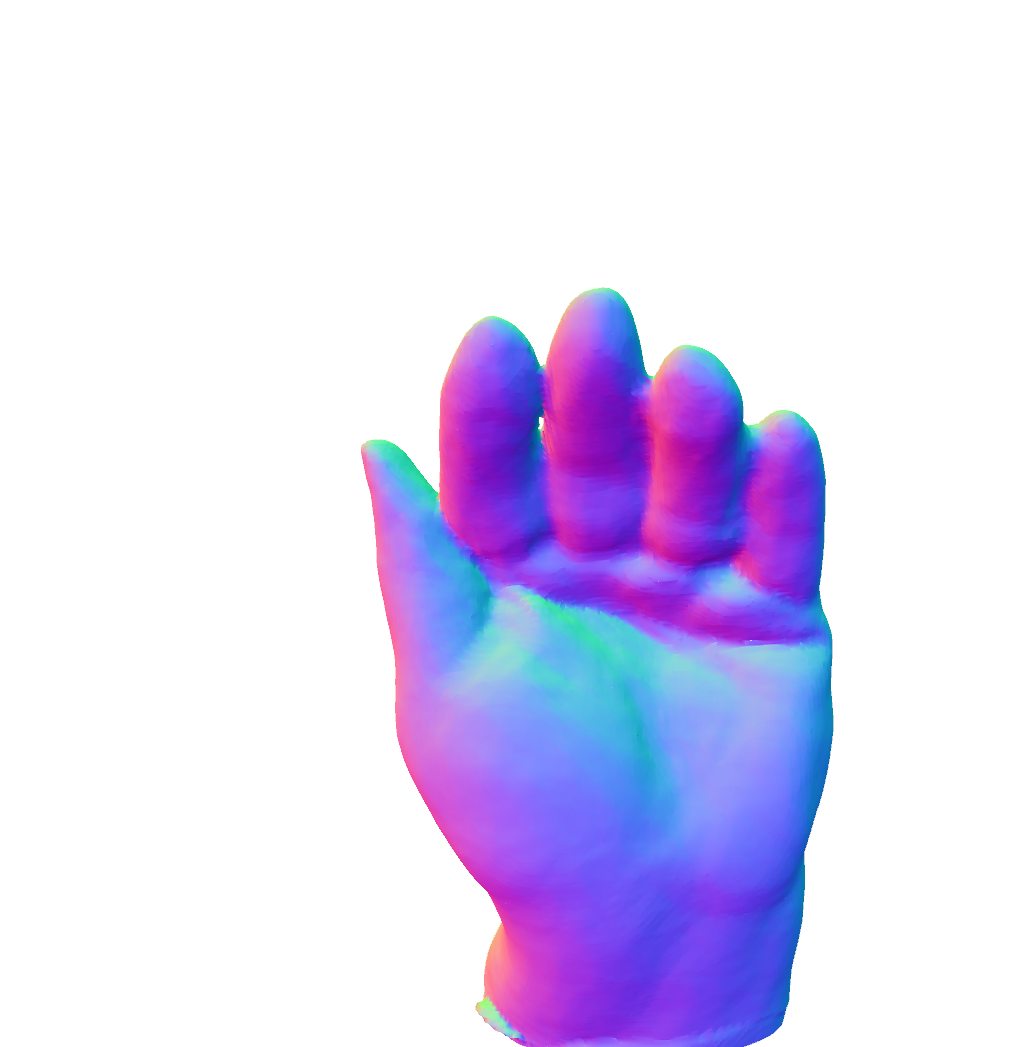}};
		
		 \draw[draw=red] (-0.7,0.7) rectangle ++(\recz,-\recz);
		 \draw[draw=red] (15.3,0.7) rectangle ++(\recz,-\recz);
	\end{tikzpicture}
	\caption{Artifacts caused by the linear interpolation of two deformation codes that are not close enough on the embedding manifold. }
	\vspace{-1em}
	\label{fig:linear_interpolation}
\end{figure*}

%% file: supp/tsne.tex
\subsection{t-SNE visualization}
\begin{figure}
\centering
\def\imw{15ex}
\def\yoff{2pt}
\begin{tikzpicture}[inner sep=0pt]
    \begin{scope}
      \node(tsne) at(0,0)                    {\includegraphics[height=40ex]{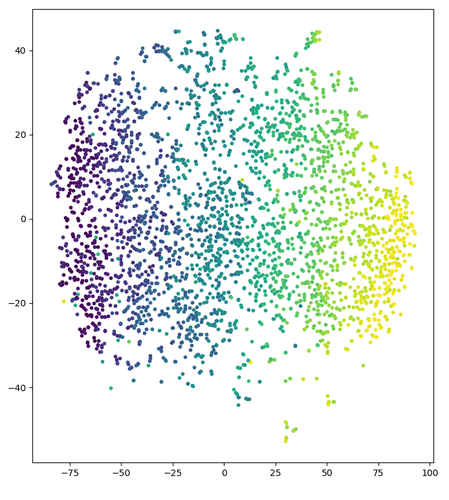}};
      \node(cmap) at(tsne.east)[anchor=west,xshift=3ex] {\includegraphics[height=40ex]{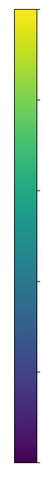}};
      \node(traj) at(tsne.east)[anchor=west,xshift=8ex,yshift=1ex] {\includegraphics[width=0.33\textwidth, angle=-90, origin=c]{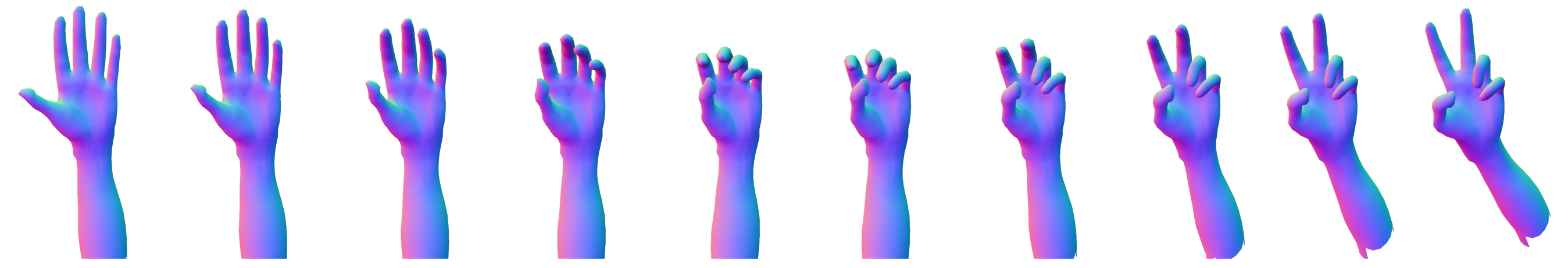}};
      \draw[->, >=stealth] [thin] (3.2,-2.7) -- ++(0, 5.7);
      \node at(3.2, -2.8)[font=\footnotesize]{time};
    \end{scope}

    \begin{scope}[xshift=-20ex, yshift=-30ex]
    
    \end{scope}

\end{tikzpicture}
  \caption{The t-SNE visualization of deformation codes for the synthetic dataset, which contains 40 test hand shapes deforms along the same trajectory. For each hand shape, the same deformation is colored the same.}
  \label{fig:tsne}
\end{figure}

To better understand the learnt deformation space, we further conduct an experiment with t-SNE analysis. 
Due to the lack of ground truth annotation on the 3DH dataset, we construct a synthetic hand dataset. 
We use a LBS-based hand model similar to MANO, which is controlled by a set of pose parameters and shape parameters. The pose parameters can drive different hand shapes to deform alike. 
For the experiment, 140 random hand shapes are sampled, and a smooth deformation trajectory of 80 timestamps is applied to each hand to obtain the meshes. \cref{fig:tsne} shows the deformation trajectory.
We randomly select 100 hands for training, then reconstruct the unseen hands. 
\cref{fig:tsne} shows the t-SNE analysis of the obtained deformation codes for the 40 test hands. 
For visualization, we colormap the results acoordingly to the trajectory. This means the same color is applied to 40 hand shapes at the same timestamp, since these codes corresponds to the same deformation.
The results indicate the same color tends to cluster, while the distribution of deformation codes smoothly transite according to the trajectory, suggesting the learnt deformation space is smooth.

%% file: supp/table_4dscan.tex
\begin{table}
	\centering
	\footnotesize
	\caption{Quantitative evaluation of end-to-end 4D dynamic reconstruction. For all methods, the first frame is initialized with the same 6DoF pose computed by our proposed pose prediction network.  }
	\label{tab:eval4d}
	\thickmuskip=0mu
	\setlength{\tabcolsep}{0.5pt}
	\begin{tabular}{C{3ex} L{8ex} R{8ex} R{8ex} R{9ex} R{9ex} R{9ex} R{9ex}}
		\toprule
		&method &$\mu(\text{CD})$ &$\sigma(\text{CD})$ &$\mu\left(^r_g\text{CD}\right)$ &$\sigma\left(^r_g\text{CD}\right)$ &$\mu\left(^g_r\text{CD}\right)$ &$\sigma\left(^g_r\text{CD}\right)$ \\
		\midrule
		\multirow{3}{*}{\rot{\shortstack[c]{seq-1}}} %
		&IGR~\cite{Gropp:etal:ICML2020}      &1.8525& 0.1645& 3.0747& 0.3294& \textbf{0.6295} & 0.0160 \\
		&IGR-PE      &1.9427& 0.1571& 3.2375& 0.3141& 0.6481& 0.0254 \\
		&\method& \textbf{1.6501}& \textbf{0.1416} & \textbf{2.6677} & \textbf{0.2826} & 0.6314& \textbf{0.0153} \\
		\midrule
		\multirow{3}{*}{\rot{\shortstack[c]{seq-2}}} %
		&IGR~\cite{Gropp:etal:ICML2020}      &1.1276& 0.1249& 1.5742& 0.2371& 0.6814& \textbf{0.0270}\\
		&IGR-PE      &1.1452& 0.0479& 1.6083& 0.0964& 0.6816& 0.0314 \\
		&\method& \textbf{1.0378} & \textbf{0.0402}& \textbf{1.4059}& \textbf{0.0959}& \textbf{0.6694}& 0.0308\\
		\midrule
		\multirow{3}{*}{\rot{\shortstack[c]{seq-3}}} %
		&IGR~\cite{Gropp:etal:ICML2020}      &1.7722& 0.2356& 2.8745& 0.4866& 0.6706& \textbf{0.0345}\\
		&IGR-PE      &1.2376& \textbf{0.0652}& 1.8063& \textbf{0.1351}& \textbf{0.6703}& 0.0433\\
		&\method &\textbf{1.1419}& 0.1095& \textbf{1.6050}& 0.2316& 0.6785& 0.0379\\
		\midrule
		\multirow{3}{*}{\rot{\shortstack[c]{seq-4}}} %
		&IGR~\cite{Gropp:etal:ICML2020}      &2.2321& 0.2051& 3.8129& 0.4067& 0.6511& 0.0270\\
		&IGR-PE      &1.6215& 0.1005& 2.5742& 0.1768& 0.6685& 0.0412\\
		&\method &\textbf{1.2992}& \textbf{0.0647}& \textbf{1.9350}& \textbf{0.1246}& \textbf{0.6631}& \textbf{0.0252} \\
		\midrule
		\multirow{3}{*}{\rot{\shortstack[c]{seq-5}}} %
		&IGR~\cite{Gropp:etal:ICML2020}      &1.3223& 0.4343& 1.9493& 0.8763& \textbf{0.6954}& \textbf{0.0409}\\
		&IGR-PE      &1.2079& 0.2024& 1.6732& 0.4069& 0.7419& 0.0647\\
		&\method &\textbf{1.1705}& \textbf{0.1668}& \textbf{1.6193}& \textbf{0.3503}& 0.7210& 0.0503\\
		\midrule
		\multirow{3}{*}{\rot{\shortstack[c]{seq-6}}} %
		&IGR~\cite{Gropp:etal:ICML2020}      &2.1419& 0.9500& 3.6017& 1.9295& \textbf{0.6823}& \textbf{0.0411}\\
		&IGR-PE      &1.6259& 0.5190& 2.5412& 1.0420& 0.7114& 0.1254\\
		&\method &\textbf{1.4510}& \textbf{0.3493}& \textbf{2.2123}& \textbf{0.7380}& 0.6896& 0.0766\\
		\midrule
		\multirow{3}{*}{\rot{\shortstack[c]{seq-7}}} %
		&IGR~\cite{Gropp:etal:ICML2020}      &1.3333& \textbf{0.0642}& 1.9999& \textbf{0.1492}& \textbf{0.6673}& \textbf{0.0548 }\\
		&IGR-PE      &1.2585& 0.1227& 1.7821& 0.1518& 0.7354& 0.1228\\
		&\method &\textbf{1.1122}& 0.1284& \textbf{1.5214}& 0.1766& 0.7028& 0.1152\\
		\midrule
		\multirow{3}{*}{\rot{\shortstack[c]{seq-8}}} %
		&IGR~\cite{Gropp:etal:ICML2020}      &1.2618& 0.0749& 1.8547& 0.1505& 0.6695& \textbf{0.0254}\\
		&IGR-PE      &1.0717& 0.0804& 1.4675& 0.1743& 0.6756& 0.0380\\
		&\method &\textbf{1.0275}& \textbf{0.0535}& \textbf{1.3960}& \textbf{0.1140}& \textbf{0.6585}& 0.0294\\
		\midrule
		\multirow{3}{*}{\rot{\shortstack[c]{seq-9}}} %
		&IGR~\cite{Gropp:etal:ICML2020}      &1.4103& 0.4432& 1.9185& 0.5472& 0.9024& 0.3944\\
		&IGR-PE      &1.2189& \textbf{0.0937}& 1.7887& \textbf{0.1844}& \textbf{0.6489}& \textbf{0.0345}\\
		&\method &\textbf{1.0005}& 0.1220& \textbf{1.3454}& 0.1878& 0.6561& 0.0776\\	
		\midrule
		\multirow{3}{*}{\rot{\shortstack[c]{seq-10}}} %
		&IGR~\cite{Gropp:etal:ICML2020}      &1.3663& 0.2815& 2.0968& 0.5447& \textbf{0.6364}& \textbf{0.0283}\\
		&IGR-PE      &1.4180& 0.2263& 2.1672& 0.4131& 0.6698& 0.0680\\
		&\method &\textbf{1.2926}& \textbf{0.2044}& \textbf{1.9042}& \textbf{0.3564}& 0.6817& 0.1128\\		
		\midrule
		\multirow{3}{*}{\rot{\shortstack[c]{seq-11}}} %
		&IGR~\cite{Gropp:etal:ICML2020}      &1.0424& \textbf{0.0598}& 1.4323& \textbf{0.1096}& \textbf{0.6530}& 0.0258\\
		&IGR-PE      &1.0781& 0.0653& 1.4818& 0.1202& 0.6747& 0.0206\\
		&\method &\textbf{1.0110}& 0.0608& \textbf{1.3575}& 0.1099& 0.6642& \textbf{0.0161}\\
		\midrule
		\multirow{3}{*}{\rot{\shortstack[c]{seq-12}}} %
		&IGR~\cite{Gropp:etal:ICML2020}      &1.3322& 0.2404& 2.0517& 0.4678& 0.6132& \textbf{0.0308}\\
		&IGR-PE      &1.6677& 0.3661& 2.6930& 0.6646& 0.6431& 0.0975\\
		&\method &\textbf{1.2244}& \textbf{0.1883}& \textbf{1.8404}& \textbf{0.3528}& \textbf{0.6078}& 0.0426\\	
		\midrule
		\multirow{3}{*}{\rot{\shortstack[c]{seq-13}}} %
		&IGR~\cite{Gropp:etal:ICML2020}      &1.1210& \textbf{0.1436}& 1.6314& \textbf{0.2639}& 0.6109& 0.0820\\
		&IGR-PE      &1.1903& 0.3033& 1.7062& 0.4356& 0.6731& 0.1944\\
		&\method &\textbf{0.8979}& 0.1496& \textbf{1.1892}& 0.2755& \textbf{0.6070}& \textbf{0.0425}\\	
		\toprule
    	 \multirow{3}{*}{\rot{\shortstack[c]{\textbf{average}}}}
    	&IGR~\cite{Gropp:etal:ICML2020}      &1.4858& 0.2632& 2.2979& 0.4999& 0.6741& 0.0637 \\
    	&IGR-PE      &1.3603& 0.1808& 2.0406& 0.3319& 0.6802& 0.0698 \\
    	&\method &\textbf{1.1782}& \textbf{0.1368}& \textbf{1.6923}& 0\textbf{.2612}& \textbf{0.6639}& \textbf{0.0517} \\
		\bottomrule
	\end{tabular}
\end{table}